\documentclass[3p,preprint,authoryear,review]{elsarticle}
\usepackage{paralist}

\usepackage{amssymb}
\usepackage{lipsum}

\usepackage{amsmath}
\usepackage{mathtools}
\usepackage{amsthm}
\usepackage{bm}
\usepackage{bbm}
\usepackage{enumerate}
\usepackage{subcaption}
\usepackage{graphicx}
\usepackage{booktabs}
\usepackage{makecell}
\usepackage{multicol}
\usepackage{multirow}

\usepackage{framed} %
\usepackage{nomencl} %
\makenomenclature
\renewcommand*\nompreamble{\begin{framed}\footnotesize\begin{multicols}{2}}
\renewcommand*\nompostamble{\end{multicols}\end{framed}}

\usepackage[table,xcdraw]{xcolor}
\definecolor{bestgreen}{rgb}{1.0,0.99,0.82} %
\definecolor{bestblue}{rgb}{0.67, 0.92, 0.99}%
\definecolor{lightgray}{rgb}{0.9,0.9,0.9}

\DeclareMathOperator*{\argmin}{arg\,min}

\newcommand{\plen}{H}
\newcommand{\clen}{C}
\newcommand{\rlen}{C_{\text{ref}}}
\newcommand{\tlen}{T}
\newcommand{\ttlen}{T_\text{test}}
\newcommand{\ypast}{\overleftarrow{Y}}
\newcommand{\ynext}{\overrightarrow{Y}}
\newcommand{\ypasttrue}{\overleftarrow{Y}^{\sf true}}
\newcommand{\ytrue}{\overrightarrow{Y}^{\sf true}}

\journal{}
\begin{document}
\begin{frontmatter}

\title{{\bfseries Probabilistic  Forecasting for Building Energy Systems using Time-Series Foundation Models}}

\author[mit]{Young-Jin Park\corref{intern}}
\affiliation[mit]{organization={Massachusetts Institute of Technology (MIT)},%
city={Cambridge},
state={MA},
country={USA}}
\cortext[intern]{Work done during YJP's internship at MERL.}
\cortext[cor]{Corresponding author.}
\author[merl]{Fran\c{c}ois Germain}
\author[merl]{Jing Liu}
\author[merl]{Ye Wang}
\author[merl]{Toshiaki Koike-Akino}
\author[merl]{Gordon Wichern}
\author[mit]{Navid Azizan}
\author[merl]{Christopher Laughman}
\author[merl]{Ankush Chakrabarty\corref{cor}}
\ead{achakrabarty@ieee.org}
\affiliation[merl]{organization={Mitsubishi Electric Research Laboratories (MERL)},
city={Cambridge},
state={MA},
country={USA}}

\begin{abstract}
Decision-making in building energy systems critically depends on the predictive accuracy of relevant time-series models.  In scenarios lacking extensive data from a target building, foundation models (FMs) represent a promising technology that can leverage prior knowledge from vast and diverse pre-training datasets to construct accurate probabilistic predictors for use in decision-making tools. This paper investigates the applicability and fine-tuning strategies of time-series foundation models (TSFMs) in building energy forecasting. We analyze both full fine-tuning and parameter-efficient fine-tuning approaches, particularly low-rank adaptation (LoRA), by using real-world data from a commercial net-zero energy building to capture signals such as room occupancy, carbon emissions, plug loads, and HVAC energy consumption. Our analysis reveals that the zero-shot predictive performance of TSFMs is generally suboptimal. To address this shortcoming, we demonstrate that employing either full fine-tuning or parameter-efficient fine-tuning significantly enhances forecasting accuracy, even with limited historical data. Notably, fine-tuning with low-rank adaptation (LoRA) substantially reduces computational costs without sacrificing accuracy. Furthermore, fine-tuned TSFMs consistently outperform state-of-the-art deep forecasting models (e.g., temporal fusion transformers) in accuracy, robustness, and generalization across varying building zones and seasonal conditions. These results underline the efficacy of TSFMs for practical, data-constrained building energy management systems, enabling improved decision-making in pursuit of energy efficiency and sustainability.
\end{abstract}

\end{frontmatter}

\newpage
\nomenclature{ARIMA}{Auto-regressive integrated moving average}
\nomenclature{BES}{Building Energy System}
\nomenclature{FM}{Foundation model}
\nomenclature{FT}{Fine-tuning}
\nomenclature{FLOPS}{Floating-point operations}
\nomenclature{GPU}{Graphics processing unit}
\nomenclature{CPU}{Central processing unit}
\nomenclature{IEA}{International Energy Agency}
\nomenclature{LoRA}{Low-Rank Adaptation}
\nomenclature{LSTM}{Long-short term memory}
\nomenclature{MAE}{Mean absolute error}
\nomenclature{MASE}{Mean absolute scaled error}
\nomenclature{MPC}{Model predictive control}
\nomenclature{MSIS}{Mean scaled interval score}
\nomenclature{PEFT}{Parameter-efficient fine-tuning}
\nomenclature{RL}{Reinforcement learning}
\nomenclature{RMSSE}{Root-mean-squared scaled error}
\nomenclature{TFT}{Temporal fusion transformer}
\nomenclature{TSFM}{Time-Series Foundation Model}
\nomenclature{wQL}{Weighted quantile loss}
\nomenclature{$Y$}{Measured output (univariate time-series)}
\nomenclature{$\pi$}{Probability density function}
\nomenclature{$t$}{Time index}
\nomenclature{$\plen$}{Forecasting predictive window}
\nomenclature{$\clen$}{Forecasting context window}
\nomenclature{$\theta$}{Neural weights}
\nomenclature{$\mathcal D$}{Time-series dataset}
\nomenclature{$\mathcal L$}{Loss function}
\nomenclature{$r$}{LoRA rank}
\nomenclature{$\omega$}{LoRA-adapted weights}
\nomenclature{$Q$}{Query tensor for attention}
\nomenclature{$K$}{Key tensor for attention}
\nomenclature{$V$}{Value tensor for attention}
\nomenclature{$\zeta$}{Performance evaluation metric}
\nomenclature{$\beta$}{Quantile}
\nomenclature{$\delta\omega$}{LoRA adaptation}
\printnomenclature

\section{Introduction}
\label{sec:introduction}

Building operations account for a substantial share of global energy consumption and carbon dioxide (CO\textsubscript{2}) emissions, as repeatedly highlighted by the International Energy Agency (IEA) \citep{perez2008review, nejat2015global, bouckaert2021net}.  Automatic decision-making and control algorithms can be used to reduce the negative environmental impact of these emissions by regulating the building behavior to manage occupant comfort and minimize energy consumption.  Accurate predictive models for building system dynamics are an integral component of these control algorithms~\citep{cox2019real}, as their forecasts of building energy consumption and other internal signals play a critical role in evaluating the energy efficiency of buildings, detecting system faults, and enabling intelligent control and optimization of energy use in building energy management systems~\citep{yang2014developing, ahmad2018comprehensive, lei2021building, khalil2022machine}. Furthermore, these models support the integration of renewable energy sources and the implementation of advanced energy-saving strategies, ultimately contributing to the creation of more sustainable built environments~\citep{sorourifar2024ccta,chakrabarty2024ccta,azizan2020optimization}. From a macro perspective, these predictive models can not only enhance the operational efficiency of buildings, but also align with global efforts to mitigate climate change and promote sustainable development~\citep{zhang2025deep}. 

While physical models or interpretable reduced-order models can be used to predict the behavior of some internal model dynamics such as heat flow, other variables such as occupant-induced effects and ambient conditions require models that can learn directly from data, as the underlying behaviors may be too complex to abstract at the level of predictive models needed for decision-making. Since we are predominantly interested in time series signals in this paper, time-series forecasting represents the critical technology under consideration and is an essential step for decision-making. Whereas classical decision-making frameworks rely on deterministic forecasting, recent advancements in stochastic model predictive control (MPC) and reinforcement learning (RL) have exploited probabilistic forecasting models with success. In particular, control policies determined by taking stochasticity into account have been shown to balance robustness, safety, and optimality more effectively than deterministic control~\citep{mohebi2025chance,arroyo2022reinforced,heidari2022reinforcement}.

While the first stochastic model variants were proposed almost a century ago, such as autoregressive integrated moving average (ARIMA) models, they often fall short in providing sufficiently accurate predictions due to their inability to capture complex patterns~\citep{ahmad2018comprehensive, bourdeau2019modeling, geraldi2022data}.  In recent years, there has thus been a shift towards the use of machine learning algorithms and deep learning architectures, which can autonomously identify patterns in historical time-series data to predict future trends more effectively. Deep neural network approaches (e.g., long short-term memory (LSTM) recurrent networks~\citep{graves2012long}) and attention-based models~\citep{vaswani2017attention} (e.g., the temporal fusion transformer (TFT)~\citep{lim2021temporal}), have been successfully applied in the building system context for interpretable learning via attention mechanisms~\citep{zheng2023interpretable}, knowledge transfer between buildings or building classes~\citep{liang2023domain, kim2024forecasting, xing2024transfer,sun2025deep}, incorporating spatio-temporal interactions~\citep{dong2025building}, and forecasting occupant-centric signals, such as plug loads~\citep{botman2024building}.

Deep networks possess the capability to represent and reproduce complex time-series patterns, but are typically highly over-parameterized~\citep{azizan2021stochastic} and require copious amounts of training data to generalize well~\citep{park2022large}. Without significant training data, they can produce unreliable forecasts, while deep network-based methods are prone to issues like overfitting or mode collapse~\citep{fan2017short, jung2020worrying, morcillo2024deep}. Such limitations are further exacerbated in the context of probabilistic forecasting, where learning a good distribution (not just a mean) is dependent not only on the sheer quantity of data, but also upon the availability of repeated instances of data over the forecast horizon. 
This significant but common limitation is a major bottleneck to uncertainty quantification in general, and probabilistic forecasting in particular. In most practical settings, large quantities of data are not available from the target building energy system under consideration when a decision-making algorithm has to be deployed. Such scenarios arise in new construction, in which case the amount of time sensors have been installed is small, or if the building occupant is unwilling to share data due to privacy concerns. In such situations, we suggest that one elegant framework for learning such probabilistic forecasting networks leverages available data from other systems. While data for the same signal (e.g., occupancy) from another system will most likely not be identical to the target system under consideration, similarities in the distributional form can be exploited to speed up and improve learning even without a large amount of target data. 

Time-series foundation models (TSFM) represent a prime candidate to enable such an approach for learning distributions with limited data from the target system. Foundation models (FMs)~\citep{bommasani2021opportunities} have generally emerged as a powerful tool, demonstrating excellent performance across numerous machine learning domains, particularly in natural language processing and computer vision~\citep{brown2020language, radford2021learning, kirillov2023segment}. Conceptually, FMs are very large networks, often containing billions of parameters, that are trained on simple self-supervised generative tasks (e.g., next-token prediction) using massive datasets that span multiple domains of machine learning. For example, TSFMs are pre-trained on gigantic datasets spanning time-series data from domains that include healthcare, traffic, energy, stocks, and weather. The pre-trained model, or parts of it, can then be used for downstream tasks with or without added fine-tuning, i.e., a few additional training iterations with typically limited task-specific data. This approach contrasts with classical deep learning, which typically learns from scratch using only task-specific data. FMs possess two beneficial characteristics that allow them to overcome the bottleneck of limited target data. First, as these networks exploit the most recent deep learning architectural advances and are heavily over-parameterized, they have the capacity to adapt to various function landscapes, allowing them to be expressive enough to contribute to different problem domains. Second, FMs have a  strong generalization capability, as they are incentivized to identify cross-domain patterns within a diverse range of pre-training datasets that would be effective at solving their pre-training generative task (i.e., learning the underlying data distributions).  These cross-domain patterns can then be exploited for specific use cases or downstream tasks through transfer learning processes such as fine-tuning. Despite their impressive track record in language and vision, the applicability and effectiveness of FMs in real-world time-series forecasting problems remain largely unreported. 

This work investigates an underexplored application of probabilistic TSFMs for building energy systems with \textbf{real building sensor data}.
While there have been recent contributions in the general area of using pre-trained generative transformers in building applications~\citep{zhang2025automated,liao2025timegpt}, our work extends these ideas in significant ways. First, we consider GPT-like foundation models that are explicitly tailored towards time-series rather than language modeling. This is important because recent results show that language models do not perform well on time-series tasks~\citep{tan2024language}. Second, we provide a comparison between TSFMs, not one TSFM against other non-FM deep forecasting methodologies. Third, we present our results on multi-signal, multi-month real data, and present the use of architectural adaptation mechanisms (as opposed to prompt tuning, which may change from version to version of the GPT) for improving performance.
Furthermore, while prior work has demonstrated the potential of TSFMs for building energy systems~\citep{mulayim2024time}, this paper offers a more in-depth and comprehensive analysis of how different ways of leveraging TSFMs can influence forecasting performance.

This paper serves as both an investigation into the utility of TSFMs on probabilistic building energy forecasting as well as a guide to improving the performance of existing (usually open-source) TSFMs that have been pre-trained on a wide variety of general time-series signals. By design, TSFMs are initially set up to output probabilistic forecasts, so that they can be used for `zero-shot' forecasting (i.e., without fine-tuning, using only the information from a limited context of past data from the target system). However, we will demonstrate that leveraging these models in such a zero-shot setting does not always yield accurate results for specific real-world building energy systems due to a variety of factors. For instance, zero-shot performance is expected to be hampered if the pre-training and target data distribution shift is too large, if there is severe imbalance in the pre-training data, or if there are domain-related constraints or long-term covariates that are impossible to incorporate through context alone and would necessitate adaptation. We provide evidence of these phenomena using real data from the SUSTIE sustainable net-zero energy building of Mitsubishi Electric, for which we have collected data for multiple months over 2021--2023. Note that this dataset is representative of a commercial office building, and therefore affected by covariates such as national holidays, work-week, corporate office hours, as well as general trends like seasonality and diurnal variations. We therefore explore different fine-tuning methods to adapt the TSFM to the specific characteristics of the SUSTIE building data. While effective, fine-tuning all the parameters of the TSFM can be computationally intensive and may risk overfitting, especially with limited data. Alternatively, we investigate Low-Rank Adaptation (LoRA)~\citep{hu2021lora}, a recently proposed technique that injects trainable low-rank decomposition matrices into each layer of the model, significantly reducing the number of trainable parameters and computational resources required.

The main \textbf{contributions} of this paper are as follows: (i) We first investigate the applicability of TSFMs on real office building data by evaluating the zero-shot prediction quality of the base TSFMs to assess their performance without any task-specific fine-tuning. This is to understand the claim made about zero-shot generalization performance of TSFMs in real engineering scenarios; i.e., to investigate whether current TSFMs could be used without customization as a plug-and-play solution to energy forecasting. (ii) We next demonstrate the case study using the state-of-the-art TSFM model, \texttt{Chronos}, demonstrating the effectiveness of fine-tuning in TSFMs by examining different fine-tuning approaches and comparing the zero-shot baselines and prominent deep forecasting benchmark models. We further investigate the trade-off between fine-tuning and context length at inference to analyze the effectiveness and amount of fine-tuning required to get satisfactory prediction accuracy. (iii) We then assess the accuracy and robustness of the proposed TSFM approaches across different environments and seasons, as well as in limited data settings. This is to simulate practical forecasting problems where unknown external factors affect the underlying dynamics, and such factors have to be accounted for during forecasting without relying on massive datasets. (iv) Finally, we evaluate the generalization capabilities of fine-tuned TSFMs to unseen tasks and systems by testing their performance on building zones not seen at training time. This is especially critical for long-term adoption of TSFM-like technology in decision-making for buildings.

The remainder of this paper is organized as follows. Section~\ref{sec:method} describes the probabilistic time-series forecasting problem more formally, and delineates the various fine-tuning approaches considered in this work. Section~\ref{sec:experiments}  presents the experimental setups in our real-world building system and the metrics of forecasting performance. In section~\ref{sec:results}, we discuss the overall predictive quality in various settings including zero-shot and with fine-tuning. We also provide comparisons to well-known deep forecasting algorithms and evaluate the effectiveness of TSFMs in practical settings, such as across seasons, on unseen building zones, and with severely limited data. Finally, Section~\ref{sec:conclusion} concludes the paper and outlines future research directions.

\section{Methodology}
\label{sec:method}
We propose a probabilistic forecasting framework that leverages time-series foundation models for building energy systems.
We begin by formulating the problem as a probabilistic time-series forecasting task, as detailed in Section~\ref{sec:forecast}.
Subsequently, in Section~\ref{sec:fm}, we detail the fine-tuning methods applied to adapt the foundation models to our specific application, enhancing their accuracy and reliability in the context of building energy systems.

\subsection{Probabilistic Forecasting in Building Energy Systems}
\label{sec:forecast}

Consider a zone in the building equipped with sensors that collect time-series data, such as occupancy, CO\textsubscript{2}, or plug loads. Let $Y:= \{y_1, \ldots, y_t, \ldots\}$ denote the univariate time-series of one of the monitored signals. The primary objective of probabilistic forecasting is to predict the conditional probability distribution
\[
\pi_\theta := \pi_\theta(\ynext_{t,\plen}\mid\ypast_{t,\clen})
\]
of the future sequence
\begin{equation}\label{eq:ynext}
	\ynext_{t,\plen}:= \{y_{t+1}, y_{t+2}, \ldots, y_{t+\plen}\}
\end{equation}   
based on a past context sequence 
\begin{equation}\label{eq:ypast}
\ypast_{t,\clen}:=\{y_{t-\clen+1},y_{t-\clen+2}, \ldots, y_t\}
\end{equation}
for a given predictive window length $\plen\in\mathbb N$ and context window length $\clen\in\mathbb N$.
Consequently, a data-driven approach identifies a model, with parameters $\theta$, that best describes the conditional distribution (usually by minimizing a loss function) by taking windows of training data.

In this paper, we formulate the building forecasting problem as univariate forecasting each of the time-series signals. This is mainly to make the study fair as most state-of-the-art deep learning models, especially TSFMs, currently do not support multivariate forecasting. Additionally, empirical evidence suggests that the state-of-the-art multivariate approach, while theoretically consistent, often results in lower performance compared to univariate methods \citep{du2003univariate, woo2024unified}, although this may be debated.

\subsection{Time-Series Foundation Models}
\label{sec:fm}

Foundation models (FMs) \citep{bommasani2021opportunities} have recently emerged as a significant paradigm in artificial intelligence (AI), demonstrating state-of-the-art performance across a wide range of applications. The core concept of FMs involves using large datasets to \emph{pre-train} models with a very large number of parameters on generic self-supervised generative tasks (typically next-sample prediction), which can then be applied to downstream tasks with little (fine-tuning) to zero adaptation. Since FMs are compositions of a very large number of parameterized modules, they are flexible enough to reconstruct a varied landscape of functions. 
In this paper, we restrict our study to time-series foundation models, of which we consider three available at the time this paper is written: 
 \textsc{Moirai}, \textsc{TimesFM}, and \textsc{Chronos}.
\textsc{Moirai}, with its multimodal and hierarchical design, is reported to excel at capturing both short-term fluctuations and long-term trends, making it particularly adept at handling fragmented and high-variance inputs~\citep{woo2024unified}. 
\textsc{TimesFM}, on the other hand, emphasizes cross-task transfer learning by providing a flexible framework that seamlessly adapts its forecasting layers to new tasks without extensive re-training, thus boosting efficiency in real-world deployment scenarios~\citep{das2023decoder}. 
\textsc{Chronos} leverages a customized attention mechanism that natively handles irregular sampling and missing data, and is touted to perform well in tasks with incomplete contextual data~\citep{ansari2024Chronos}.
Despite some structural and algorithmic differences, the TSFM networks share many similarities: the foremost being that they are all probabilistic time-series forecasting models, trained with self-supervised generative task consistent with the foundation model paradigm. Furthermore, they are essentially built around transformer networks~\citep{vaswani2017attention}, and comprise over $10^8$ trainable parameters.
Additionally, they have all been made available on GitHub with weights pre-trained on broad collections of publicly available datasets covering various domains such as energy, transport, climate/weather, sales, economy, healthcare, and web data, often along with synthetic datasets, all of which have been considered at multiple time-scales. For these reasons, the authors of these works often refer to these TSFMs as `universal forecasters'. 
We opted not to discuss other time-series foundation models in this study primarily because \textsc{Moirai}, \textsc{TimesFM}, and \textsc{Chronos} collectively epitomize the predominant design innovations: hierarchical multi-modality, cross-task transfer capabilities, and adaptive attention-based architectures. Although there exist additional models that rely on variations of these strategies, they rarely depart substantially in either methodological underpinnings or empirical contributions. 

\emph{Note that the objective of this work is not to choose a `winner' between the TSFMs, but instead to understand applicability and challenges in adopting them for time-series forecasting in practical building energy applications.}

Due to their training on myriad datasets, TSFMs acquire the capability to autonomously identify temporal patterns such as periodicity and other generic yet distinctive trends within the pre-training data. This characteristic enables TSFMs to approximately generalize to unseen domains without task-specific fine-tuning, a feature known as \emph{zero-shot} inference. This means that zero adaptation/fine-tuning iterations are spent training the TSFM on task-specific data, and only the network's intrinsic adaptation mechanism, usually via attention layers, provides the conditional distribution $\pi_\theta(\cdot)$. The feasibility of zero-shot prediction differentiates FMs from classical deep learning models, offering significant convenience by eliminating the cumbersome process of data preparation and model retraining for each new task.

Unfortunately, zero-shot inference requires a sufficiently long context dataset to yield good predictive accuracy (we provide evidence later in \S\ref{sec:result_icl} of this paper), which may not always be available for reasons discussed in the introduction. In fact, allowing for long context lengths is contrary to the major premise of this paper, which is to be able to generate good predictions with limited data. Therefore, we focus our efforts on investigating efficient adaptation mechanisms with limited contextual data, which is discussed next.

\subsection{Fine-Tuning TSFMs for Building Energy Systems}
\label{sec:ft}
Fine-tuning is a post-training process that not only enables rapid adaptation to a specific task (i.e., signals from a particular target building energy system), but also leverages the synergy between the information embedded within the pre-trained model and the information contained in the downstream task-specific data. By fine-tuning the TSFM, we are effectively transfer-learning from a wide array of tasks (pre-training) to a specific one.

Formally, let us consider a pre-trained TSFM parameterized by the weights $\theta_0$, and a training ground-truth time-series $\ytrue_{0,T}:=\{y^{\sf true}_{1},y^{\sf true}_{2}, \ldots, y^{\sf true}_T\}$ collected from the target building under consideration. 
Given a context window length $\clen$ and a predictive window length $\plen$, we construct a fine-tuning dataset $\mathcal{D}_{\text{FT}}$ as follows:
\begin{equation}
	\mathcal{D}_{\text{FT}} := \left\{ \big( \ypasttrue_{i,\clen}, \ytrue_{i,\plen} \big) \mid i = \clen, \dots, T-\plen \right\}.
\end{equation}

\subsection*{Full fine-tuning (FullFT)}
Full fine-tuning involves re-training all the TSFM parameters by minimizing a loss function $\mathcal{L}$ suitable for the target building forecasting problem, using gradient descent methods such as Adam \citep{kingma2014adam} or its variants~\citep{loshchilov2017decoupled}. The optimization is performed over the fine-tuning dataset $\mathcal{D}_{\text{FT}}$ and we denote the resulting model parameters update by \begin{equation}\label{eq:fullft}
	\Delta \theta_{\sf{FullFT}} = \argmin_{\Delta \theta} \sum_t \mathcal{L} \big( \ytrue_{t,\plen}, \ynext_{t,\plen}; \theta _0 +\Delta \theta\big), 
\end{equation}
where $\ynext_{t,\plen}$ denotes statistics or sample predictions drawn from the probabilistic model, given a context input $\ypasttrue_{t,\clen}$.

The specific form of the loss function $\mathcal{L}$ depends on the architecture of the TSFM and the target task. For instance, \textsc{TimesFM} employs a weighted quantile loss, while \textsc{Chronos} discretizes the target values to reformulate the regression problem as a classification task, subsequently applying the categorical cross-entropy loss.
A major issue with solving~\eqref{eq:fullft} is that TSFMs typically contain hundreds of millions of parameters, and therefore the training procedure requires significant compute as well as care to avoid overfitting and, in extreme cases, catastrophic forgetting \citep{kirkpatrick2017catastrophic,min2022one}.

\noindent\subsection*{Parameter-Efficient Fine-Tuning (PEFT)}
To mitigate the issues of full fine-tuning, we adopt the common strategy of training on a lower-dimensional subspace of $\theta$. Since this requires training much fewer number of parameters, it is often referred to as parameter-efficient fine-tuning (PEFT). 

In particular, we employ low-rank adaptation (LoRA), which is one of the most prominent PEFT techniques at the time of writing~\citep{hu2021lora}. The concept of LoRA is simple: rather than updating all $\theta$, LoRA adapts on a low-rank subspace within $\theta$. A more mathematical description of the approach is provided below. 

Previously, we have assumed $\theta$ encapsulates all the weights in all the layers of the TSFM as a flattened tensor. Implementing LoRA involves first selecting a subset of layers, indexed by $\mathcal I$, whose weights will be adapted. Let $\omega_i$ denote a matrix of weights in the $i$-th layer of the TSFM, where $i\in\mathcal I$; the size of $\omega_i$ is $d_{i,1}\times d_{i,2}$. For each of the adapted layers, we define a low-rank adapter matrix $\delta\omega_i = \omega_{i,L} \omega_{i,R}$ where $\omega_{i,L}$ has size $d_{i,1}\times r_i$ and $\omega_{i,R}$ has size $r_i\times d_{i,2}$ for some prescribed rank $r_i\ll\min \{d_{i,1},d_{i,2}\}$. For simplicity of implementation, one often selects a uniform rank $r$ across all layers which satisfies $r\ll \min_{i\in\mathcal I} \{d_{i,1},d_{i,2}\}$. By doing this, the fine-tuning optimization problem can be recast as
\begin{equation}\label{eq:peft}
	\Delta \theta_{\sf{PEFT}} = \argmin_{\delta \omega_{\mathcal I}} \sum_t \mathcal{L} \big( \ytrue_{t,\plen}, \ynext_{t,\plen}; \theta_0 +\Delta \theta \big), 
\end{equation}
which results in training $\sum_{i\in\mathcal I}(d_{i,1}+d_{i,2})\times r$ weights. Since $r$ is quite small compared to the number of neurons per layer, this can be made significantly less than the original dimensionality $|\theta| = \sum_{i\in\mathcal I} d_{i,1}d_{i,2} + \sum_{i\not\in\mathcal I}d_{i,1}d_{i,2}$. Since the original parameter $\omega_i$ remains fixed throughout the PEFT stage, much fewer parameters are involved in back-propagation, resulting in significantly improved computational efficiency during training, making it feasible on resource-limited hardware on-site.
After solving~\eqref{eq:peft}, the learned adaptation matrices $\delta \omega_i$ are added into the original weight matrices, that is $\omega_i \leftarrow \omega_i + \delta\omega_i$. 
Furthermore, as noted in \citep{hu2021lora, zeng2023expressive}, the lower number of fine-tunable parameters means LoRA also protects against overfitting on small-sized target datasets.

In particular, for LoRA fine-tuning within the transformer components of the TSFM, the adaptation is as follows. The self-attention mechanism enables each token in an input sequence to dynamically attend to all other tokens by computing a weighted sum of value vectors based on the similarity between query and key vectors. Given input tokens $\tau$, the mechanism first projects $\tau$ into queries, keys, and values via $Q = \tau\omega_Q$, $K = \tau \omega_K$, and $V = \tau \omega_V$, where $\omega_Q, \omega_K, \omega_V$ are learnable weight matrices. The attention output is then computed as
\[
\textsf{Att}(Q, K, V) = \textsf{softmax}\left(\frac{QK^\top}{\sqrt{d_K}}\right)V,
\]
with $d_K$ being the dimensionality of the keys, and 
$\mathsf{softmax}(\mathbf{z}) = \tfrac{\exp(\mathbf{z})}{\mathbf{1}^\top \exp(\mathbf{z})}$, exponents taken component-wise. To adapt pre-trained models efficiently, LoRA fine-tuning is applied by injecting trainable low-rank matrices into the query and value weights. Concretely, instead of updating $\omega_Q$ and $\omega_V$ in their entirety, we update $\omega_Q := \omega_Q + \omega_{Q,L}\omega_{Q,R}$ and $\omega_V := \omega_V + \omega_{V,L}\omega_{V,R}$, with low-rank matrices $\omega_{Q,L}$, $\omega_{Q,R}$, $\omega_{V,L}$, and $\omega_{V,R}$. This approach retains the original pre-trained parameters while enabling task-specific adjustments with significantly fewer trainable parameters.

\section{Experimental Setup}
\label{sec:experiments}

\subsection{Data Collection}

For the purpose of training TSFMs, we use real experimental data collected from SUSTIE, which is a next-generation commercial office building located in Japan. The name SUSTIE combines the words ``Sustainability'' and ``Energy'' and the building is designed to research and demonstrate energy savings while insuring workers’ health and comfort. The four-story SUSTIE building has a total floor area of approximately 6456~m$^2$ which includes nine office spaces that are regularly used by around 260 office workers, an open-feel atrium area, a cafeteria, and a gym. SUSTIE's building management system collects data on electrical energy consumption, meteorological and indoor environment conditions, occupancy, and equipment operational data to analyze and control energy consumption and comfort during building operations.  The electrical energy consumption is measured separately for different types of equipment (air-conditioning, ventilation, lighting, hot water supply, and elevators) and for each room.  The occupancy, i.e., the number of people in each room, is counted by the access control system using card readers installed in each area. A dataset was collected at SUSTIE contiguously from January 2021 to August 2024, although only the most recent subset (i.e., September 2023 to August 2024) is used in this work. We adopted a number of data pre-processing steps as described below to make this dataset tractable and usable for training models.  Any missing values in data signals were filled using linear interpolation, and all data signals were synchronized to identical sampling times with a sampling rate of 15 minutes. The electrical energy consumption (kW$\cdot$h) signals of different equipment were also converted to power consumption (W) by assuming piece-wise constant power signals between two successive energy measurements. 

We consider four different types of occupant-centric time-series signals, collected at 15-minute intervals (i.e., 4 samples per hour) from SUSTIE during office workdays. We exclude weekend data as such signals are sometimes trivial to forecast and artificially boost predictive accuracy; e.g., zero occupancy and almost flat energy outputs. 
The signals are room occupancy (\texttt{Occ}), carbon emissions (\texttt{CO2}), power consumption for illumination and appliances (\texttt{Light}), and energy consumption for heating, ventilation, and air conditioning equipment (\texttt{HVAC}). 
To evaluate the efficacy of the proposed framework across seasons, we conducted the experiments using $4$ different test periods, each spanning the last 40 workdays of each season.
For each target variable (i.e., each signal at each split), the last $40$ office workdays of signals are hidden at training time and used as test periods.
Following the selection of 24-hr context windows in relevant literature~\citep{xing2024transfer,zheng2023interpretable}, the TSFM is set up to accept a 24-hr window as input\footnote{We focus on a 24-hr input context window as our main target scenario, but we examine longer windows in Section~\ref{sec:result_icl}.} and predict the next 6-hr.
This is also motivated by the fact that control algorithms such as model predictive control for building control often use 6--8-hr predictions with which to compute a control action~\citep{drgovna2020all,arroyo2022reinforced}.
Given that the data is collected at 15-minute intervals (i.e., 4 samples per hour), this corresponds to a context and a predictive window length of $\clen=24\times4=96$ and $\plen=6\times4=24$ steps, respectively.

Data was collected from eight zones spanning three floors within the building. The floors serve different primary functions: six zones on two floors (2F and 4F) are designated as office spaces, while the two zones on another floor (3F) are designated for relaxation purposes. 
As a consequence, we obtained a total of 32 different time series ($8$ zones $\times$ $4$ seasons) for each signal type.
By utilizing this diverse dataset, we aim to evaluate the proposed TSFM-based forecasting scheme’s ability to handle various operational conditions and environmental factors inherent in real-world building energy systems. This approach allows us to test the model’s robustness and generalization capabilities across different zones with varying functionalities and seasonal influences.

\subsection{Evaluation Metrics}

To assess the predictive performance of the proposed models, we consider two point forecast measures: mean absolute scaled error (MASE) and root-mean-squared scaled error (RMSSE) which are designed to understand how well the mean predictions are, as well as two distributional forecast accuracy metrics designed to understand how well the statistics of the time-series have been reconstructed: weighted quantile loss (wQL) and mean-scaled interval score (MSIS) based on prediction $\ynext_{t, \plen}$ defined in~\eqref{eq:ynext} and its corresponding ground truth
\[
\ytrue_{t,\plen} = \{y^{\sf true}_{t+1}, y^{\sf true}_{t+2}, \ldots, y^{\sf true}_{t+\plen}\},
\]
for a context window $\ypast_{t,\clen}$, defined in~\eqref{eq:ypast}. Suppose $\bar{y}_t$ denotes the (empirical) mean prediction for time-series value $y_t$ at time $t$. We define these evaluation metrics next.

The MASE is calculated as
\begin{equation*}
	\mathsf{MASE}(\ynext_{t, \plen}, \ytrue_{t,\plen}) \triangleq \frac{1}{\zeta_{\sf MAE}}\cdot\frac{1}{\plen} \sum_{\tau=t+1}^{t+\plen} |\bar{y}_\tau - y^{\sf true}_\tau|
\end{equation*}
and the RMSSE is computed as
\begin{equation*}
	\mathsf{RMSSE}(\ynext_{t, \plen}, \ytrue_{t,\plen}) \triangleq \frac{1}{\zeta_{\sf RMSE}}\cdot\sqrt{\frac{1}{\plen} \sum_{\tau=t+1}^{t+\plen} |\bar{y}_\tau - y^{\sf true}_\tau|^2},
\end{equation*}
where
\begin{align}\label{eq:zeta_mae}
	\zeta_{\sf MAE} &\triangleq \frac{1}{\tlen-\rlen} \sum_{\tau=\rlen+1}^{\tlen} |y_\tau^{\sf true} - y^{\sf true}_{\tau-\rlen}|, 
\text{ and }	\zeta_{\sf RMSE} \triangleq \sqrt{\frac{1}{\tlen-\rlen} \sum_{\tau=\rlen+1}^{\tlen} |y_\tau^{\sf true} - y^{\sf true}_{\tau-\rlen}|^2}	
\end{align}
are scaling factors derived from the naive forecasting errors over the training period. As shown, these scaling factors are calculated using the mean absolute error (MAE) between the ground truth series $\ytrue_{0,T}$ and its 24-hour (i.e., $\rlen = 24 \times 4 = 96$ steps, given the 15-minute sampling interval) lagged signal.
By scaling the error measures in this manner, we assess the predictive performance relative to a naive baseline and present the results as percentages for clearer interpretation. This approach allows for a standardized comparison across different models and datasets.

It is important to note that we apply a constant scaling factor averaged over the entire training series rather than using varying scales for different sub-series based on the test time $t$. This approach ensures that the metrics do not disproportionately emphasize sub-series that are either easy to forecast or have near-zero values, which may hold less practical significance.

To evaluate the quality of distributional learning, we consider two distributional forecast accuracy metrics: wQL and MSIS. These metrics assess not only the accuracy of the point forecasts but also the quality of the uncertainty estimates provided by the predictive distributions.
Suppose $y^{(\beta)}_t$ denotes the $\beta$-th quantile, for time-series value $y_t$ at time $t$. Note that $\beta\in (0,1)$ and $\beta=0.1$ indicates the 10-th quantile.
The wQL assesses the accuracy of quantile forecasts by averaging the quantile losses over pre-selected quantiles, which can be written as
\begin{equation*}
	\mathsf{wQL}(\pi_\theta, \ytrue_{t,\plen}) \triangleq \frac{1}{N_q} \sum_{n=1}^{N_q}  \mathsf{QL}^{(\beta_n)}(\pi_\theta, \ytrue_{t,\plen}),
\end{equation*}
where the quantile loss at the $\beta_n$-th quantile is given by
\begin{align*}
\mathsf{QL}^{(\beta_n)}(\pi_\theta, \ytrue_{t,\plen}) \triangleq \frac{2}{\plen\zeta_{\sf{QL}}} \sum_{\tau=t+1}^{t+\plen} \left[ \beta_n \cdot \max\left(0,\; y_t^{\sf true} - y_{t}^{(\beta_n)}\right) + (1 - \beta_n) \cdot \max\left(0,\; y_{t}^{(\beta_n)} - y_t^{\sf true}\right) \right]
\end{align*}
with the scaling factor
\[
\zeta_{\sf{QL}} \triangleq \frac{1}{\tlen-\rlen} \sum_{\tau=\rlen+1}^{\tlen} |y^{\sf true}_\tau|.
\]
In this paper, we evaluate the quantile losses at the $10$-th, $50$-th, and $90$-th quantiles; therefore $\beta_1=0.1, \beta_2=0.5, \beta_3=0.9$ and $N_q=3$. This reflects a lower confidence bound, median, and upper confidence bound of the predictive distribution, respectively. A smaller value of wQL indicates better probabilistic forecasting performance.

For MSIS, some additional notation is required.
Suppose $u_t = y^{(1-\beta)}_t$ and $l_t=y^{(\beta)}_t$ respectively denote the upper and lower confidence intervals, where $\beta\in (0, 0.5)$. Let $\mathsf I(\cdot)$ denote the indicator function that is $1$ when the conditional argument is true, and $0$ if false.
The MSIS evaluates the quality of the prediction intervals by balancing penalties on the width of the interval with penalties on observations falling outside the interval, and can be written as
\begin{align*}
	\mathsf{MSIS}(\pi_\theta, \ytrue_{t,\plen}) &\triangleq \frac{1}{\plen\zeta_{\sf MAE}}  \sum_{\tau=t+1}^{t+\plen} \Bigl[ (u_t - l_t)  + \frac{2}{\beta} (l_t - y^{\sf true}_t) \cdot \mathsf{I}(y_t^{\sf true} < l_t)  + \frac{2}{\beta} (y_t^{\sf true} - u_t) \cdot \mathsf{I}(y_t^{\sf true} > u_t) \Bigr],
\end{align*}
with $\zeta_{\sf MAE}$ defined in~\eqref{eq:zeta_mae}. In this paper, we select $\beta=0.1$.

Then, MSIS rewards the competing objectives of finding a prediction interval as narrow as possible (i.e., showing as little uncertainty as possible) while minimizing the number of true values found to fall outside of it (i.e., accurately capturing the observed estimate uncertainty).
A lower MSIS indicates better probabilistic forecasting performance.

Given that the test period spans $\ttlen = 40~\text{days} \times 24~\text{hr/day} \;\times\; 4~\text{steps/hr} = 3840$ steps, which is longer than the predictive window length $H = 24$ steps, we employ a rolling-window analysis as suggested in~\citep{zivot2006modeling}.
Specifically, for each time step $t$ from $T$ to $T+\ttlen-\plen$, we generate forecasts using the model and compute with a given evaluation metric. We then report the average scores over all these test time steps to assess the model's overall predictive performance.
\subsection{Comparison Study with State-of-the-Art}

We consider three TSFM architectures: \textsc{Moirai}, \textsc{TimesFM}, and \textsc{Chronos}. These models utilize transformers as their underlying architecture, with 91M, 200M, and 200M parameters, respectively. For zero-shot predictions, we use the pre-trained weights provided by their respective official implementations\footnote{\textsc{Moirai}: \texttt{github.com/redoules/moirai, huggingface.co/Salesforce/moirai-1.0-R-base}\\\textsc{TimesFM}: \texttt{github.com/google-research/timesfm, huggingface.co/google/timesfm-1.0-200m}\\ \textsc{Chronos}: \texttt{github.com/amazon-science/chronos-forecasting, huggingface.co/amazon/chronos-t5-base}}.

We evaluate our proposed time-series foundation models against three prominent deep learning baselines, each implemented via the GluonTS library \citep{gluonts_jmlr}. The first, \textsc{DeepAR} \citep{salinas2020deepar}, employs an LSTM-based architecture to generate probabilistic forecasts by learning global models across multiple time series, making it well-suited to heterogeneous datasets. The second, \textsc{N-BEATS} \citep{oreshkin2019n}, is a purely feedforward architecture organized into backward and forward residual stacks that excel at capturing long-term temporal patterns; it is inherently deterministic, and thus we do not report its distributional accuracy. Lastly, temporal fusion transformers (\textsc{TFT}) \citep{lim2021temporal} combine attention mechanisms and a gating methodology to handle temporal dynamics and feature interactions, offering both interpretability and strong predictive performance across diverse forecasting tasks.

\section{Results and Discussions}
\label{sec:results}
\newcommand{\yj}[1]{{\color{purple}#1}}

We begin our empirical analysis by evaluating the effectiveness of applying TSFMs both in a zero-shot setting and with fine-tuning. To provide a comprehensive comparison, we also report the forecasting performance metrics of benchmark deep learning models trained from scratch. In Sections~\ref{sec:result_comparison}, \ref{sec:result_icl}, and \ref{sec:result_ft}, we focus primarily on the overall performance of the framework; therefore, we present the mean $\pm$ standard deviation of errors across all 32 different time series, encompassing 8 zones over 4 seasons.
For each signal, the time-series data from the 3 months immediately preceding each test period is used for training or fine-tuning, with context and predictive window lengths fixed as 24 and 6 hours, respectively, unless otherwise stated.
Detailed analyses for each zone and season are discussed in subsequent Section~\ref{sec:result_specific}. 
In addition, we investigate the robustness and effectiveness of the proposed fine-tuning for TSFMs with limited data in Sections~\ref{sec:result_robust}.
Lastly, the generalization capability of the TSFM approach to unseen zones is examined in Section~\ref{sec:result_unseen}.

\begin{table*}[ht!]
	\centering
	\normalsize
	\caption{
		Comparison of forecasting performance.
		The winner/runner-up is highlighted in light blue/yellow, respectively.
		Fine-tuned TSFMs outperform baseline TSFMs and state-of-the-art deep forecasting models.
		The average scores along with their standard deviations across different zones and seasons are presented together.}
	\label{tab:comparison}
	\resizebox{1.0\linewidth}{!}{
		\begin{tabular}{ll|rrr|rrr|rr}
			\toprule
			&  & \multicolumn{3}{c}{\textbf{Deep Forecasting Models}} & \multicolumn{3}{c}{\textbf{TSFMs (Zero-Shot)}} & \multicolumn{2}{c}{\textbf{Fine-tuned TSFMs}} \\
			\cmidrule(lr){3-5} \cmidrule(lr){6-8} \cmidrule(lr){9-10}
			\textbf{Signal} & \textbf{Metric} & \textsc{N-BEATS} & \textsc{DeepAR} & \textsc{TFT} & \textsc{Moirai} & \textsc{TimesFM} & \textsc{Chronos} & \makecell[r]{\textsc{Chronos}\\+FullFT} & \makecell[r]{\textsc{Chronos}\\+PEFT}  \\
			\midrule
			\midrule
			\multirow{4}{*}{\texttt{Occ}}
			& MASE & 0.25 {\scriptsize $\pm$ 0.04} & 0.21 {\scriptsize $\pm$ 0.03} & 0.20 {\scriptsize $\pm$ 0.04} & 0.55 {\scriptsize $\pm$ 0.08} & 0.41 {\scriptsize $\pm$ 0.06} & 0.41 {\scriptsize $\pm$ 0.05} & \cellcolor{bestgreen} 0.18 {\scriptsize $\pm$ 0.03} & \cellcolor{bestblue} 0.18 {\scriptsize $\pm$ 0.03} \\
			& RMSSE & 0.26 {\scriptsize $\pm$ 0.04} & 0.22 {\scriptsize $\pm$ 0.03} & 0.21 {\scriptsize $\pm$ 0.04} & 0.57 {\scriptsize $\pm$ 0.08} & 0.42 {\scriptsize $\pm$ 0.06} & 0.41 {\scriptsize $\pm$ 0.05} & \cellcolor{bestgreen} 0.20 {\scriptsize $\pm$ 0.03} & \cellcolor{bestblue} 0.19 {\scriptsize $\pm$ 0.03} \\
			& MSIS & 5.09 {\scriptsize $\pm$ 0.75} & 3.49 {\scriptsize $\pm$ 0.66} & \cellcolor{bestgreen} 2.83 {\scriptsize $\pm$ 0.80} & 6.24 {\scriptsize $\pm$ 1.06} & 8.80 {\scriptsize $\pm$ 1.30} & 5.58 {\scriptsize $\pm$ 0.87} & 2.91 {\scriptsize $\pm$ 0.54} & \cellcolor{bestblue} 2.71 {\scriptsize $\pm$ 0.64} \\
			& wQL & 0.64 {\scriptsize $\pm$ 0.09} & 0.37 {\scriptsize $\pm$ 0.05} & \cellcolor{bestgreen} 0.34 {\scriptsize $\pm$ 0.08} & 0.87 {\scriptsize $\pm$ 0.15} & 0.94 {\scriptsize $\pm$ 0.15} & 0.77 {\scriptsize $\pm$ 0.11} & 0.35 {\scriptsize $\pm$ 0.05} & \cellcolor{bestblue} 0.32 {\scriptsize $\pm$ 0.05} \\
			\midrule
			\multirow{4}{*}{\texttt{CO2}}
			& MASE & 0.02 {\scriptsize $\pm$ 0.01} & 0.04 {\scriptsize $\pm$ 0.02} & 0.02 {\scriptsize $\pm$ 0.01} & 0.04 {\scriptsize $\pm$ 0.01} & 0.04 {\scriptsize $\pm$ 0.01} & 0.03 {\scriptsize $\pm$ 0.01} & \cellcolor{bestgreen} 0.02 {\scriptsize $\pm$ 0.00} & \cellcolor{bestblue} 0.02 {\scriptsize $\pm$ 0.00} \\
			& RMSSE & 0.03 {\scriptsize $\pm$ 0.01} & 0.04 {\scriptsize $\pm$ 0.02} & 0.03 {\scriptsize $\pm$ 0.01} & 0.05 {\scriptsize $\pm$ 0.01} & 0.04 {\scriptsize $\pm$ 0.01} & 0.04 {\scriptsize $\pm$ 0.01} & \cellcolor{bestgreen} 0.03 {\scriptsize $\pm$ 0.01} & \cellcolor{bestblue} 0.02 {\scriptsize $\pm$ 0.01} \\
			& MSIS & 0.47 {\scriptsize $\pm$ 0.10} & 0.66 {\scriptsize $\pm$ 0.35} & \cellcolor{bestgreen} 0.37 {\scriptsize $\pm$ 0.09} & 0.56 {\scriptsize $\pm$ 0.14} & 0.84 {\scriptsize $\pm$ 0.25} & 0.44 {\scriptsize $\pm$ 0.11} & 0.37 {\scriptsize $\pm$ 0.08} & \cellcolor{bestblue} 0.30 {\scriptsize $\pm$ 0.08} \\
			& wQL & 0.05 {\scriptsize $\pm$ 0.01} & 0.06 {\scriptsize $\pm$ 0.03} & \cellcolor{bestgreen} 0.04 {\scriptsize $\pm$ 0.01} & 0.06 {\scriptsize $\pm$ 0.02} & 0.07 {\scriptsize $\pm$ 0.02} & 0.05 {\scriptsize $\pm$ 0.01} & 0.04 {\scriptsize $\pm$ 0.01} & \cellcolor{bestblue} 0.03 {\scriptsize $\pm$ 0.01} \\
			\midrule
			\multirow{4}{*}{\texttt{Light}}
			& MASE & 0.36 {\scriptsize $\pm$ 0.06} & 0.18 {\scriptsize $\pm$ 0.02} & 0.17 {\scriptsize $\pm$ 0.05} & 0.50 {\scriptsize $\pm$ 0.07} & 0.33 {\scriptsize $\pm$ 0.07} & 0.32 {\scriptsize $\pm$ 0.05} & \cellcolor{bestblue} 0.15 {\scriptsize $\pm$ 0.03} & \cellcolor{bestgreen} 0.15 {\scriptsize $\pm$ 0.03} \\
			& RMSSE & 0.40 {\scriptsize $\pm$ 0.06} & 0.21 {\scriptsize $\pm$ 0.02} & 0.20 {\scriptsize $\pm$ 0.04} & 0.52 {\scriptsize $\pm$ 0.07} & 0.34 {\scriptsize $\pm$ 0.07} & 0.34 {\scriptsize $\pm$ 0.04} & \cellcolor{bestgreen} 0.18 {\scriptsize $\pm$ 0.03} & \cellcolor{bestblue} 0.18 {\scriptsize $\pm$ 0.03} \\
			& MSIS & 7.29 {\scriptsize $\pm$ 1.24} & 2.97 {\scriptsize $\pm$ 0.56} & 2.72 {\scriptsize $\pm$ 1.18} & 6.32 {\scriptsize $\pm$ 0.72} & 6.91 {\scriptsize $\pm$ 1.53} & 4.70 {\scriptsize $\pm$ 0.60} & \cellcolor{bestgreen} 2.32 {\scriptsize $\pm$ 0.52} & \cellcolor{bestblue} 2.12 {\scriptsize $\pm$ 0.56} \\
			& wQL & 0.75 {\scriptsize $\pm$ 0.12} & 0.26 {\scriptsize $\pm$ 0.04} & 0.25 {\scriptsize $\pm$ 0.08} & 0.66 {\scriptsize $\pm$ 0.08} & 0.59 {\scriptsize $\pm$ 0.14} & 0.48 {\scriptsize $\pm$ 0.07} & \cellcolor{bestgreen} 0.23 {\scriptsize $\pm$ 0.04} & \cellcolor{bestblue} 0.21 {\scriptsize $\pm$ 0.04} \\
			\midrule
			\multirow{4}{*}{\texttt{HVAC}}
			& MASE & 0.28 {\scriptsize $\pm$ 0.23} & 0.25 {\scriptsize $\pm$ 0.20} & 0.24 {\scriptsize $\pm$ 0.18} & 0.43 {\scriptsize $\pm$ 0.34} & 0.31 {\scriptsize $\pm$ 0.28} & 0.27 {\scriptsize $\pm$ 0.21} & \cellcolor{bestblue} 0.20 {\scriptsize $\pm$ 0.14} & \cellcolor{bestgreen} 0.21 {\scriptsize $\pm$ 0.14} \\
			& RMSSE & 0.29 {\scriptsize $\pm$ 0.22} & 0.26 {\scriptsize $\pm$ 0.20} & 0.24 {\scriptsize $\pm$ 0.17} & 0.46 {\scriptsize $\pm$ 0.39} & 0.30 {\scriptsize $\pm$ 0.25} & 0.27 {\scriptsize $\pm$ 0.20} & \cellcolor{bestblue} 0.21 {\scriptsize $\pm$ 0.14} & \cellcolor{bestgreen} 0.22 {\scriptsize $\pm$ 0.14} \\
			& MSIS & 5.54 {\scriptsize $\pm$ 4.53} & 4.83 {\scriptsize $\pm$ 4.18} & 3.74 {\scriptsize $\pm$ 3.20} & 4.53 {\scriptsize $\pm$ 3.97} & 6.72 {\scriptsize $\pm$ 6.04} & 4.51 {\scriptsize $\pm$ 3.69} & \cellcolor{bestblue} 3.35 {\scriptsize $\pm$ 2.65} & \cellcolor{bestgreen} 3.37 {\scriptsize $\pm$ 2.64} \\
			& wQL & 0.83 {\scriptsize $\pm$ 0.69} & 0.60 {\scriptsize $\pm$ 0.50} & 0.48 {\scriptsize $\pm$ 0.36} & 0.72 {\scriptsize $\pm$ 0.63} & 0.85 {\scriptsize $\pm$ 0.77} & 0.66 {\scriptsize $\pm$ 0.54} & \cellcolor{bestgreen} 0.47 {\scriptsize $\pm$ 0.35} & \cellcolor{bestblue} 0.46 {\scriptsize $\pm$ 0.33} \\
			\bottomrule
		\end{tabular}
	}
\end{table*}

\subsection{Comparing predictions of TSFMs and Competitors}
\label{sec:result_comparison}
Table~\ref{tab:comparison} reports the results of predictive performance of the state-of-the-art forecasting models N-BEATS, DeepAR, and TFT, along with the zero-shot TSFMs and the fine-tuned \textsc{Chronos}, which exhibit the best zero-shot performance. It is  especially interesting that the zero-shot performance of the TSFMs is similar to, and sometimes better than the non-transformer-based deep forecasting models that have explicitly been trained on the task-specific context data. For instance, both \textsc{Chronos} and \textsc{Moirai} perform similar on the \texttt{CO2}, \texttt{Light}, and \texttt{HVAC} tasks in terms of the weighted quantile loss and the point estimate error metrics, with \textsc{Chronos} doing better overall than \textsc{Moirai}. It is worth noting that the signals collected from the building energy systems are not completely unrelated to the TSFM pre-training data, which includes datasets such as \texttt{SpanishEnergyAndWeather}, and \texttt{AustralianElectricity}. This explains why the distribution learning in the zero-shot paradigm is meaningful. A potential reason for \textsc{Chronos}'s improved performance, in line with current conventional wisdom regarding generative models for continuous data (e.g., images), is that models based on tokenized representation often outperform models with continuous variables~\citep{vanoord2017vqvae}. In our case, we see \textsc{Chronos}, which operates on a tokenized representation of time series data and is trained via a classification cross-entropy loss, performs better than \textsc{TimesFM}, i.e., a fully-connected network trained via a regression loss. Encouraged by \textsc{Chronos}' zero-shot performance, we also report the fine-tuned \textsc{Chronos} performance in the final two columns of the table; fine-tuning involved 1000 re-training iterations with FullFT (taking 1.4~s/iter) and PEFT (taking 0.6~s/iter). We infer that fine-tuned \textsc{Chronos} outperforms all the deep forecasting models, including TFT, in most cases, and that PEFT helps more than FullFT in most cases. Even when PEFT is slightly worse, such as for HVAC energy prediction, the difference is marginal, given that PEFT requires 2x less GPU time than FullFT to re-train.

Overall, the zero-shot performance of TSFMs with a 24-hr context window falls short compared to models specifically trained on the downstream datasets. As discussed in the following Section~\ref{sec:result_icl}, this performance gap is largely due to the short context window length. Although building signals exhibit daily periodicity, a context window of 24 hours is insufficient for the zero-shot inference scheme, as the models are unable to fully identify (daily) periodic patterns by only looking back at the past 24 hours.

\subsection{Leveraging Longer Context for Improved Zero-Shot Forecasts}
\label{sec:result_icl}

\begin{figure}[!ht]
	\centering
	\begin{subfigure}[]{0.45\textwidth}
		\centering
		\includegraphics[width=\linewidth]{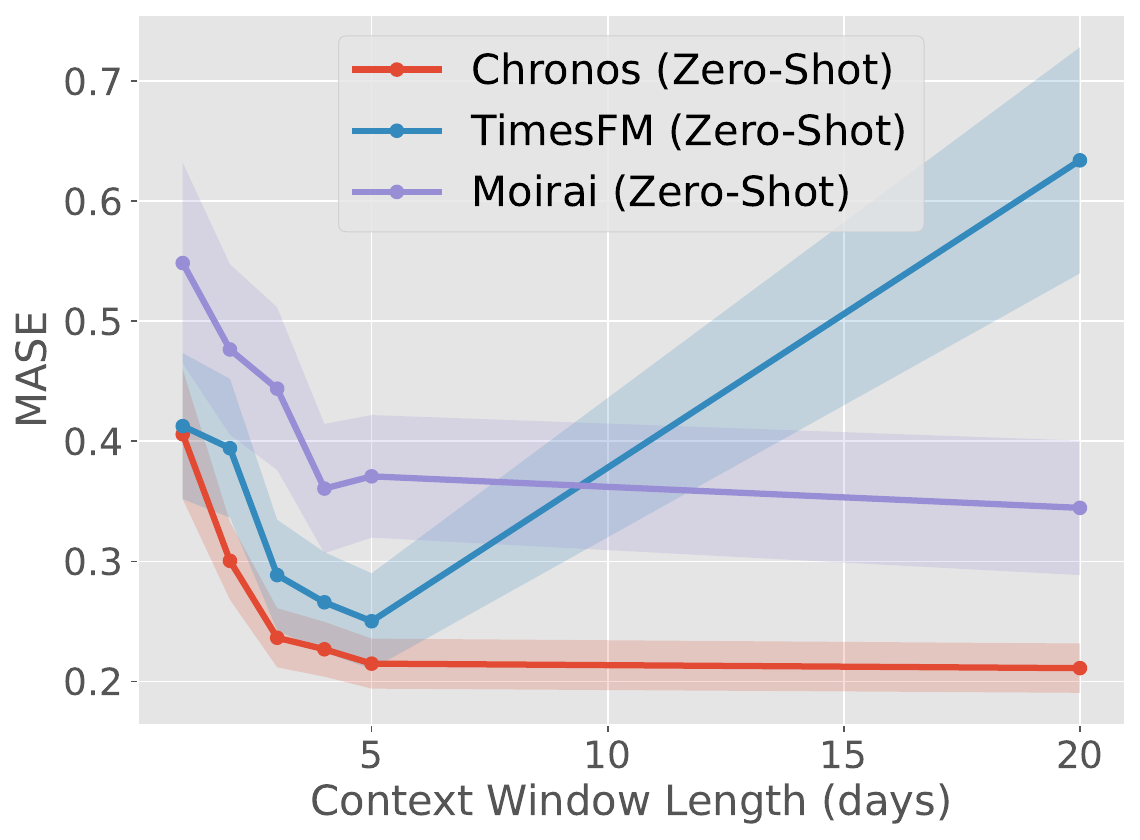}
	\end{subfigure}
	\begin{subfigure}[]{0.45\textwidth}
		\centering
		\includegraphics[width=\linewidth]{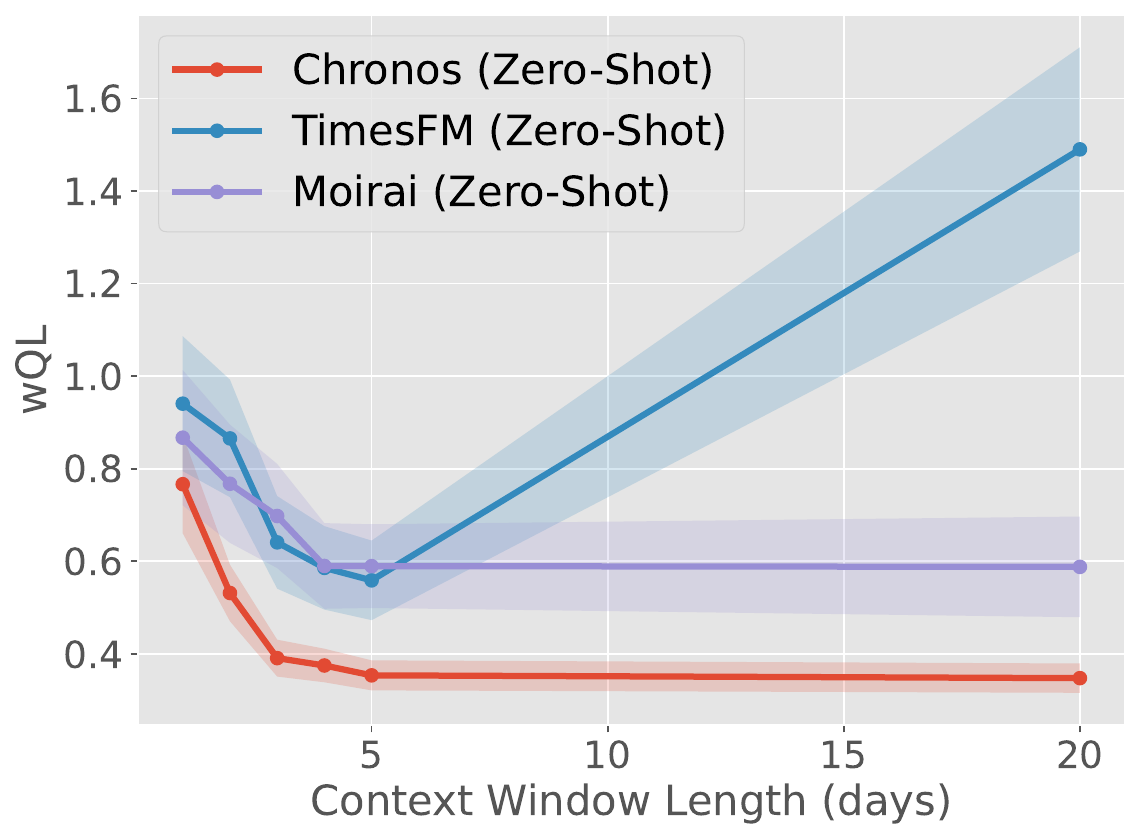}
	\end{subfigure}
	\caption{Forecasting accuracies of the zero-shot in-context inference with TSFM models across different context window sizes on the Occupancy (\texttt{Occ}) signal. The shades represent the standard deviation across different zones and seasons.}
	\vspace{-1em}
	\label{fig:icl-occ}
\end{figure}

One of the defining features of pre-trained TSFMs is their innate ability to uncover periodic temporal patterns when provided with sufficiently long context data, without requiring an additional training stage.
Here, we report how longer contexts 
can lead to substantial improvement of the TSFM zero-shot forecasts.
Specifically, we experiment with varying context window lengths: $\clen \in \{1, 2, 3, 4, 5, 20\}$ office workdays, where each workday comprises 96 time steps and we focus on the occupancy channel (other channels exhibit similar trends). Longer context windows were infeasible due to GPU memory constraints, as they would involve input sequences spanning thousands of time steps. Our experiments are conducted on all three of the TSFMs.

Figure~\ref{fig:icl-occ} shows that to achieve high-performance zero-shot predictions, the model requires a sufficiently long context. Specifically, when the context length is too short ($\le 48$ hours), the performance of TSFMs is substantially degraded, i.e., results in high forecasting errors. As expected, accuracy improves with increasing context window length up to 5 workdays (i.e., 480 time steps). This is justified by the fact that commercial building signals are highly periodic with respect to the work-day as well as the work-week, and providing a work-week (5 days) worth of context reveals trends unseen over a single 24hr period. 
Choosing a very long context window is also not a good method for chasing performance. As evident from the figure, over 5 days results in diminishing returns for two of the TSFMs. Moreover, we find that when \textsc{TimesFM} is provided with a 20-day context window (1,920 time steps), its prediction error is alarmingly elevated. This is not unreasonable, given that the model was optimized for a maximum context length of 512 time steps during the pre-training stage. This is another key factor in selecting the context window length: one needs to choose a context window that is within the maximum length chosen during pre-training. We also recognize that the computational complexity of transformer-based TSFMs scales quadratically with the context length $\mathcal{O}(\clen^2)$ \citep{dao2022flashattention}, making test-time inference increasingly expensive, even with engineering improvements such as flash attention and key-value caching.
To reiterate, there may be circumstances when the pre-trained TSFM is too cumbersome to fine-tune, e.g., the necessary data or hardware is not available. In such cases, one way to deploy a TSFM directly without fine-tuning to achieve good predictive performance is by carefully selecting the appropriate length of the context window, which, for our commercial building, is between 3 to 5 days.

\subsection{Effectiveness of (Parameter-Efficient) Fine-Tuning}
\label{sec:result_ft}
\begin{table}[!ht]
	\centering
	\caption{
		Comparison of forecasting accuracy between zero-shot, PEFT with varying LoRA ranks, and FullFT.
	}
	\label{tab:lora_ablation}
	\resizebox{0.6\linewidth}{!}{
		\begin{tabular}{ll|r|rrrr}
			\toprule
			\cmidrule(lr){3-6} \cmidrule(lr){7-7}
			\textbf{Signal} & \textbf{Metric} & Zero-Shot & \makecell[r]{LoRA \\ (rank=4)} & \makecell[r]{LoRA \\ (rank=16)} & \makecell[r]{LoRA \\ (rank=64)} & \makecell[r]{FullFT} \\
			\midrule
			\midrule
			\multirow{4}{*}{\texttt{Occ}}
			& MASE & 0.41 {\scriptsize $\pm$ 0.05} & \cellcolor{bestgreen} 0.18 {\scriptsize $\pm$ 0.03} & \cellcolor{bestblue} 0.18 {\scriptsize $\pm$ 0.03} & 0.18 {\scriptsize $\pm$ 0.03} & 0.18 {\scriptsize $\pm$ 0.03} \\
			& RMSSE & 0.41 {\scriptsize $\pm$ 0.05} & \cellcolor{bestgreen} 0.19 {\scriptsize $\pm$ 0.03} & \cellcolor{bestblue} 0.19 {\scriptsize $\pm$ 0.03} & 0.19 {\scriptsize $\pm$ 0.03} & 0.20 {\scriptsize $\pm$ 0.03} \\
			& MSIS & 5.58 {\scriptsize $\pm$ 0.87} &   \cellcolor{bestblue} 2.71 {\scriptsize $\pm$ 0.64} & \cellcolor{bestgreen} 2.72 {\scriptsize $\pm$ 0.65} & 2.72 {\scriptsize $\pm$ 0.65} & 2.91 {\scriptsize $\pm$ 0.54} \\
			& wQL & 0.77 {\scriptsize $\pm$ 0.11} & \cellcolor{bestblue} 0.32 {\scriptsize $\pm$ 0.05} & \cellcolor{bestgreen} 0.32 {\scriptsize $\pm$ 0.05} & 0.32 {\scriptsize $\pm$ 0.05} & 0.35 {\scriptsize $\pm$ 0.05} \\
			\midrule
			\multirow{4}{*}{\texttt{CO2}}
			& MASE & 0.03 {\scriptsize $\pm$ 0.01} & 0.02 {\scriptsize $\pm$ 0.00} & \cellcolor{bestblue} 0.02 {\scriptsize $\pm$ 0.00} & \cellcolor{bestgreen} 0.02 {\scriptsize $\pm$ 0.00} & 0.02 {\scriptsize $\pm$ 0.00} \\
			& RMSSE & 0.04 {\scriptsize $\pm$ 0.01} & 0.02 {\scriptsize $\pm$ 0.01} & \cellcolor{bestgreen} 0.02 {\scriptsize $\pm$ 0.01} & \cellcolor{bestblue} 0.02 {\scriptsize $\pm$ 0.01} & 0.03 {\scriptsize $\pm$ 0.01} \\
			& MSIS & 0.44 {\scriptsize $\pm$ 0.11} & 0.30 {\scriptsize $\pm$ 0.08} & \cellcolor{bestgreen} 0.30 {\scriptsize $\pm$ 0.08} & \cellcolor{bestblue} 0.30 {\scriptsize $\pm$ 0.08} & 0.37 {\scriptsize $\pm$ 0.08} \\
			& wQL & 0.05 {\scriptsize $\pm$ 0.01} & \cellcolor{bestgreen} 0.03 {\scriptsize $\pm$ 0.01} & \cellcolor{bestblue} 0.03 {\scriptsize $\pm$ 0.01} & 0.03 {\scriptsize $\pm$ 0.01} & 0.04 {\scriptsize $\pm$ 0.01} \\
			\midrule
			\multirow{4}{*}{\texttt{Light}}
			& MASE & 0.32 {\scriptsize $\pm$ 0.05} & 0.15 {\scriptsize $\pm$ 0.03} & \cellcolor{bestblue} 0.15 {\scriptsize $\pm$ 0.03} & 0.15 {\scriptsize $\pm$ 0.03} & \cellcolor{bestgreen} 0.15 {\scriptsize $\pm$ 0.03} \\
			& RMSSE & 0.34 {\scriptsize $\pm$ 0.04} &  \cellcolor{bestgreen} 0.18 {\scriptsize $\pm$ 0.03} & \cellcolor{bestblue} 0.18 {\scriptsize $\pm$ 0.03} & 0.18 {\scriptsize $\pm$ 0.03} & 0.18 {\scriptsize $\pm$ 0.03} \\
			& MSIS & 4.70 {\scriptsize $\pm$ 0.60} & \cellcolor{bestblue} 2.12 {\scriptsize $\pm$ 0.56} & 2.13 {\scriptsize $\pm$ 0.58} & \cellcolor{bestgreen} 2.13 {\scriptsize $\pm$ 0.57} & 2.32 {\scriptsize $\pm$ 0.52} \\
			& wQL & 0.48 {\scriptsize $\pm$ 0.07}  & \cellcolor{bestblue} 0.21 {\scriptsize $\pm$ 0.04} & \cellcolor{bestgreen} 0.21 {\scriptsize $\pm$ 0.04} & 0.21 {\scriptsize $\pm$ 0.04} & 0.23 {\scriptsize $\pm$ 0.04} \\
			\midrule
			\multirow{4}{*}{\texttt{HVAC}}
			& MASE & 0.27 {\scriptsize $\pm$ 0.21} & 0.21 {\scriptsize $\pm$ 0.14} & \cellcolor{bestgreen} 0.21 {\scriptsize $\pm$ 0.14} & 0.21 {\scriptsize $\pm$ 0.14} & \cellcolor{bestblue} 0.20 {\scriptsize $\pm$ 0.14} \\
			& RMSSE & 0.27 {\scriptsize $\pm$ 0.20} & 0.22 {\scriptsize $\pm$ 0.14} & \cellcolor{bestgreen} 0.21 {\scriptsize $\pm$ 0.14} & 0.21 {\scriptsize $\pm$ 0.14} & \cellcolor{bestblue} 0.21 {\scriptsize $\pm$ 0.14} \\
			& MSIS & 4.51 {\scriptsize $\pm$ 3.69} & 3.37 {\scriptsize $\pm$ 2.64} & \cellcolor{bestblue} 3.33 {\scriptsize $\pm$ 2.62} & \cellcolor{bestgreen} 3.34 {\scriptsize $\pm$ 2.60} & 3.35 {\scriptsize $\pm$ 2.65} \\
			& wQL & 0.66 {\scriptsize $\pm$ 0.54} & 0.46 {\scriptsize $\pm$ 0.33} & \cellcolor{bestblue} 0.46 {\scriptsize $\pm$ 0.33} & \cellcolor{bestgreen} 0.46 {\scriptsize $\pm$ 0.33} & 0.47 {\scriptsize $\pm$ 0.35} \\
			\bottomrule
		\end{tabular}
	}
\end{table}

Of course, the ideal situation is if one can fine-tune the TSFM to the target system. Here, we demonstrate that we can further improve the forecasting performance of TSFMs via fine-tuning without relying on a computationally expensive long context window. Concretely, we test the effects of FullFT as well as PEFT via LoRA, the latter of which has shown promise in fine-tuning large language models (LLMs) but has been under-explored for TSFMs.
Given that \textsc{Chronos} demonstrates the most promising preliminary zero-shot performance and the lack of (parameter efficient) fine-tuning support or HuggingFace~\citep{wolf-etal-2020-transformers} compatibility in other TSFMs, we focus our efforts on \textsc{Chronos}.

As mentioned before, in Table~\ref{tab:comparison}, we observe that fine-tuned~\textsc{Chronos} clearly and consistently outperforms the benchmark deep forecasting models with both FullFT and PEFT, reducing some of the zero-shot error metrics by more than 50\%.
To better understand the impact of LoRA rank $r$, we vary $r=4$, $16$, and $64$ for 1,000 fine-tuning iterations and report performance in Table~\ref{tab:lora_ablation}.
We did not test larger ranks, as they only yielded marginal performance improvements while incurring higher computational costs.
We see that LoRA mostly outperforms FullFT except for point-estimates on HVAC energy consumption. This is consistent with the observations of LoRA in language tasks~\citep{hu2021lora}, and mainly because LoRA structurally maintains a balance between preserving prior knowledge from the pre-trained model and adapting to task-specific data, while mitigating the risk of overfitting. 

Since the PEFT results are all similarly good, we recommend a lower rank (e.g., $r=4$ or $r=16$) to reduce the fine-tuning computational expense for limited data and few FT iterations. In fact, using LoRA $r=4$ reduces our total training floating-point operations (FLOPS) by 33\% and accelerates training by 2.3$\times$ compared to FullFT, on an Nvidia RTX 2080Ti GPU and 6 CPU cores.

\subsection{Prediction Qualities across Different Zones and Seasons}
\label{sec:result_specific}

\begin{figure}[!ht]
	\centering
	\begin{subfigure}[]{0.45\textwidth}
		\centering
		\includegraphics[width=\linewidth]{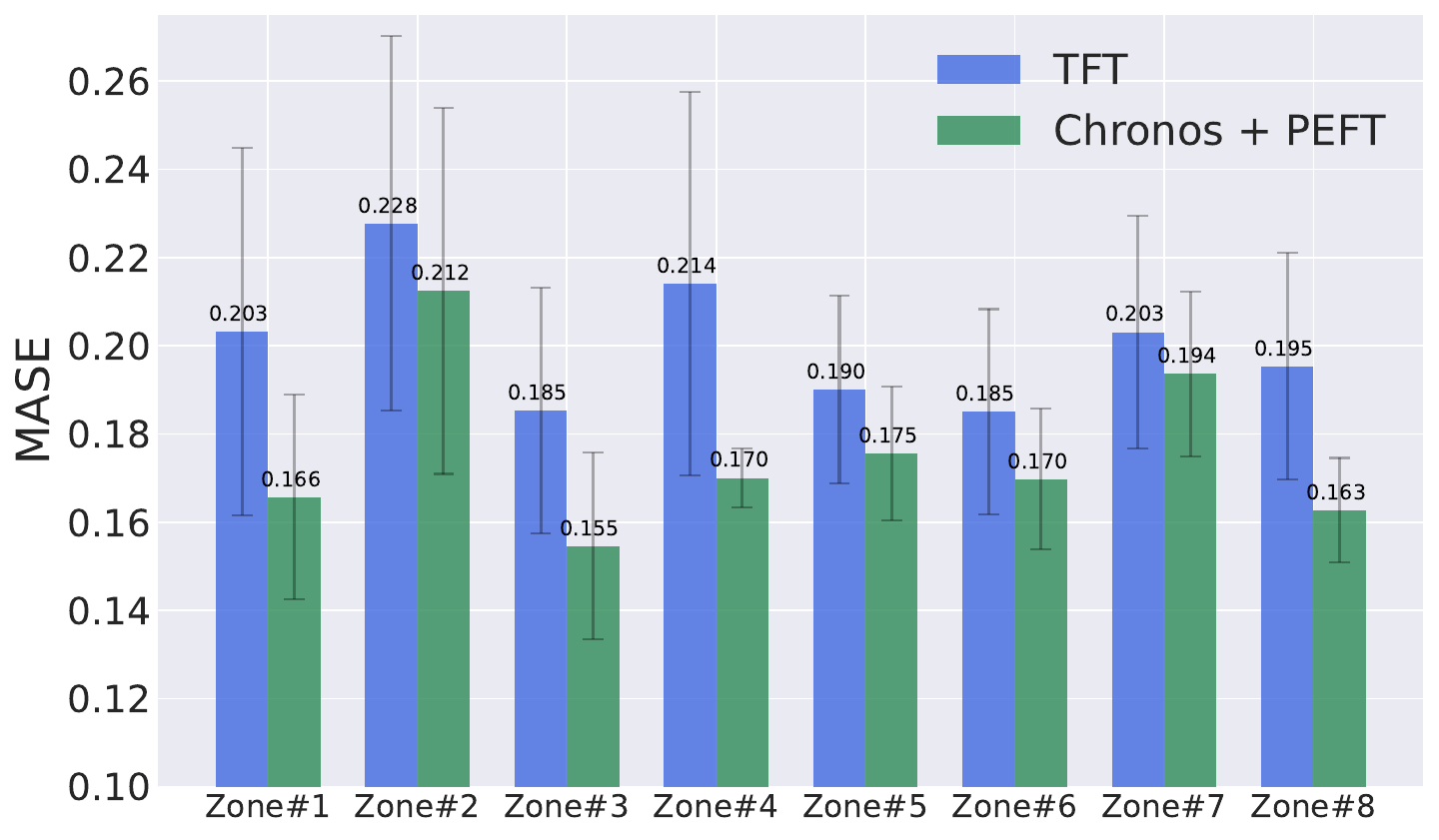}
	\end{subfigure}
	\begin{subfigure}[]{0.45\textwidth}
		\centering
		\includegraphics[width=\linewidth]{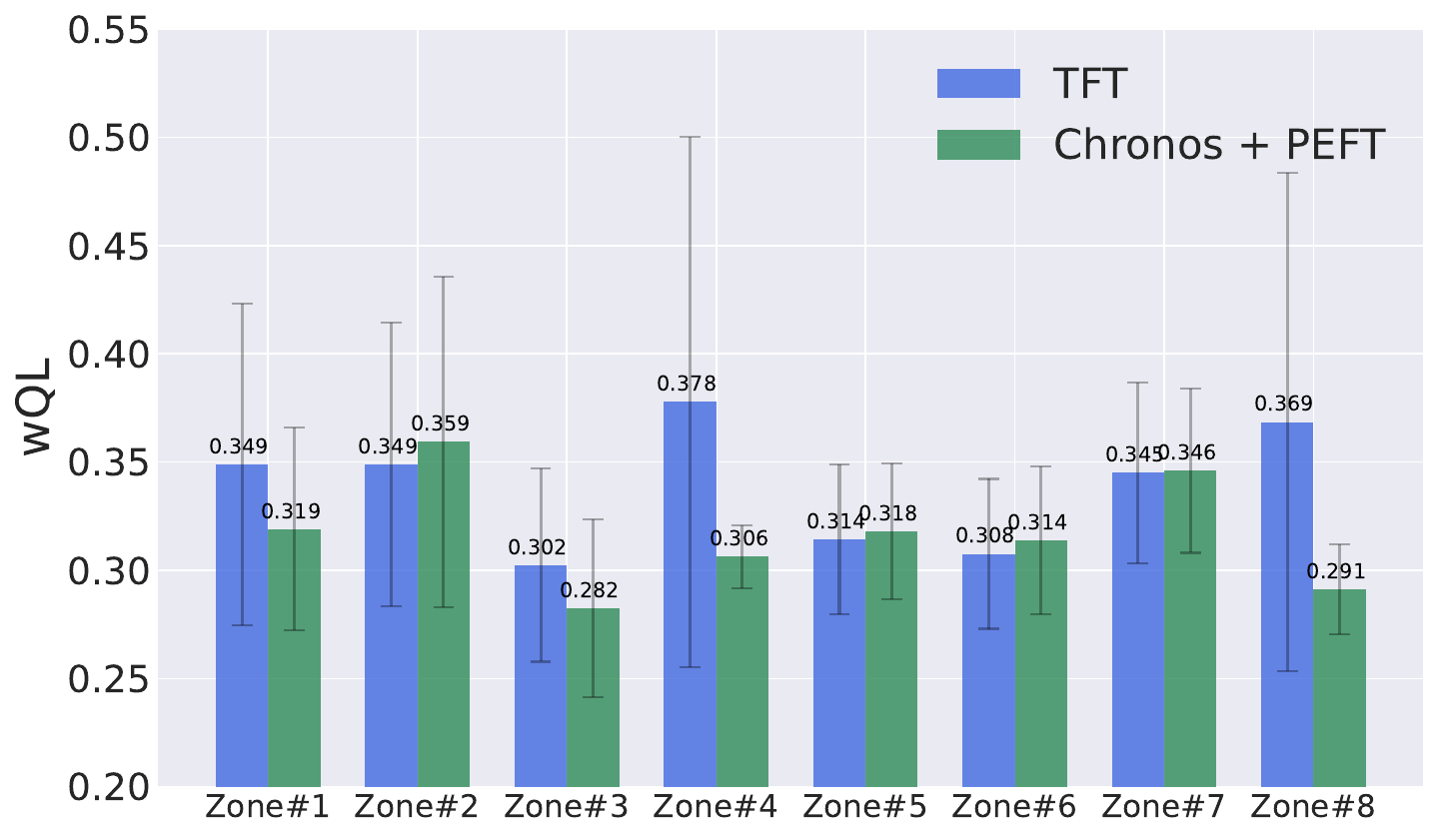}
	\end{subfigure}
	\caption{Forecasting accuracy of the benchmark TFT model and the proposed TSFM approach using PEFT across different zones on the Occupancy (\texttt{Occ}) signal. The error bars represent the standard deviation across different seasons at each zone.}
	\vspace{-1em}
	\label{fig:zone-occ}
\end{figure}

\begin{figure}[!ht]
	\centering
	\begin{subfigure}[]{0.45\textwidth}
		\centering
		\includegraphics[width=\linewidth]{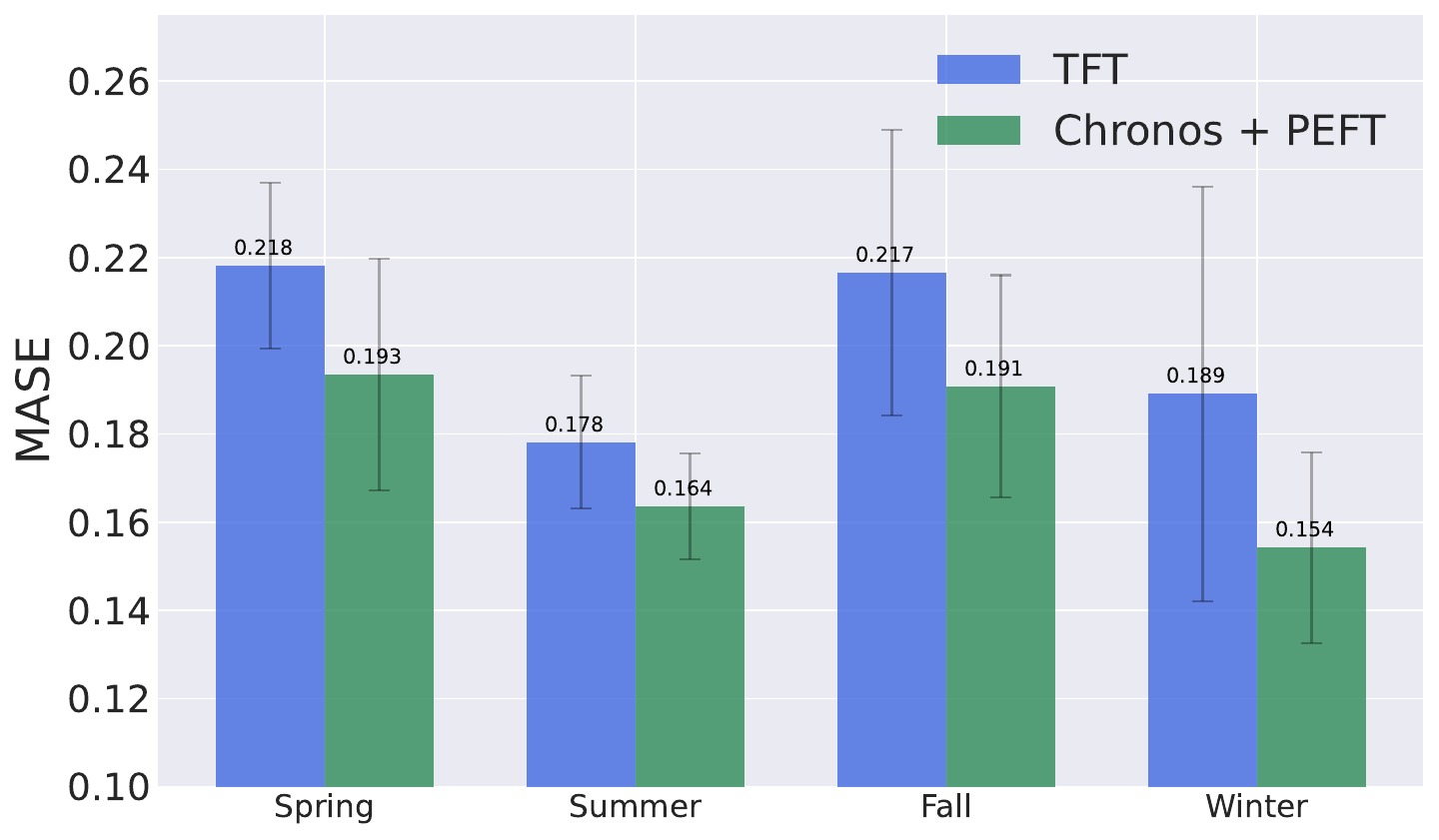}
	\end{subfigure}
	\begin{subfigure}[]{0.45\textwidth}
		\centering
		\includegraphics[width=\linewidth]{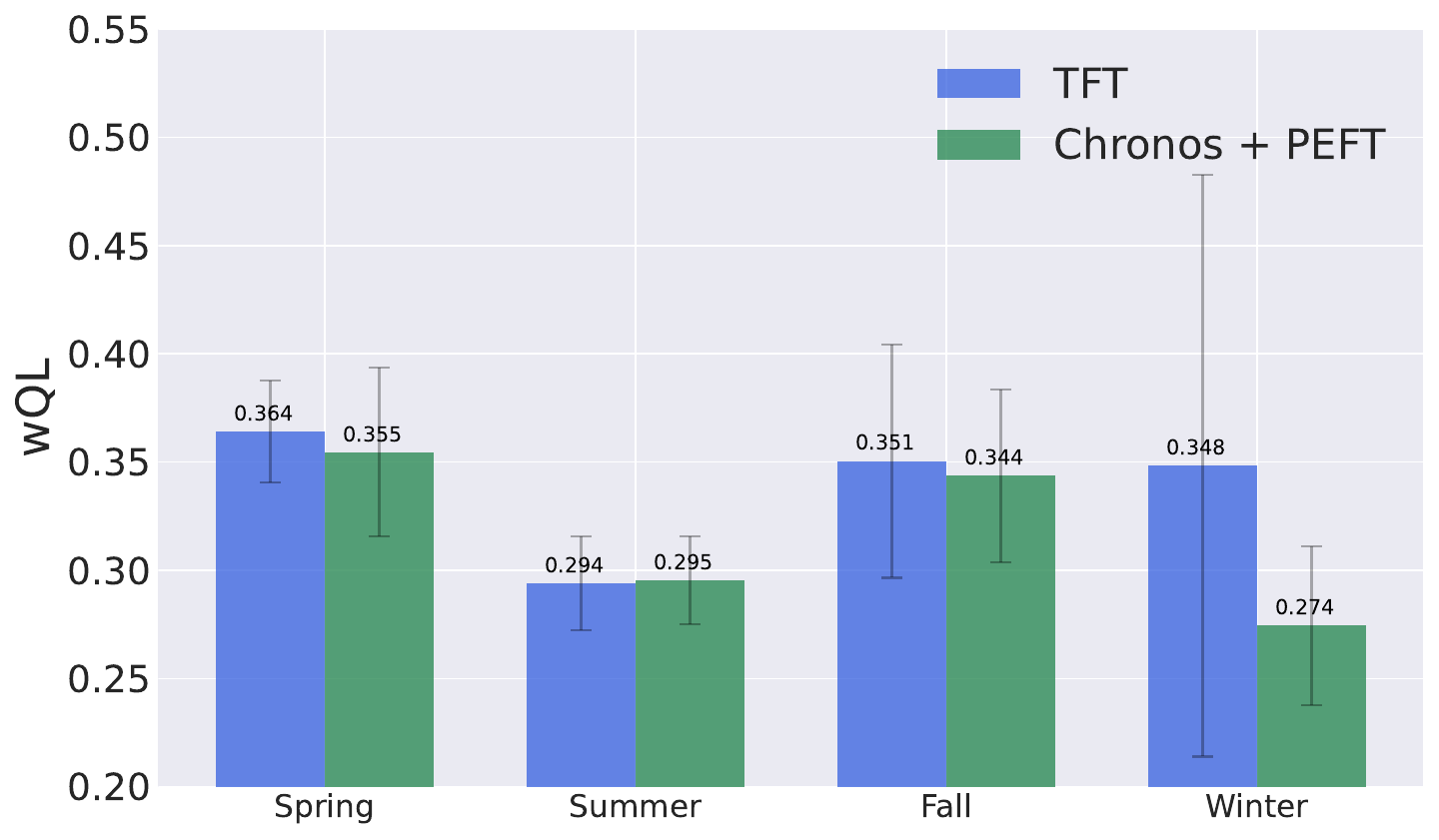}
	\end{subfigure}
	\caption{Forecasting accuracy of the benchmark TFT model and the proposed TSFM approach using PEFT across different seasons on the Occupancy (\texttt{Occ}) signal. The error bars represent the standard deviation across different zones during each season.}
	\vspace{-1em}
	\label{fig:season-occ}
\end{figure}

Beyond evaluating the overall performance, we further compare the prediction accuracies of the fine-tuned TSFM via PEFT against the TFT, which demonstrates the highest performance among deep forecasting models in Table~\ref{tab:comparison}. We present the \texttt{Occ} signal as a representative example in Figure~\ref{fig:zone-occ} and \ref{fig:season-occ} for ease of exposition, although the other signals are provided in Table~\ref{tab:comparison_zone_1_8} in the Appendix.

Notably, the proposed TSFM approach consistently outperforms the TFT, demonstrating up to approximately 25\% improvement in both deterministic and probabilistic metrics (i.e., MASE and wQL). Although the TFT once surpasses the TSFM in the wQL metric (e.g., during summer), the performance gap is only $0.001$ in the mean. In contrast to the TFT, the TSFM approach consistently exhibits low error across various situations. More importantly, the TSFM approach demonstrates consistently smaller standard deviations across different seasons and zones, whereas the TFT often exhibits inconsistent performance. This empirical analysis underscores the robustness of the TSFM approach in its predictive capabilities, making it more reliable for application in BEMS with varying environmental conditions and seasonality.

\subsection{Robustness of TSFM-based Approach to Limited Data}
\label{sec:result_robust}

\begin{figure}[!ht]
    \centering
    \begin{subfigure}[]{0.45\textwidth}
        \centering
        \includegraphics[width=\linewidth]{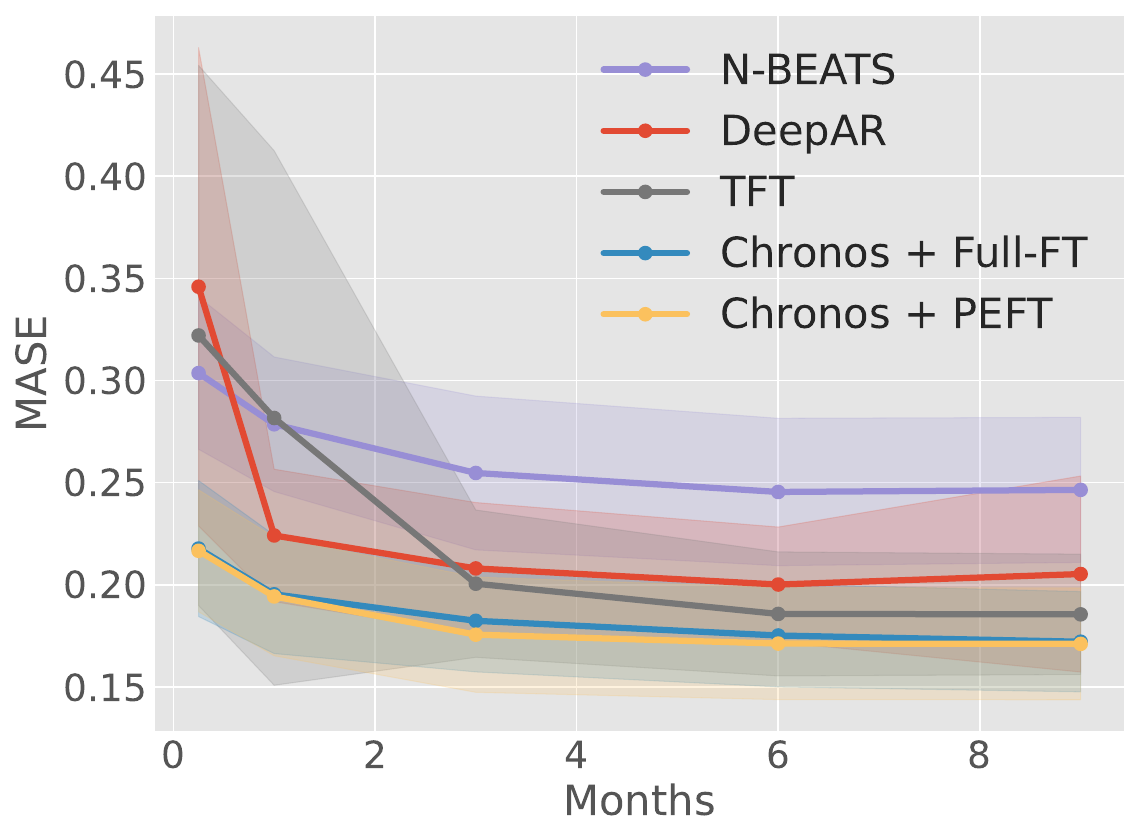}
    \end{subfigure}
    \begin{subfigure}[]{0.45\textwidth}
        \centering
        \includegraphics[width=\linewidth]{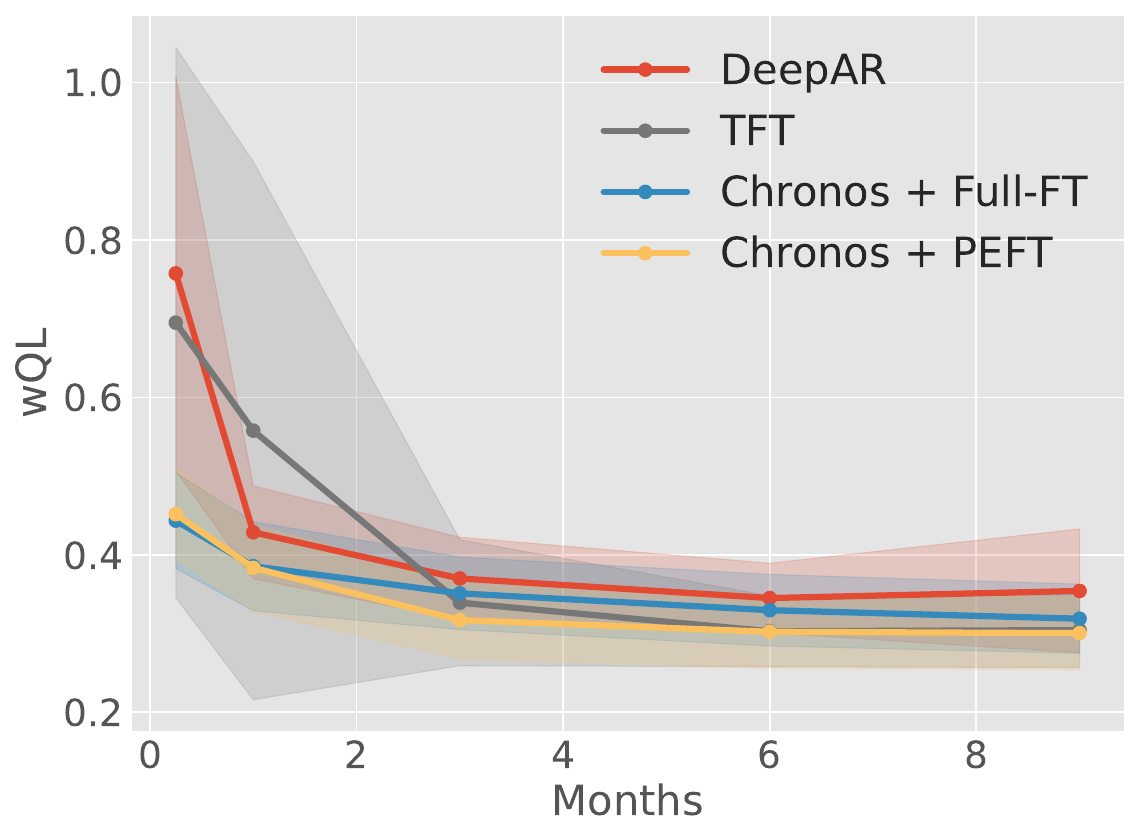}
    \end{subfigure}
    \caption{Forecasting accuracies of the benchmark models and the proposed TSFM approach using PEFT across different training dataset size on the Occupancy (\texttt{Occ}) signal. The shades represent the standard deviation across different zones and seasons.}%
    \vspace{-1em}
    \label{fig:abl-occ}
\end{figure}

Thus far, we have demonstrated that TSFMs can successfully forecast real-world building time-series data with fine-tuning.
An immediate question is how much time-series data is needed to fine-tune such large models, and whether a fine-tuned model can achieve high accuracy with less data.
To explore this, we conduct an ablation study by varying the training dataset size, using periods of $\{1, 4\}$ weeks and $\{3, 6, 9\}$ months.
For a fair comparison, we keep the test period (i.e., split) fixed while extending the (preceding) training duration accordingly.
The plot of forecasting accuracy across different models for the \texttt{Occ} signal is shown in Figure~\ref{fig:abl-occ}, with the results for the other signals provided in Figure~\ref{fig:abl-occ-full}, \ref{fig:abl-co2-full}, \ref{fig:abl-light-full}, and \ref{fig:abl-hvac-full} in the Appendix.
When a large quantity of training data is available, \textsc{TFT} demonstrates performance comparable to that of TSFMs, indicating that TSFMs are most critically useful in the limited data scenario, e.g., less than 6 months. Indeed, the pre-training stage of TSFMs enables models to leverage general patterns found in large diverse pre-training datasets, which then helps to both mitigate overfitting and enhance generalization in their fine-tuned versions~\citep{devlin2018bert}. This is also evidenced by the fact that when the available data is severely limited, below 2 months, the TSFM performance degrades more gracefully than TFT or DeepAR.

\begin{table*}[ht!]
	\centering
	\caption{
		Comparison of forecasting accuracy between TFT and TSFMs on \emph{unseen} zone. The model is trained on a zone on the 2nd floor, then tested on another zone on the 3rd floor. \smallskip
	}
	\label{tab:unseen_1}
	\resizebox{1.0\linewidth}{!}{
		\begin{tabular}{llccc|llccc}
			\toprule
			&  & & \multicolumn{2}{c|}{\textbf{Fine-tuned TSFMs}} &  & & & \multicolumn{2}{c}{\textbf{Fine-tuned TSFMs}} \\
			\cmidrule(lr){4-5}
			\cmidrule(lr){9-10}
			\textbf{Signal} & \textbf{Metric} & \textsc{TFT} & \makecell[r]{\textsc{Chronos}\\+FullFT} & \makecell[r]{\textsc{Chronos}\\+PEFT} & \textbf{Signal} & \textbf{Metric} & \textsc{TFT} & \makecell[r]{\textsc{Chronos}\\+FullFT} & \makecell[r]{\textsc{Chronos}\\+PEFT}  \\
			\midrule
			\midrule
			\multirow{4}{*}{\texttt{Occ}}  & MASE  & 0.271 { \scriptsize $\pm$ 0.071 }  & \cellcolor{bestgreen} 0.231 { \scriptsize $\pm$ 0.061 }  & \cellcolor{bestblue} 0.229 { \scriptsize $\pm$ 0.064 } & 
			\multirow{4}{*}{\texttt{Light}}  & MASE  & 0.262 { \scriptsize $\pm$ 0.061 }  & \cellcolor{bestblue} 0.216 { \scriptsize $\pm$ 0.056 }  & \cellcolor{bestgreen} 0.223 { \scriptsize $\pm$ 0.049 } \\
			& RMSSE  & 0.278 { \scriptsize $\pm$ 0.070 }  & \cellcolor{bestgreen} 0.248 { \scriptsize $\pm$ 0.062 }  & \cellcolor{bestblue} 0.243 { \scriptsize $\pm$ 0.063 } & 
			& RMSSE  & 0.292 { \scriptsize $\pm$ 0.056 }  & \cellcolor{bestblue} 0.255 { \scriptsize $\pm$ 0.057 }  & \cellcolor{bestgreen} 0.258 { \scriptsize $\pm$ 0.050 } \\
			& MSIS  & 4.040 { \scriptsize $\pm$ 1.030 }  & \cellcolor{bestgreen} 3.693 { \scriptsize $\pm$ 1.179 }  & \cellcolor{bestblue} 3.629 { \scriptsize $\pm$ 1.173 } & 
			& MSIS  & 3.818 { \scriptsize $\pm$ 0.936 }  & \cellcolor{bestgreen} 3.368 { \scriptsize $\pm$ 1.045 }  & \cellcolor{bestblue} 3.239 { \scriptsize $\pm$ 0.744 } \\
			& wQL  & \cellcolor{bestgreen} 0.432 { \scriptsize $\pm$ 0.110 }  & 0.435 { \scriptsize $\pm$ 0.114 }  & \cellcolor{bestblue} 0.416 { \scriptsize $\pm$ 0.117 } &
			& wQL  & 0.305 { \scriptsize $\pm$ 0.062 }  & \cellcolor{bestgreen} 0.281 { \scriptsize $\pm$ 0.076 }  & \cellcolor{bestblue} 0.259 { \scriptsize $\pm$ 0.059 } \\
			\midrule
			\multirow{4}{*}{\texttt{CO2}}  & MASE  & 0.345 { \scriptsize $\pm$ 0.087 }  & \cellcolor{bestgreen} 0.274 { \scriptsize $\pm$ 0.074 }  & \cellcolor{bestblue} 0.255 { \scriptsize $\pm$ 0.070 } & 
			\multirow{4}{*}{\texttt{HVAC}}  & MASE  & 0.436 { \scriptsize $\pm$ 0.310 }  & \cellcolor{bestgreen} 0.410 { \scriptsize $\pm$ 0.235 }  & \cellcolor{bestblue} 0.383 { \scriptsize $\pm$ 0.270 } \\
			& RMSSE  & 0.309 { \scriptsize $\pm$ 0.074 }  & \cellcolor{bestgreen} 0.255 { \scriptsize $\pm$ 0.068 }  & \cellcolor{bestblue} 0.239 { \scriptsize $\pm$ 0.064 } & 
			& RMSSE  & 0.380 { \scriptsize $\pm$ 0.292 }  & \cellcolor{bestgreen} 0.363 { \scriptsize $\pm$ 0.263 }  & \cellcolor{bestblue} 0.347 { \scriptsize $\pm$ 0.276 } \\
			& MSIS  & 4.875 { \scriptsize $\pm$ 1.384 }  & \cellcolor{bestgreen} 4.742 { \scriptsize $\pm$ 1.230 }  & \cellcolor{bestblue} 3.624 { \scriptsize $\pm$ 1.009 } &
			& MSIS  & 6.895 { \scriptsize $\pm$ 3.724 }  & \cellcolor{bestgreen} 6.462 { \scriptsize $\pm$ 4.056 }  & \cellcolor{bestblue} 6.285 { \scriptsize $\pm$ 4.443 } \\
			& wQL  & \cellcolor{bestgreen} 0.100 { \scriptsize $\pm$ 0.024 }  & 0.101 { \scriptsize $\pm$ 0.023 }  & \cellcolor{bestblue} 0.079 { \scriptsize $\pm$ 0.018 } & 
			& wQL  & \cellcolor{bestgreen} 0.984 { \scriptsize $\pm$ 0.494 }  & 0.988 { \scriptsize $\pm$ 0.427 }  & \cellcolor{bestblue} 0.924 { \scriptsize $\pm$ 0.438 } \\
			\bottomrule
		\end{tabular}
	}
\end{table*}

\begin{table*}[ht!]
	\centering
	\caption{
		Comparison of forecasting accuracy between TFT and TSFMs on \emph{unseen} zone. The model is trained on a zone on the 2nd floor, then tested on another zone on the 4th floor. 
	}
	\label{tab:unseen_2}
	\resizebox{1.0\linewidth}{!}{
		\begin{tabular}{llccc|llccc}
			\toprule
			&  & & \multicolumn{2}{c|}{\textbf{Fine-tuned TSFMs}} &  & & & \multicolumn{2}{c}{\textbf{Fine-tuned TSFMs}} \\
			\cmidrule(lr){4-5}
			\cmidrule(lr){9-10}
			\textbf{Dataset} & \textbf{Metric} & \textsc{TFT} & \makecell[r]{\textsc{Chronos}\\+FullFT} & \makecell[r]{\textsc{Chronos}\\+PEFT} & \textbf{Dataset} & \textbf{Metric} & \textsc{TFT} & \makecell[r]{\textsc{Chronos}\\+FullFT} & \makecell[r]{\textsc{Chronos}\\+PEFT}  \\
			\midrule
			\midrule
			\multirow{4}{*}{\texttt{Occ}}  & MASE  & 0.250 { \scriptsize $\pm$ 0.077 }  & \cellcolor{bestgreen} 0.224 { \scriptsize $\pm$ 0.061 }  & \cellcolor{bestblue} 0.213 { \scriptsize $\pm$ 0.064 } & 
			\multirow{4}{*}{\texttt{Light}}  & MASE  & 0.225 { \scriptsize $\pm$ 0.054 }  & \cellcolor{bestgreen} 0.223 { \scriptsize $\pm$ 0.054 }  & \cellcolor{bestblue} 0.218 { \scriptsize $\pm$ 0.048 } \\
			& RMSSE  & 0.260 { \scriptsize $\pm$ 0.073 }  & \cellcolor{bestgreen} 0.242 { \scriptsize $\pm$ 0.062 }  & \cellcolor{bestblue} 0.229 { \scriptsize $\pm$ 0.063 } & 
			& RMSSE  & \cellcolor{bestgreen} 0.256 { \scriptsize $\pm$ 0.050 }  & 0.261 { \scriptsize $\pm$ 0.055 }  & \cellcolor{bestblue} 0.249 { \scriptsize $\pm$ 0.049 } \\
			& MSIS  & 3.799 { \scriptsize $\pm$ 1.144 }  & \cellcolor{bestgreen} 3.417 { \scriptsize $\pm$ 1.241 }  & \cellcolor{bestblue} 3.117 { \scriptsize $\pm$ 1.230 } & 
			& MSIS  & 3.495 { \scriptsize $\pm$ 0.804 }  & \cellcolor{bestgreen} 3.423 { \scriptsize $\pm$ 0.981 }  & \cellcolor{bestblue} 3.063 { \scriptsize $\pm$ 0.707 } \\
			& wQL  & 0.413 { \scriptsize $\pm$ 0.112 }  & \cellcolor{bestgreen} 0.409 { \scriptsize $\pm$ 0.114 }  & \cellcolor{bestblue} 0.366 { \scriptsize $\pm$ 0.117 } &
			& wQL  & \cellcolor{bestblue} 0.256 { \scriptsize $\pm$ 0.055 }  & 0.299 { \scriptsize $\pm$ 0.074 }  & \cellcolor{bestgreen} 0.261 { \scriptsize $\pm$ 0.059 } \\
			\midrule
			\multirow{4}{*}{\texttt{CO2}}  & MASE  & 0.348 { \scriptsize $\pm$ 0.081 }  & \cellcolor{bestgreen} 0.294 { \scriptsize $\pm$ 0.063 }  & \cellcolor{bestblue} 0.285 { \scriptsize $\pm$ 0.057 } & 
			\multirow{4}{*}{\texttt{HVAC}}  & MASE  & 0.262 { \scriptsize $\pm$ 0.288 }  & \cellcolor{bestblue} 0.216 { \scriptsize $\pm$ 0.211 }  & \cellcolor{bestgreen} 0.231 { \scriptsize $\pm$ 0.254 } \\
			& RMSSE  & 0.309 { \scriptsize $\pm$ 0.070 }  & \cellcolor{bestgreen} 0.264 { \scriptsize $\pm$ 0.059 }  & \cellcolor{bestblue} 0.259 { \scriptsize $\pm$ 0.054 } & 
			& RMSSE  & 0.256 { \scriptsize $\pm$ 0.273 }  & \cellcolor{bestgreen} 0.227 { \scriptsize $\pm$ 0.242 }  & \cellcolor{bestblue} 0.225 { \scriptsize $\pm$ 0.262 } \\
			& MSIS  & 5.314 { \scriptsize $\pm$ 1.177 }  & \cellcolor{bestgreen} 5.129 { \scriptsize $\pm$ 1.056 }  & \cellcolor{bestblue} 4.169 { \scriptsize $\pm$ 0.679 } &
			& MSIS  & 3.553 { \scriptsize $\pm$ 3.418 }  & \cellcolor{bestgreen} 3.398 { \scriptsize $\pm$ 3.711 }  & \cellcolor{bestblue} 2.928 { \scriptsize $\pm$ 4.215 } \\
			& wQL  & \cellcolor{bestgreen} 0.079 { \scriptsize $\pm$ 0.027 }  & 0.081 { \scriptsize $\pm$ 0.025 }  & \cellcolor{bestblue} 0.065 { \scriptsize $\pm$ 0.018 } & 
			& wQL  & 0.486 { \scriptsize $\pm$ 0.450 }  & \cellcolor{bestgreen} 0.453 { \scriptsize $\pm$ 0.373 }  & \cellcolor{bestblue} 0.421 { \scriptsize $\pm$ 0.402 } \\
			\bottomrule
		\end{tabular}
	}
\end{table*}

\subsection{Generalization to Unseen Zones} \label{sec:result_unseen}
In practical situations, one is often faced with predicting time-series for a new client/user: i.e., with severely limited prior data available for training. Data may also not be available for all areas requiring forecasting, due to cost or privacy concerns. In such circumstances, we evaluate the generalizability of TSFMs to unseen (but known to be similar) thermal zones. We fine-tuned \textsc{Chronos} using data for 3 months from 
a zone on the 2nd floor of our commercial building, and then tested it on the one from the 3rd (Table~\ref{tab:unseen_1}) and 4th (Table~\ref{tab:unseen_2}) without any additional tuning. Note that these floors serve completely different purposes (one floor is for relaxation, one has computer equipment, and one floor is a standard office layout).
As in the previous sections, we conducted tests across different seasons.

We observe that TSFM with fine-tuning outperforms TFT predictions comprehensively in terms of point estimate errors. And, as in previous sections, PEFT often does slightly better than FullFT. The improvement of fine-tuned TSFMs over TFT is especially apparent from the \texttt{Light} and \texttt{HVAC} categories, where the TSFM competitors generalize significantly better, possibly due to an informative prior learned from a pre-training dataset containing energy data (albeit from different source in a completely different context). For \texttt{Occ} and \texttt{CO2} (which are correlated), since usage patterns across zones are not very different, TFT is not as far behind in the probabilistic metrics, but remains outclassed by fine-tuned TSFMs. This demonstrates clearly how the pre-training prior of \textsc{Chronos} helps regularize the training and build predictions more robust to unseen patterns.
At the same time, with probabilistic forecasting deemed more complex, we surmise it is more susceptible to overfitting, occasionally competing against the benefits from the pre-trained prior and potentially explaining the few rare cases when TFT outperforms FullFT on probabilistic metrics.

\section{Summary and conclusions}
\label{sec:conclusion}
This study investigates the use of time-series foundation models (TSFMs) for probabilistic forecasting in building energy systems. In particular, the work addressed how pre-trained TSFMs can be leveraged to tackle challenges related to limited quantities of target data in building applications. The paper compares the zero-shot performance of TSFMs with that obtained after fine-tuning, including both full fine-tuning and parameter-efficient fine-tuning (PEFT) using techniques such as low-rank adaptation (LoRA). Experimental results on real-world data from a commercial net-zero energy building demonstrate that TSFMs not only provide competitive zero-shot predictions with appropriate context lengths but also yield significant improvements in forecast accuracy and computational efficiency when fine-tuned, outperforming several state-of-the-art deep forecasting models across various operational scenarios. Specific findings reported include:
\begin{enumerate}
\item A systematic assessment of three current publicly-available TSFMs on GitHub on a forecasting task with real building energy data. We demonstrated that zero-shot predictions (i.e., \textit{without any task-specific fine-tuning}) by these TSFMs are often outperformed by state-of-the-art deep forecasting models such as N-BEATS, DeepAR, and TFT.
\item A comparison of full fine-tuning (updating all model parameters) with parameter-efficient fine-tuning (PEFT) using LoRA. We showed that fine-tuning significantly improves prediction accuracy; in particular, LoRA achieves better predictive performance than full fine-tuning while reducing training time by approximately 2.3× and lowering FLOPS by 33\% on a GPU.
\item An evaluation of TSFMs on multiple experimentally measured time-series signals (room occupancy, carbon emissions, plug loads, HVAC energy consumption) across different building zones and seasons to test generalization. We reported evidence that fine-tuned TSFMs consistently deliver lower forecast errors and smaller performance variances compared to state-of-the-art benchmarks, even under limited data conditions (e.g., training with less than 2 months of data). In fact, with such limited data, even the best time-series forecasting models like TFT do not perform well compared to TSFMs.
\item A study of the impact of varying the length of the context window on zero-shot inference. While a 24-hour context is insufficient to capture daily periodicity, we found that extending the context to 3--5 days significantly enhances prediction accuracy without incurring excessive computational cost.
\end{enumerate}

Some open opportunities include extending the current TSFM framework from univariate to multivariate forecasting to capture correlations among multiple building signals. Additionally, integrating TSFMs with real-time building management systems such as model predictive control or reinforcement learning frameworks could improve the optimality of the control policy without penalizing safety. Finally, assessing the generalizability of TSFMs by applying them to a wider range of building types and climatic conditions will be critical for verification and deployment.

\section{CRediT Author Statement}
The following CRediT roles have been assigned to the authors:
\begin{inparaitem}[\textbullet]
    \item \textbf{Conceptualization:} All authors;
    \item \textbf{Methodology:} All authors;
    \item \textbf{Software:} YJP, FG, JL, AC;
    \item \textbf{Validation:} YJP, FG, AC;
    \item \textbf{Data Curation:} CL, AC;
    \item \textbf{Writing - Original Draft:} YJP, FG, AC;
    \item \textbf{Writing - Review \& Editing:} All authors;
    \item \textbf{Supervision:} FG, CL, AC.
\end{inparaitem}

\bibliographystyle{elsarticle-harv} 
\bibliography{refs}

\begin{thebibliography}{59}
\expandafter\ifx\csname natexlab\endcsname\relax\def\natexlab#1{#1}\fi
\providecommand{\url}[1]{\texttt{#1}}
\providecommand{\href}[2]{#2}
\providecommand{\path}[1]{#1}
\providecommand{\DOIprefix}{doi:}
\providecommand{\ArXivprefix}{arXiv:}
\providecommand{\URLprefix}{URL: }
\providecommand{\Pubmedprefix}{pmid:}
\providecommand{\doi}[1]{\href{http://dx.doi.org/#1}{\path{#1}}}
\providecommand{\Pubmed}[1]{\href{pmid:#1}{\path{#1}}}
\providecommand{\bibinfo}[2]{#2}
\ifx\xfnm\relax \def\xfnm[#1]{\unskip,\space#1}\fi
\bibitem[{Ahmad et~al.(2018)Ahmad, Chen, Guo and Wang}]{ahmad2018comprehensive}
\bibinfo{author}{Ahmad, T.}, \bibinfo{author}{Chen, H.}, \bibinfo{author}{Guo,
  Y.}, \bibinfo{author}{Wang, J.}, \bibinfo{year}{2018}.
\newblock \bibinfo{title}{A comprehensive overview on the data driven and large
  scale based approaches for forecasting of building energy demand: A review}.
\newblock \bibinfo{journal}{Energy and Buildings} \bibinfo{volume}{165},
  \bibinfo{pages}{301--320}.
\bibitem[{Alexandrov et~al.(2020)Alexandrov, Benidis, Bohlke-Schneider,
  Flunkert, Gasthaus et~al.}]{gluonts_jmlr}
\bibinfo{author}{Alexandrov, A.}, \bibinfo{author}{Benidis, K.},
  \bibinfo{author}{Bohlke-Schneider, M.}, \bibinfo{author}{Flunkert, V.},
  \bibinfo{author}{Gasthaus, J.}, et~al., \bibinfo{year}{2020}.
\newblock \bibinfo{title}{{GluonTS: Probabilistic and Neural Time Series
  Modeling in Python}}.
\newblock \bibinfo{journal}{Journal of Machine Learning Research}
  \bibinfo{volume}{21}, \bibinfo{pages}{1--6}.
\bibitem[{Ansari et~al.(2024)Ansari, Stella, Turkmen, Zhang, Mercado
  et~al.}]{ansari2024Chronos}
\bibinfo{author}{Ansari, A.F.}, \bibinfo{author}{Stella, L.},
  \bibinfo{author}{Turkmen, C.}, \bibinfo{author}{Zhang, X.},
  \bibinfo{author}{Mercado, P.}, et~al., \bibinfo{year}{2024}.
\newblock \bibinfo{title}{Chronos: Learning the language of time series}.
\newblock \bibinfo{journal}{arXiv preprint arXiv:2403.07815} .
\bibitem[{Arroyo et~al.(2022)Arroyo, Manna, Spiessens and
  Helsen}]{arroyo2022reinforced}
\bibinfo{author}{Arroyo, J.}, \bibinfo{author}{Manna, C.},
  \bibinfo{author}{Spiessens, F.}, \bibinfo{author}{Helsen, L.},
  \bibinfo{year}{2022}.
\newblock \bibinfo{title}{Reinforced model predictive control (rl-mpc) for
  building energy management}.
\newblock \bibinfo{journal}{Applied Energy} \bibinfo{volume}{309},
  \bibinfo{pages}{118346}.
\bibitem[{Azizan(2020)}]{azizan2020optimization}
\bibinfo{author}{Azizan, N.}, \bibinfo{year}{2020}.
\newblock \bibinfo{title}{Optimization algorithms for large-scale systems: From
  deep learning to energy markets}.
\newblock \bibinfo{journal}{ACM SIGMETRICS Performance Evaluation Review}
  \bibinfo{volume}{47}, \bibinfo{pages}{2--5}.
\bibitem[{Azizan et~al.(2021)Azizan, Lale and Hassibi}]{azizan2021stochastic}
\bibinfo{author}{Azizan, N.}, \bibinfo{author}{Lale, S.},
  \bibinfo{author}{Hassibi, B.}, \bibinfo{year}{2021}.
\newblock \bibinfo{title}{Stochastic mirror descent on overparameterized
  nonlinear models}.
\newblock \bibinfo{journal}{IEEE Transactions on Neural Networks and Learning
  Systems} \bibinfo{volume}{33}, \bibinfo{pages}{7717--7727}.
\bibitem[{Bommasani et~al.(2021)Bommasani, Hudson, Adeli, Altman, Arora
  et~al.}]{bommasani2021opportunities}
\bibinfo{author}{Bommasani, R.}, \bibinfo{author}{Hudson, D.A.},
  \bibinfo{author}{Adeli, E.}, \bibinfo{author}{Altman, R.},
  \bibinfo{author}{Arora, S.}, et~al., \bibinfo{year}{2021}.
\newblock \bibinfo{title}{On the opportunities and risks of foundation models}.
\newblock \bibinfo{journal}{arXiv preprint arXiv:2108.07258} .
\bibitem[{Botman et~al.(2024)Botman, Lago, Fu, Chia, Wolf
  et~al.}]{botman2024building}
\bibinfo{author}{Botman, L.}, \bibinfo{author}{Lago, J.}, \bibinfo{author}{Fu,
  X.}, \bibinfo{author}{Chia, K.}, \bibinfo{author}{Wolf, J.}, et~al.,
  \bibinfo{year}{2024}.
\newblock \bibinfo{title}{Building plug load mode detection, forecasting and
  scheduling}.
\newblock \bibinfo{journal}{Applied Energy} \bibinfo{volume}{364},
  \bibinfo{pages}{123098}.
\bibitem[{Bouckaert et~al.(2021)Bouckaert, Pales, McGlade, Remme, Wanner
  et~al.}]{bouckaert2021net}
\bibinfo{author}{Bouckaert, S.}, \bibinfo{author}{Pales, A.F.},
  \bibinfo{author}{McGlade, C.}, \bibinfo{author}{Remme, U.},
  \bibinfo{author}{Wanner, B.}, et~al., \bibinfo{year}{2021}.
\newblock \bibinfo{title}{Net zero by 2050: A roadmap for the global energy
  sector}.
\bibitem[{Bourdeau et~al.(2019)Bourdeau, qiang Zhai, Nefzaoui, Guo and
  Chatellier}]{bourdeau2019modeling}
\bibinfo{author}{Bourdeau, M.}, \bibinfo{author}{qiang Zhai, X.},
  \bibinfo{author}{Nefzaoui, E.}, \bibinfo{author}{Guo, X.},
  \bibinfo{author}{Chatellier, P.}, \bibinfo{year}{2019}.
\newblock \bibinfo{title}{Modeling and forecasting building energy consumption:
  A review of data-driven techniques}.
\newblock \bibinfo{journal}{Sustainable Cities and Society}
  \bibinfo{volume}{48}, \bibinfo{pages}{101533}.
\bibitem[{Brown et~al.(2020)Brown, Mann, Ryder, Subbiah, Kaplan
  et~al.}]{brown2020language}
\bibinfo{author}{Brown, T.}, \bibinfo{author}{Mann, B.},
  \bibinfo{author}{Ryder, N.}, \bibinfo{author}{Subbiah, M.},
  \bibinfo{author}{Kaplan, J.D.}, et~al., \bibinfo{year}{2020}.
\newblock \bibinfo{title}{Language models are few-shot learners}.
\newblock \bibinfo{journal}{Advances in Neural Information Processing Systems}
  \bibinfo{volume}{33}, \bibinfo{pages}{1877--1901}.
\bibitem[{Chakrabarty et~al.(2024)Chakrabarty, Vanfretti, Tang, Paulson, Zhan
  et~al.}]{chakrabarty2024ccta}
\bibinfo{author}{Chakrabarty, A.}, \bibinfo{author}{Vanfretti, L.},
  \bibinfo{author}{Tang, W.T.}, \bibinfo{author}{Paulson, J.A.},
  \bibinfo{author}{Zhan, S.}, et~al., \bibinfo{year}{2024}.
\newblock \bibinfo{title}{Assessing building control performance using
  physics-based simulation models and deep generative networks}, in:
  \bibinfo{booktitle}{2024 IEEE Conference on Control Technology and
  Applications (CCTA)}, \bibinfo{organization}{IEEE}. pp.
  \bibinfo{pages}{547--554}.
\bibitem[{Cox et~al.(2019)Cox, Kim, Cho and Mago}]{cox2019real}
\bibinfo{author}{Cox, S.J.}, \bibinfo{author}{Kim, D.}, \bibinfo{author}{Cho,
  H.}, \bibinfo{author}{Mago, P.}, \bibinfo{year}{2019}.
\newblock \bibinfo{title}{Real time optimal control of district cooling system
  with thermal energy storage using neural networks}.
\newblock \bibinfo{journal}{Applied energy} \bibinfo{volume}{238},
  \bibinfo{pages}{466--480}.
\bibitem[{Dao et~al.(2022)Dao, Fu, Ermon, Rudra and
  R{\'e}}]{dao2022flashattention}
\bibinfo{author}{Dao, T.}, \bibinfo{author}{Fu, D.Y.}, \bibinfo{author}{Ermon,
  S.}, \bibinfo{author}{Rudra, A.}, \bibinfo{author}{R{\'e}, C.},
  \bibinfo{year}{2022}.
\newblock \bibinfo{title}{Flash{A}ttention: Fast and memory-efficient exact
  attention with {IO}-awareness}, in: \bibinfo{booktitle}{Advances in Neural
  Information Processing Systems}, pp. \bibinfo{pages}{16344--16359}.
\bibitem[{Das et~al.(2024)Das, Kong, Sen and Zhou}]{das2023decoder}
\bibinfo{author}{Das, A.}, \bibinfo{author}{Kong, W.}, \bibinfo{author}{Sen,
  R.}, \bibinfo{author}{Zhou, Y.}, \bibinfo{year}{2024}.
\newblock \bibinfo{title}{A decoder-only foundation model for time-series
  forecasting}, in: \bibinfo{editor}{Salakhutdinov, R.},
  \bibinfo{editor}{Kolter, Z.}, \bibinfo{editor}{Heller, K.},
  \bibinfo{editor}{Weller, A.}, \bibinfo{editor}{Oliver, N.},
  \bibinfo{editor}{Scarlett, J.}, \bibinfo{editor}{Berkenkamp, F.} (Eds.),
  \bibinfo{booktitle}{Proceedings of the 41st International Conference on
  Machine Learning}, \bibinfo{publisher}{PMLR}. pp.
  \bibinfo{pages}{10148--10167}.
\bibitem[{Devlin(2018)}]{devlin2018bert}
\bibinfo{author}{Devlin, J.}, \bibinfo{year}{2018}.
\newblock \bibinfo{title}{{BERT}: Pre-training of deep bidirectional
  transformers for language understanding}.
\newblock \bibinfo{journal}{arXiv preprint arXiv:1810.04805} .
\bibitem[{Dong et~al.(2025)Dong, Luo, Yuan, Tian, Zhang
  et~al.}]{dong2025building}
\bibinfo{author}{Dong, X.}, \bibinfo{author}{Luo, Y.}, \bibinfo{author}{Yuan,
  S.}, \bibinfo{author}{Tian, Z.}, \bibinfo{author}{Zhang, L.}, et~al.,
  \bibinfo{year}{2025}.
\newblock \bibinfo{title}{Building electricity load forecasting based on
  spatiotemporal correlation and electricity consumption behavior information}.
\newblock \bibinfo{journal}{Applied Energy} \bibinfo{volume}{377},
  \bibinfo{pages}{124580}.
\bibitem[{Drgo{\v{n}}a et~al.(2020)Drgo{\v{n}}a, Arroyo, Figueroa, Blum, Arendt
  et~al.}]{drgovna2020all}
\bibinfo{author}{Drgo{\v{n}}a, J.}, \bibinfo{author}{Arroyo, J.},
  \bibinfo{author}{Figueroa, I.C.}, \bibinfo{author}{Blum, D.},
  \bibinfo{author}{Arendt, K.}, et~al., \bibinfo{year}{2020}.
\newblock \bibinfo{title}{All you need to know about model predictive control
  for buildings}.
\newblock \bibinfo{journal}{Annual Reviews in Control} \bibinfo{volume}{50},
  \bibinfo{pages}{190--232}.
\bibitem[{Du~Preez and Witt(2003)}]{du2003univariate}
\bibinfo{author}{Du~Preez, J.}, \bibinfo{author}{Witt, S.F.},
  \bibinfo{year}{2003}.
\newblock \bibinfo{title}{Univariate versus multivariate time series
  forecasting: an application to international tourism demand}.
\newblock \bibinfo{journal}{International Journal of Forecasting}
  \bibinfo{volume}{19}, \bibinfo{pages}{435--451}.
\bibitem[{Fan et~al.(2017)Fan, Xiao and Zhao}]{fan2017short}
\bibinfo{author}{Fan, C.}, \bibinfo{author}{Xiao, F.}, \bibinfo{author}{Zhao,
  Y.}, \bibinfo{year}{2017}.
\newblock \bibinfo{title}{A short-term building cooling load prediction method
  using deep learning algorithms}.
\newblock \bibinfo{journal}{Applied Energy} \bibinfo{volume}{195},
  \bibinfo{pages}{222--233}.
\bibitem[{Geraldi and Ghisi(2022)}]{geraldi2022data}
\bibinfo{author}{Geraldi, M.S.}, \bibinfo{author}{Ghisi, E.},
  \bibinfo{year}{2022}.
\newblock \bibinfo{title}{Data-driven framework towards realistic bottom-up
  energy benchmarking using an artificial neural network}.
\newblock \bibinfo{journal}{Applied Energy} \bibinfo{volume}{306},
  \bibinfo{pages}{117960}.
\bibitem[{Graves and Graves(2012)}]{graves2012long}
\bibinfo{author}{Graves, A.}, \bibinfo{author}{Graves, A.},
  \bibinfo{year}{2012}.
\newblock \bibinfo{title}{Long short-term memory}, in:
  \bibinfo{booktitle}{Supervised Sequence Labelling With Recurrent Neural
  Networks}. \bibinfo{publisher}{Springer}, pp. \bibinfo{pages}{37--45}.
\bibitem[{Heidari et~al.(2022)Heidari, Mar{\'e}chal and
  Khovalyg}]{heidari2022reinforcement}
\bibinfo{author}{Heidari, A.}, \bibinfo{author}{Mar{\'e}chal, F.},
  \bibinfo{author}{Khovalyg, D.}, \bibinfo{year}{2022}.
\newblock \bibinfo{title}{Reinforcement learning for proactive operation of
  residential energy systems by learning stochastic occupant behavior and
  fluctuating solar energy: Balancing comfort, hygiene and energy use}.
\newblock \bibinfo{journal}{Applied Energy} \bibinfo{volume}{318},
  \bibinfo{pages}{119206}.
\bibitem[{Hu et~al.(2021)Hu, Shen, Wallis, Allen-Zhu, Li et~al.}]{hu2021lora}
\bibinfo{author}{Hu, E.J.}, \bibinfo{author}{Shen, Y.},
  \bibinfo{author}{Wallis, P.}, \bibinfo{author}{Allen-Zhu, Z.},
  \bibinfo{author}{Li, Y.}, et~al., \bibinfo{year}{2021}.
\newblock \bibinfo{title}{Lora: Low-rank adaptation of large language models}.
\bibitem[{Jung et~al.(2020)Jung, Kim, Kwak and Park}]{jung2020worrying}
\bibinfo{author}{Jung, S.}, \bibinfo{author}{Kim, K.M.}, \bibinfo{author}{Kwak,
  H.}, \bibinfo{author}{Park, Y.J.}, \bibinfo{year}{2020}.
\newblock \bibinfo{title}{A worrying analysis of probabilistic time-series
  models for sales forecasting}.
\newblock \href{http://arxiv.org/abs/2011.10715}{{\tt arXiv:2011.10715}}.
\bibitem[{Khalil et~al.(2022)Khalil, McGough, Pourmirza, Pazhoohesh and
  Walker}]{khalil2022machine}
\bibinfo{author}{Khalil, M.}, \bibinfo{author}{McGough, A.S.},
  \bibinfo{author}{Pourmirza, Z.}, \bibinfo{author}{Pazhoohesh, M.},
  \bibinfo{author}{Walker, S.}, \bibinfo{year}{2022}.
\newblock \bibinfo{title}{Machine learning, deep learning and statistical
  analysis for forecasting building energy consumption—a systematic review}.
\newblock \bibinfo{journal}{Engineering Applications of Artificial
  Intelligence} \bibinfo{volume}{115}, \bibinfo{pages}{105287}.
\bibitem[{Kim et~al.(2024)Kim, Seomun, Lee, Cho, Chin
  et~al.}]{kim2024forecasting}
\bibinfo{author}{Kim, D.}, \bibinfo{author}{Seomun, G.}, \bibinfo{author}{Lee,
  Y.}, \bibinfo{author}{Cho, H.}, \bibinfo{author}{Chin, K.}, et~al.,
  \bibinfo{year}{2024}.
\newblock \bibinfo{title}{Forecasting building energy demand and on-site power
  generation for residential buildings using long and short-term memory method
  with transfer learning}.
\newblock \bibinfo{journal}{Applied Energy} \bibinfo{volume}{368},
  \bibinfo{pages}{123500}.
\bibitem[{Kingma(2014)}]{kingma2014adam}
\bibinfo{author}{Kingma, D.P.}, \bibinfo{year}{2014}.
\newblock \bibinfo{title}{Adam: A method for stochastic optimization}.
\newblock \bibinfo{journal}{arXiv preprint arXiv:1412.6980} .
\bibitem[{Kirillov et~al.(2023)Kirillov, Mintun, Ravi, Mao, Rolland
  et~al.}]{kirillov2023segment}
\bibinfo{author}{Kirillov, A.}, \bibinfo{author}{Mintun, E.},
  \bibinfo{author}{Ravi, N.}, \bibinfo{author}{Mao, H.},
  \bibinfo{author}{Rolland, C.}, et~al., \bibinfo{year}{2023}.
\newblock \bibinfo{title}{Segment anything}, in:
  \bibinfo{booktitle}{Proceedings of the IEEE/CVF International Conference on
  Computer Vision}, \bibinfo{publisher}{IEEE}. pp. \bibinfo{pages}{4015--4026}.
\bibitem[{Kirkpatrick et~al.(2017)Kirkpatrick, Pascanu, Rabinowitz, Veness,
  Desjardins et~al.}]{kirkpatrick2017catastrophic}
\bibinfo{author}{Kirkpatrick, J.}, \bibinfo{author}{Pascanu, R.},
  \bibinfo{author}{Rabinowitz, N.}, \bibinfo{author}{Veness, J.},
  \bibinfo{author}{Desjardins, G.}, et~al., \bibinfo{year}{2017}.
\newblock \bibinfo{title}{Overcoming catastrophic forgetting in neural
  networks}.
\newblock \bibinfo{journal}{Proceedings of the National Academy of Sciences}
  \bibinfo{volume}{114}, \bibinfo{pages}{3521--3526}.
\bibitem[{Lei et~al.(2021)Lei, Chen, Wu, Chen and Liu}]{lei2021building}
\bibinfo{author}{Lei, L.}, \bibinfo{author}{Chen, W.}, \bibinfo{author}{Wu,
  B.}, \bibinfo{author}{Chen, C.}, \bibinfo{author}{Liu, W.},
  \bibinfo{year}{2021}.
\newblock \bibinfo{title}{A building energy consumption prediction model based
  on rough set theory and deep learning algorithms}.
\newblock \bibinfo{journal}{Energy and Buildings} \bibinfo{volume}{240},
  \bibinfo{pages}{110886}.
\bibitem[{Liang et~al.(2023)Liang, Chen, Zhu, Jin and Du}]{liang2023domain}
\bibinfo{author}{Liang, X.}, \bibinfo{author}{Chen, S.}, \bibinfo{author}{Zhu,
  X.}, \bibinfo{author}{Jin, X.}, \bibinfo{author}{Du, Z.},
  \bibinfo{year}{2023}.
\newblock \bibinfo{title}{Domain knowledge decomposition of building energy
  consumption and a hybrid data-driven model for 24-h ahead predictions}.
\newblock \bibinfo{journal}{Applied Energy} \bibinfo{volume}{344},
  \bibinfo{pages}{121244}.
\bibitem[{Liao et~al.(2025)Liao, Wang, Yang, Yang, Fang
  et~al.}]{liao2025timegpt}
\bibinfo{author}{Liao, W.}, \bibinfo{author}{Wang, S.}, \bibinfo{author}{Yang,
  D.}, \bibinfo{author}{Yang, Z.}, \bibinfo{author}{Fang, J.}, et~al.,
  \bibinfo{year}{2025}.
\newblock \bibinfo{title}{{TimeGPT} in load forecasting: A large time series
  model perspective}.
\newblock \bibinfo{journal}{Applied Energy} \bibinfo{volume}{379},
  \bibinfo{pages}{124973}.
\bibitem[{Lim et~al.(2021)Lim, Ar{\i}k, Loeff and Pfister}]{lim2021temporal}
\bibinfo{author}{Lim, B.}, \bibinfo{author}{Ar{\i}k, S.{\"O}.},
  \bibinfo{author}{Loeff, N.}, \bibinfo{author}{Pfister, T.},
  \bibinfo{year}{2021}.
\newblock \bibinfo{title}{Temporal fusion transformers for interpretable
  multi-horizon time series forecasting}.
\newblock \bibinfo{journal}{International Journal of Forecasting}
  \bibinfo{volume}{37}, \bibinfo{pages}{1748--1764}.
\bibitem[{Loshchilov(2017)}]{loshchilov2017decoupled}
\bibinfo{author}{Loshchilov, I.}, \bibinfo{year}{2017}.
\newblock \bibinfo{title}{Decoupled weight decay regularization}.
\newblock \bibinfo{journal}{arXiv preprint arXiv:1711.05101} .
\bibitem[{Min et~al.(2022)Min, Ahn and Azizan}]{min2022one}
\bibinfo{author}{Min, Y.}, \bibinfo{author}{Ahn, K.}, \bibinfo{author}{Azizan,
  N.}, \bibinfo{year}{2022}.
\newblock \bibinfo{title}{One-pass learning via bridging orthogonal gradient
  descent and recursive least-squares}, in: \bibinfo{booktitle}{2022 IEEE 61st
  Conference on Decision and Control (CDC)}, \bibinfo{organization}{IEEE}. pp.
  \bibinfo{pages}{4720--4725}.
\bibitem[{Mohebi et~al.(2025)Mohebi, Li and Wang}]{mohebi2025chance}
\bibinfo{author}{Mohebi, P.}, \bibinfo{author}{Li, S.}, \bibinfo{author}{Wang,
  Z.}, \bibinfo{year}{2025}.
\newblock \bibinfo{title}{Chance-constrained stochastic framework for building
  thermal control under forecast uncertainties}.
\newblock \bibinfo{journal}{Energy and Buildings} , \bibinfo{pages}{115385}.
\bibitem[{Morcillo-Jimenez et~al.(2024)Morcillo-Jimenez, Mesa,
  G{\'o}mez-Romero, Vila and Martin-Bautista}]{morcillo2024deep}
\bibinfo{author}{Morcillo-Jimenez, R.}, \bibinfo{author}{Mesa, J.},
  \bibinfo{author}{G{\'o}mez-Romero, J.}, \bibinfo{author}{Vila, M.A.},
  \bibinfo{author}{Martin-Bautista, M.J.}, \bibinfo{year}{2024}.
\newblock \bibinfo{title}{Deep learning for prediction of energy consumption:
  an applied use case in an office building}.
\newblock \bibinfo{journal}{Applied Intelligence} \bibinfo{volume}{54},
  \bibinfo{pages}{5813--5825}.
\bibitem[{Mulayim et~al.(2024)Mulayim, Quan, Han, Ouyang, Hong, Berg{\'e}s and
  Srivastava}]{mulayim2024time}
\bibinfo{author}{Mulayim, O.B.}, \bibinfo{author}{Quan, P.},
  \bibinfo{author}{Han, L.}, \bibinfo{author}{Ouyang, X.},
  \bibinfo{author}{Hong, D.}, \bibinfo{author}{Berg{\'e}s, M.},
  \bibinfo{author}{Srivastava, M.}, \bibinfo{year}{2024}.
\newblock \bibinfo{title}{Are time series foundation models ready to
  revolutionize predictive building analytics?}, in:
  \bibinfo{booktitle}{Proceedings of the 11th ACM International Conference on
  Systems for Energy-Efficient Buildings, Cities, and Transportation}, pp.
  \bibinfo{pages}{169--173}.
\bibitem[{Nejat et~al.(2015)Nejat, Jomehzadeh, Taheri, Gohari and
  Majid}]{nejat2015global}
\bibinfo{author}{Nejat, P.}, \bibinfo{author}{Jomehzadeh, F.},
  \bibinfo{author}{Taheri, M.M.}, \bibinfo{author}{Gohari, M.},
  \bibinfo{author}{Majid, M.Z.A.}, \bibinfo{year}{2015}.
\newblock \bibinfo{title}{A global review of energy consumption, co2 emissions
  and policy in the residential sector (with an overview of the top ten co2
  emitting countries)}.
\newblock \bibinfo{journal}{Renewable and Sustainable Energy Reviews}
  \bibinfo{volume}{43}, \bibinfo{pages}{843--862}.
\bibitem[{Oreshkin et~al.(2019)Oreshkin, Carpov, Chapados and
  Bengio}]{oreshkin2019n}
\bibinfo{author}{Oreshkin, B.N.}, \bibinfo{author}{Carpov, D.},
  \bibinfo{author}{Chapados, N.}, \bibinfo{author}{Bengio, Y.},
  \bibinfo{year}{2019}.
\newblock \bibinfo{title}{{N-BEATS: N}eural basis expansion analysis for
  interpretable time series forecasting}.
\newblock \bibinfo{journal}{arXiv preprint arXiv:1905.10437} .
\bibitem[{Park et~al.(2022)Park, Kim, Odermatt, Lee and Kim}]{park2022large}
\bibinfo{author}{Park, Y.J.}, \bibinfo{author}{Kim, D.},
  \bibinfo{author}{Odermatt, F.}, \bibinfo{author}{Lee, J.},
  \bibinfo{author}{Kim, K.M.}, \bibinfo{year}{2022}.
\newblock \bibinfo{title}{A large-scale ensemble learning framework for demand
  forecasting}, in: \bibinfo{booktitle}{2022 IEEE International Conference on
  Data Mining (ICDM)}, \bibinfo{organization}{IEEE}. pp.
  \bibinfo{pages}{378--387}.
\bibitem[{P{\'e}rez-Lombard et~al.(2008)P{\'e}rez-Lombard, Ortiz and
  Pout}]{perez2008review}
\bibinfo{author}{P{\'e}rez-Lombard, L.}, \bibinfo{author}{Ortiz, J.},
  \bibinfo{author}{Pout, C.}, \bibinfo{year}{2008}.
\newblock \bibinfo{title}{A review on buildings energy consumption
  information}.
\newblock \bibinfo{journal}{Energy and Buildings} \bibinfo{volume}{40},
  \bibinfo{pages}{394--398}.
\bibitem[{Radford et~al.(2021)Radford, Kim, Hallacy, Ramesh, Goh
  et~al.}]{radford2021learning}
\bibinfo{author}{Radford, A.}, \bibinfo{author}{Kim, J.W.},
  \bibinfo{author}{Hallacy, C.}, \bibinfo{author}{Ramesh, A.},
  \bibinfo{author}{Goh, G.}, et~al., \bibinfo{year}{2021}.
\newblock \bibinfo{title}{Learning transferable visual models from natural
  language supervision}, in: \bibinfo{booktitle}{International Conference on
  Machine Learning}, \bibinfo{organization}{PMLR}. pp.
  \bibinfo{pages}{8748--8763}.
\bibitem[{Salinas et~al.(2020)Salinas, Flunkert, Gasthaus and
  Januschowski}]{salinas2020deepar}
\bibinfo{author}{Salinas, D.}, \bibinfo{author}{Flunkert, V.},
  \bibinfo{author}{Gasthaus, J.}, \bibinfo{author}{Januschowski, T.},
  \bibinfo{year}{2020}.
\newblock \bibinfo{title}{{DeepAR}: Probabilistic forecasting with
  autoregressive recurrent networks}.
\newblock \bibinfo{journal}{International Journal of Forecasting}
  \bibinfo{volume}{36}, \bibinfo{pages}{1181--1191}.
\bibitem[{Sorourifar et~al.(2024)Sorourifar, Paulson, Wang, Quirynen, Laughman
  et~al.}]{sorourifar2024ccta}
\bibinfo{author}{Sorourifar, F.}, \bibinfo{author}{Paulson, J.A.},
  \bibinfo{author}{Wang, Y.}, \bibinfo{author}{Quirynen, R.},
  \bibinfo{author}{Laughman, C.R.}, et~al., \bibinfo{year}{2024}.
\newblock \bibinfo{title}{Bayesian forecasting with deep generative disturbance
  models in stochastic {MPC} for building energy systems}, in:
  \bibinfo{booktitle}{2024 IEEE Conference on Control Technology and
  Applications (CCTA)}, pp. \bibinfo{pages}{414--419}.
\newblock \DOIprefix\doi{10.1109/CCTA60707.2024.10666537}.
\bibitem[{Sun et~al.(2025)Sun, Hu, Mae and Imaizumi}]{sun2025deep}
\bibinfo{author}{Sun, L.}, \bibinfo{author}{Hu, Z.}, \bibinfo{author}{Mae, M.},
  \bibinfo{author}{Imaizumi, T.}, \bibinfo{year}{2025}.
\newblock \bibinfo{title}{Deep transfer learning strategy based on
  timesblock-cdan for predicting thermal environment and air conditioner energy
  consumption in residential buildings}.
\newblock \bibinfo{journal}{Applied Energy} \bibinfo{volume}{381},
  \bibinfo{pages}{125188}.
\bibitem[{Tan et~al.(2024)Tan, Merrill, Gupta, Althoff and
  Hartvigsen}]{tan2024language}
\bibinfo{author}{Tan, M.}, \bibinfo{author}{Merrill, M.},
  \bibinfo{author}{Gupta, V.}, \bibinfo{author}{Althoff, T.},
  \bibinfo{author}{Hartvigsen, T.}, \bibinfo{year}{2024}.
\newblock \bibinfo{title}{Are language models actually useful for time series
  forecasting?}
\newblock \bibinfo{journal}{Advances in Neural Information Processing Systems}
  \bibinfo{volume}{37}, \bibinfo{pages}{60162--60191}.
\bibitem[{Van Den~Oord et~al.(2017)Van Den~Oord, Vinyals
  et~al.}]{vanoord2017vqvae}
\bibinfo{author}{Van Den~Oord, A.}, \bibinfo{author}{Vinyals, O.}, et~al.,
  \bibinfo{year}{2017}.
\newblock \bibinfo{title}{Neural discrete representation learning}.
\newblock \bibinfo{journal}{Advances in neural information processing systems}
  \bibinfo{volume}{30}.
\bibitem[{Vaswani et~al.(2017)Vaswani, Shazeer, Parmar, Uszkoreit, Jones
  et~al.}]{vaswani2017attention}
\bibinfo{author}{Vaswani, A.}, \bibinfo{author}{Shazeer, N.},
  \bibinfo{author}{Parmar, N.}, \bibinfo{author}{Uszkoreit, J.},
  \bibinfo{author}{Jones, L.}, et~al., \bibinfo{year}{2017}.
\newblock \bibinfo{title}{Attention is all you need}.
\newblock \bibinfo{journal}{Advances in Neural Information Processing Systems}
  \bibinfo{volume}{30}.
\bibitem[{Wolf et~al.(2020)Wolf, Debut, Sanh, Chaumond, Delangue
  et~al.}]{wolf-etal-2020-transformers}
\bibinfo{author}{Wolf, T.}, \bibinfo{author}{Debut, L.}, \bibinfo{author}{Sanh,
  V.}, \bibinfo{author}{Chaumond, J.}, \bibinfo{author}{Delangue, C.}, et~al.,
  \bibinfo{year}{2020}.
\newblock \bibinfo{title}{Transformers: State-of-the-art natural language
  processing}, in: \bibinfo{booktitle}{Proceedings of the 2020 Conference on
  Empirical Methods in Natural Language Processing: System Demonstrations},
  \bibinfo{publisher}{Association for Computational Linguistics}. pp.
  \bibinfo{pages}{38--45}.
\bibitem[{Woo et~al.(2024)Woo, Liu, Kumar, Xiong, Savarese
  et~al.}]{woo2024unified}
\bibinfo{author}{Woo, G.}, \bibinfo{author}{Liu, C.}, \bibinfo{author}{Kumar,
  A.}, \bibinfo{author}{Xiong, C.}, \bibinfo{author}{Savarese, S.}, et~al.,
  \bibinfo{year}{2024}.
\newblock \bibinfo{title}{Unified training of universal time series forecasting
  transformers}.
\newblock \bibinfo{journal}{arXiv preprint arXiv:2402.02592} .
\bibitem[{Xing et~al.(2024)Xing, Pan, Yang, Yuan, Liang
  et~al.}]{xing2024transfer}
\bibinfo{author}{Xing, Z.}, \bibinfo{author}{Pan, Y.}, \bibinfo{author}{Yang,
  Y.}, \bibinfo{author}{Yuan, X.}, \bibinfo{author}{Liang, Y.}, et~al.,
  \bibinfo{year}{2024}.
\newblock \bibinfo{title}{Transfer learning integrating similarity analysis for
  short-term and long-term building energy consumption prediction}.
\newblock \bibinfo{journal}{Applied Energy} \bibinfo{volume}{365},
  \bibinfo{pages}{123276}.
\bibitem[{Yang et~al.(2014)Yang, L{\'e}tourneau and Guo}]{yang2014developing}
\bibinfo{author}{Yang, C.}, \bibinfo{author}{L{\'e}tourneau, S.},
  \bibinfo{author}{Guo, H.}, \bibinfo{year}{2014}.
\newblock \bibinfo{title}{Developing data-driven models to predict bems energy
  consumption for demand response systems}, in: \bibinfo{booktitle}{Modern
  Advances in Applied Intelligence, IEA/AIE 2014},
  \bibinfo{organization}{Springer}. pp. \bibinfo{pages}{188--197}.
\bibitem[{Zeng and Lee(2023)}]{zeng2023expressive}
\bibinfo{author}{Zeng, Y.}, \bibinfo{author}{Lee, K.}, \bibinfo{year}{2023}.
\newblock \bibinfo{title}{The expressive power of low-rank adaptation}.
\newblock \bibinfo{journal}{arXiv preprint arXiv:2310.17513} .
\bibitem[{Zhang et~al.(2025a)Zhang, Zhang, Zhao and Lu}]{zhang2025automated}
\bibinfo{author}{Zhang, C.}, \bibinfo{author}{Zhang, J.},
  \bibinfo{author}{Zhao, Y.}, \bibinfo{author}{Lu, J.}, \bibinfo{year}{2025}a.
\newblock \bibinfo{title}{{Automated data-driven building energy load
  prediction method based on generative pre-trained transformers (GPT)}}.
\newblock \bibinfo{journal}{Energy} \bibinfo{volume}{318},
  \bibinfo{pages}{134824}.
\bibitem[{Zhang et~al.(2025b)Zhang, Glaws, Cortiella, Emami and
  King}]{zhang2025deep}
\bibinfo{author}{Zhang, X.}, \bibinfo{author}{Glaws, A.},
  \bibinfo{author}{Cortiella, A.}, \bibinfo{author}{Emami, P.},
  \bibinfo{author}{King, R.N.}, \bibinfo{year}{2025}b.
\newblock \bibinfo{title}{Deep generative models in energy system applications:
  {R}eview, challenges, and future directions}.
\newblock \bibinfo{journal}{Applied Energy} \bibinfo{volume}{380},
  \bibinfo{pages}{125059}.
\bibitem[{Zheng et~al.(2023)Zheng, Zhou, Liu and
  Nakanishi}]{zheng2023interpretable}
\bibinfo{author}{Zheng, P.}, \bibinfo{author}{Zhou, H.}, \bibinfo{author}{Liu,
  J.}, \bibinfo{author}{Nakanishi, Y.}, \bibinfo{year}{2023}.
\newblock \bibinfo{title}{Interpretable building energy consumption forecasting
  using spectral clustering algorithm and temporal fusion transformers
  architecture}.
\newblock \bibinfo{journal}{Applied Energy} \bibinfo{volume}{349},
  \bibinfo{pages}{121607}.
\bibitem[{Zivot and Wang(2006)}]{zivot2006modeling}
\bibinfo{author}{Zivot, E.}, \bibinfo{author}{Wang, J.}, \bibinfo{year}{2006}.
\newblock \bibinfo{title}{Modeling financial time series with S-PLUS}.
  volume~\bibinfo{volume}{2}.
\newblock \bibinfo{publisher}{Springer}.

\end{thebibliography}

\newpage
\appendix

\section{Additional Tables and Figures}

\begin{figure}[!ht]
    \centering
    \begin{subfigure}[]{0.4\textwidth}
        \centering
        \includegraphics[width=\linewidth]{figures_journal/ds_size_occ_mase.pdf}
    \end{subfigure}
    \begin{subfigure}[]{0.4\textwidth}
        \centering
        \includegraphics[width=\linewidth]{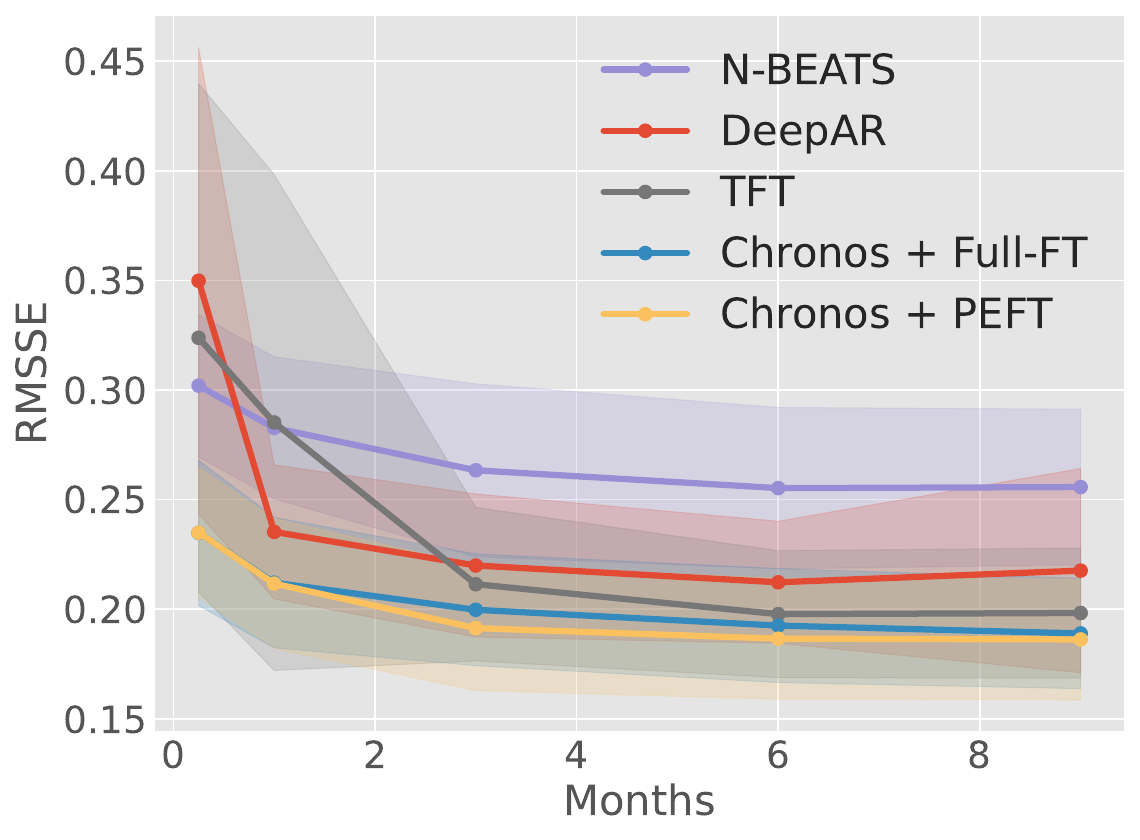}
    \end{subfigure}
    \begin{subfigure}[]{0.4\textwidth}
        \centering
        \includegraphics[width=\linewidth]{figures_journal/ds_size_occ_wql.pdf}
    \end{subfigure}
    \begin{subfigure}[]{0.4\textwidth}
        \centering
        \includegraphics[width=\linewidth]{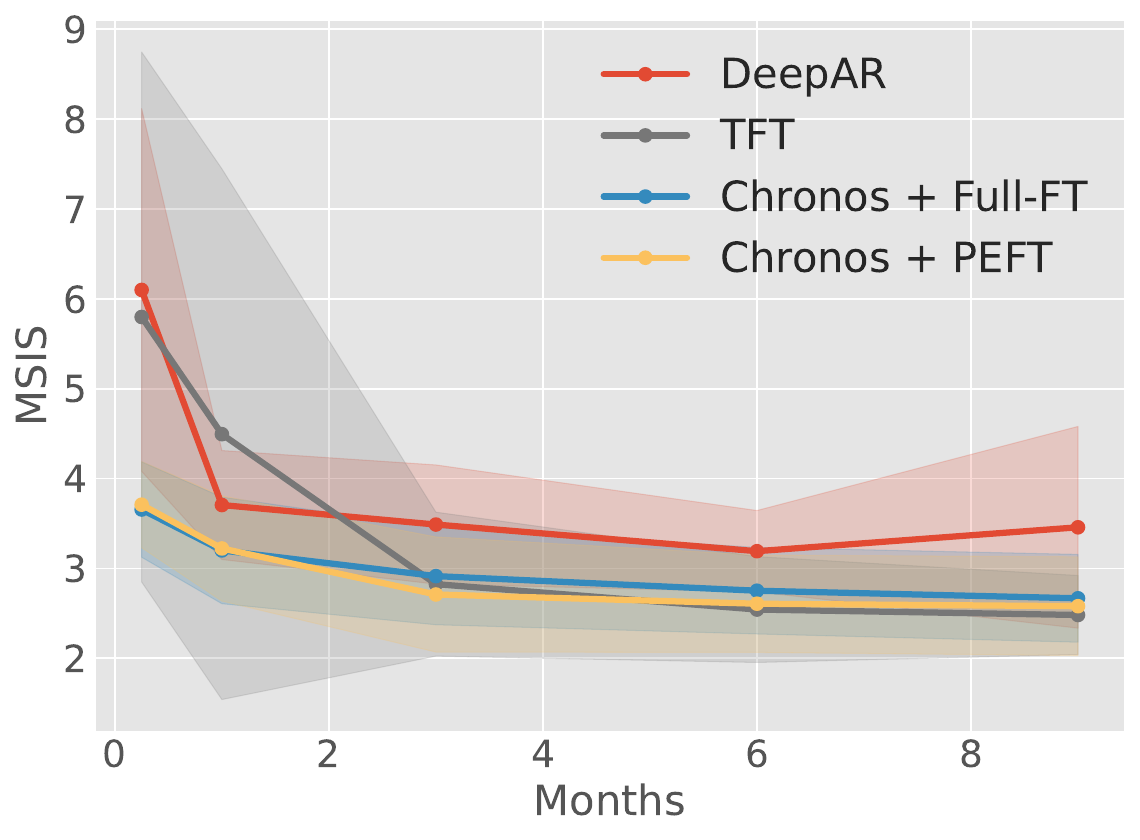}
    \end{subfigure}
    \caption{Forecasting accuracies of the benchmark models and the proposed TSFM approach using PEFT across different training dataset size on the room occupancy (\texttt{Occ}) signal. The shades represent the standard deviation across different zones and seasons.}
    \vspace{-1em}
    \label{fig:abl-occ-full}
\end{figure}

\begin{figure}[!ht]
    \centering
    \begin{subfigure}[]{0.4\textwidth}
        \centering
        \includegraphics[width=\linewidth]{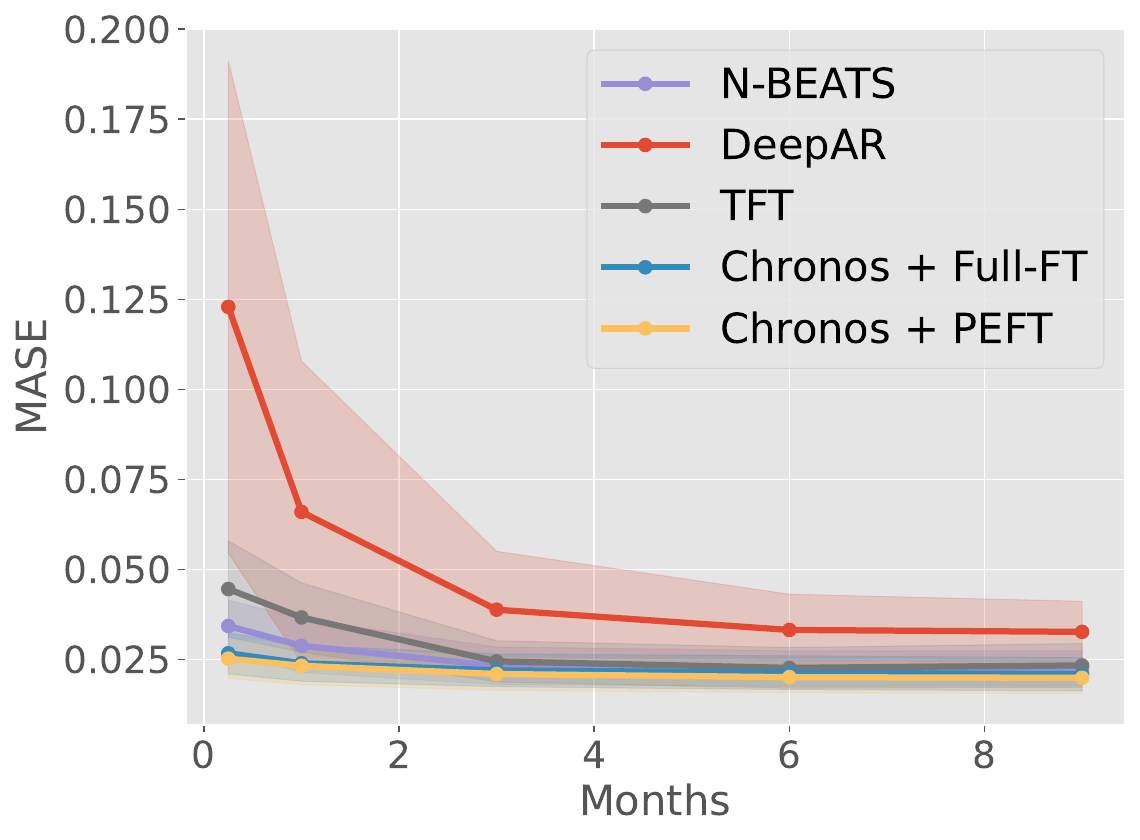}
    \end{subfigure}
    \begin{subfigure}[]{0.4\textwidth}
        \centering
        \includegraphics[width=\linewidth]{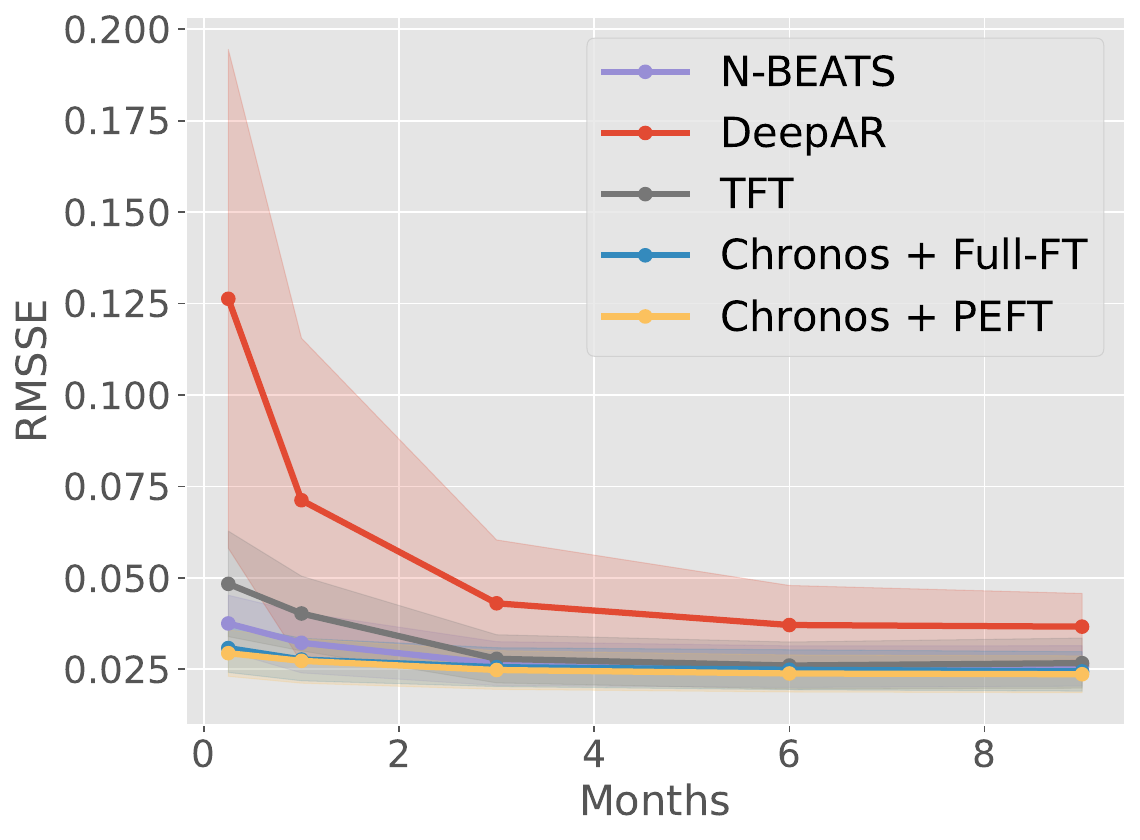}
    \end{subfigure}
    \begin{subfigure}[]{0.4\textwidth}
        \centering
        \includegraphics[width=\linewidth]{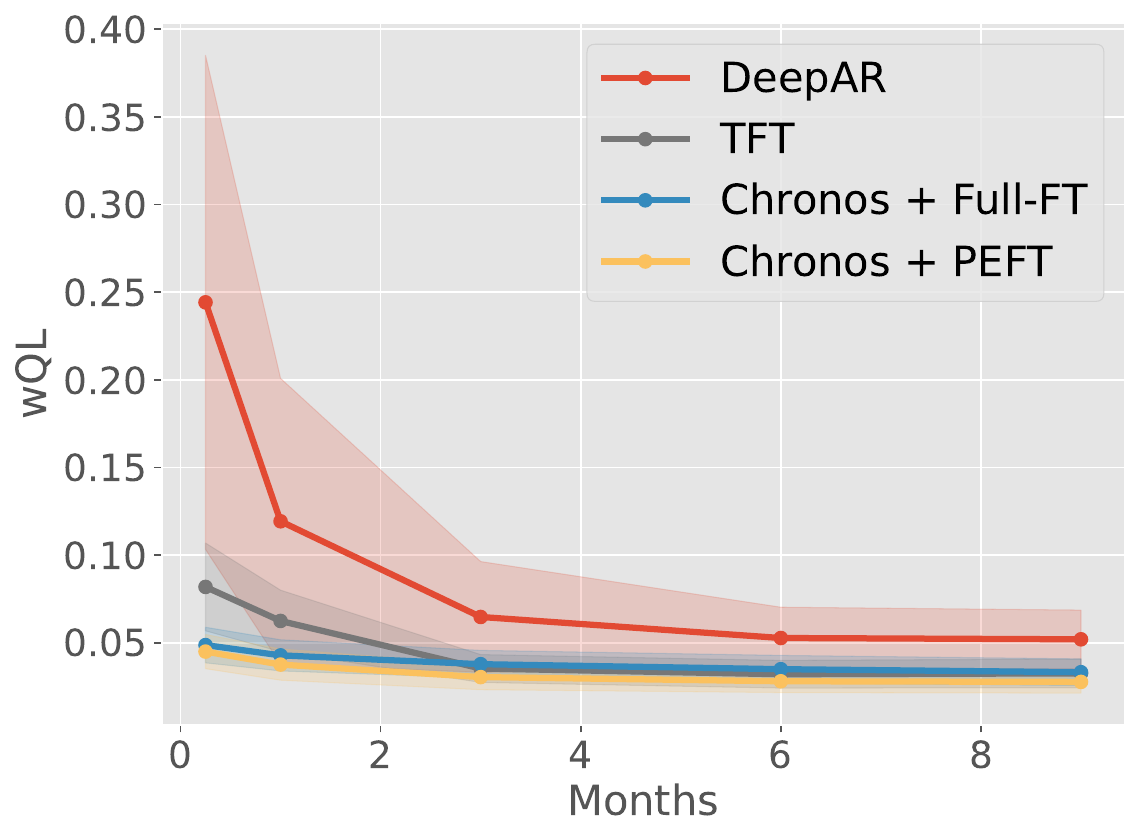}
    \end{subfigure}
    \begin{subfigure}[]{0.4\textwidth}
        \centering
        \includegraphics[width=\linewidth]{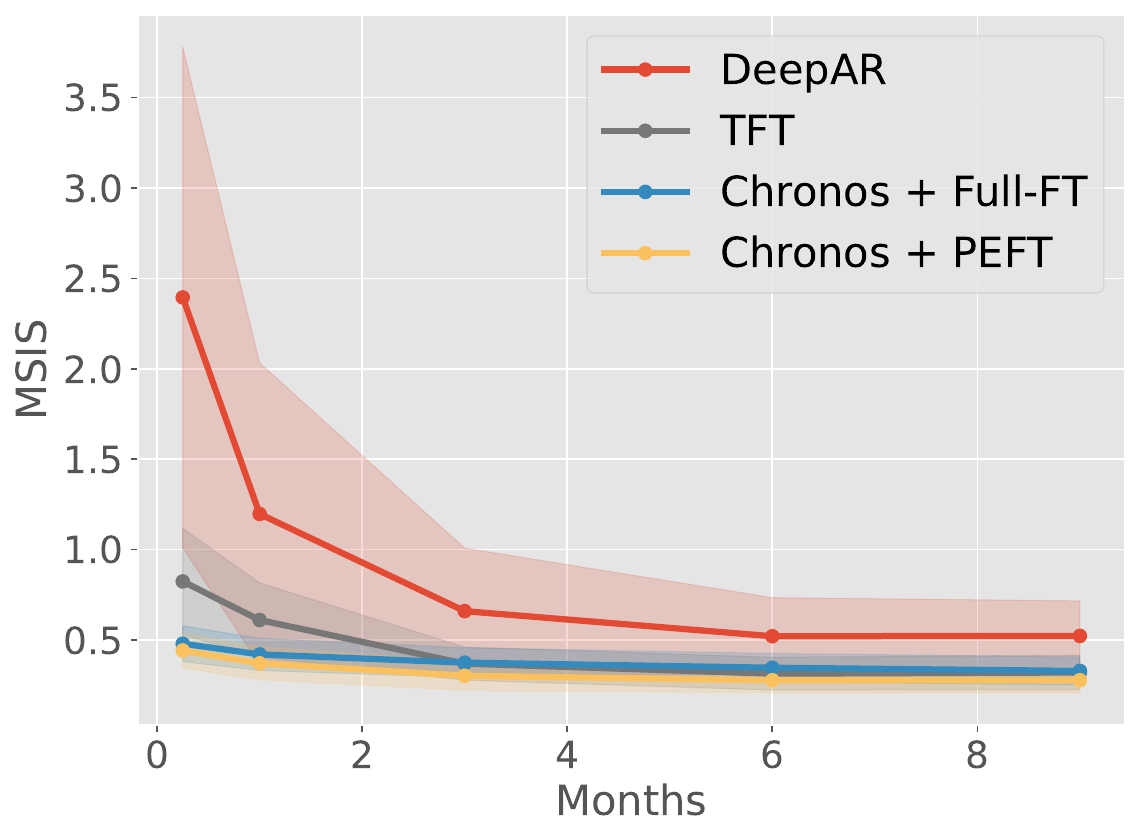}
    \end{subfigure}
    \caption{Forecasting accuracies of the benchmark models and the proposed TSFM approach using PEFT across different training dataset size on the carbon emissions (\texttt{CO2}) signal. The shades represent the standard deviation across different zones and seasons.}
    \vspace{-1em}
    \label{fig:abl-co2-full}
\end{figure}

\begin{figure}[!ht]
    \centering
    \begin{subfigure}[]{0.4\textwidth}
        \centering
        \includegraphics[width=\linewidth]{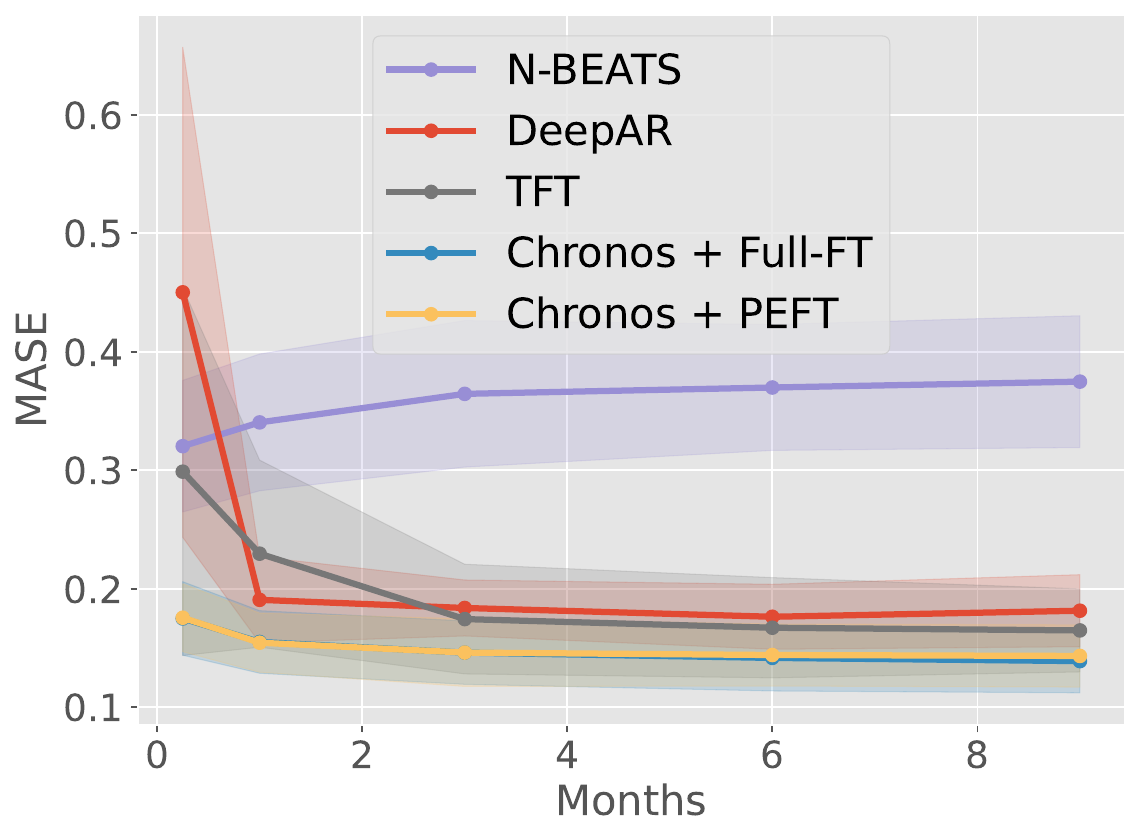}
    \end{subfigure}
    \begin{subfigure}[]{0.4\textwidth}
        \centering
        \includegraphics[width=\linewidth]{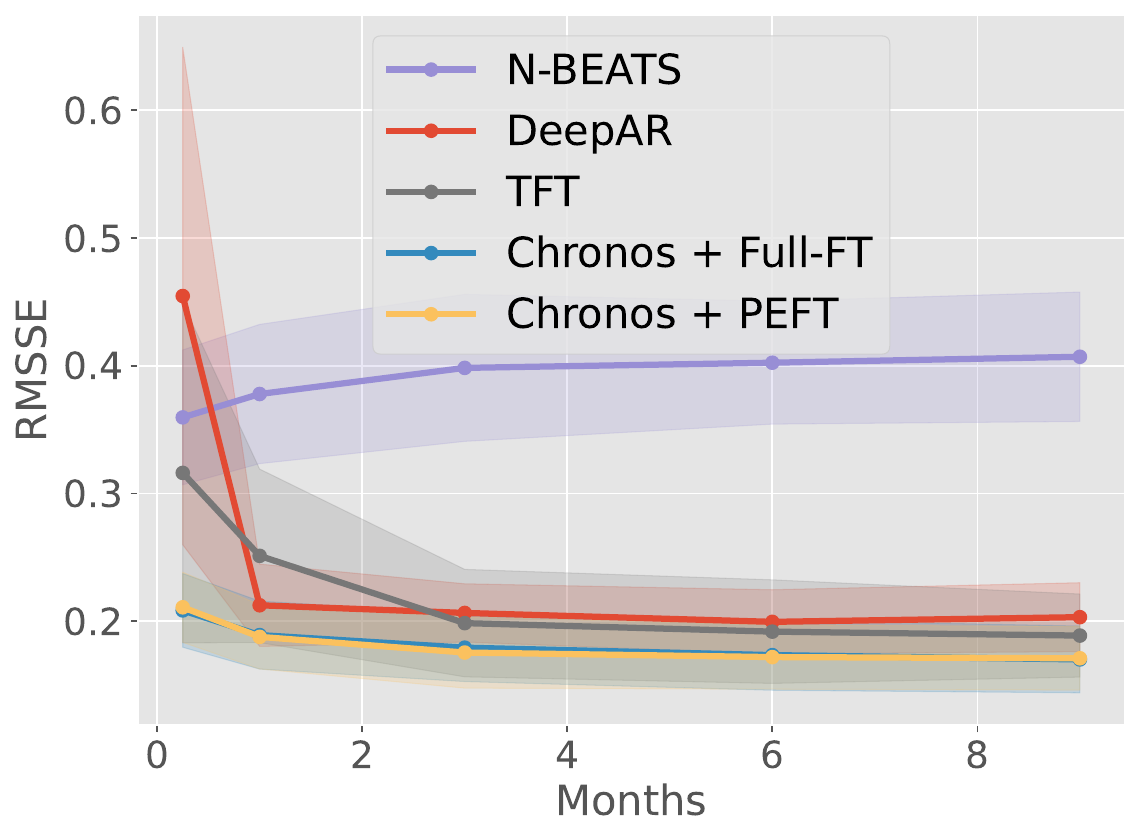}
    \end{subfigure}
    \begin{subfigure}[]{0.4\textwidth}
        \centering
        \includegraphics[width=\linewidth]{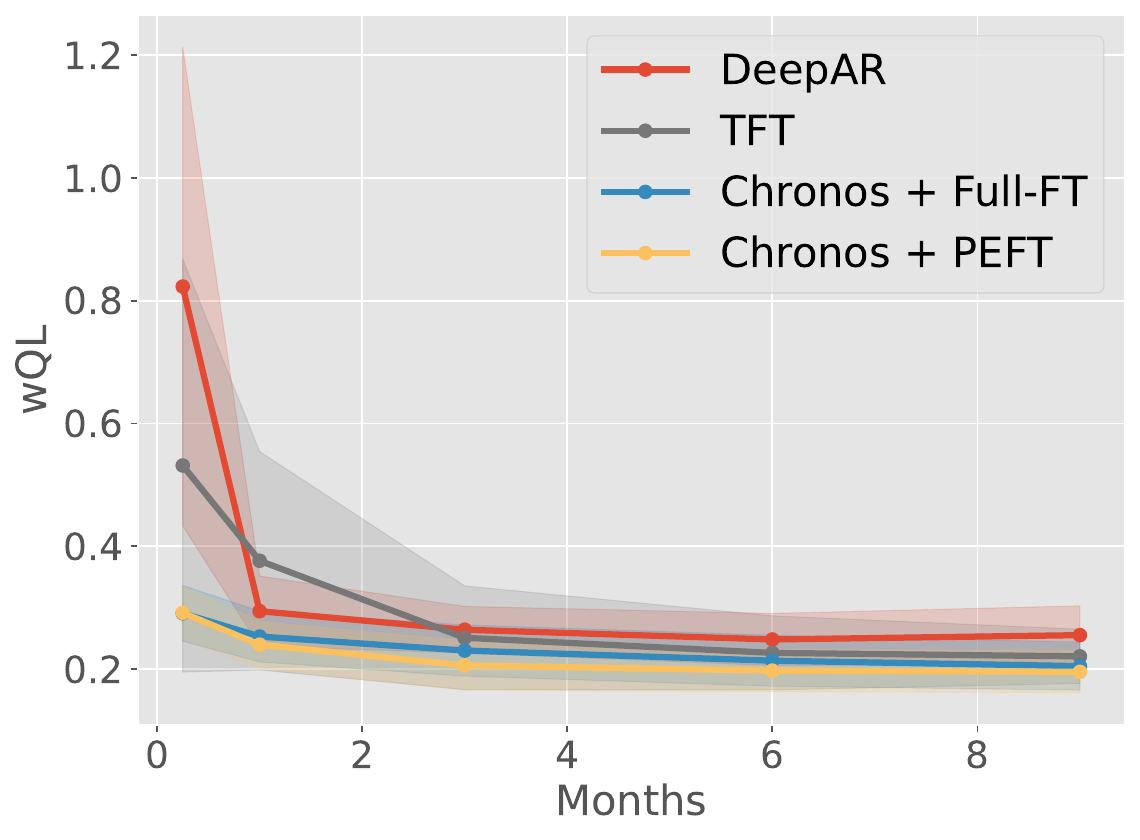}
    \end{subfigure}
    \begin{subfigure}[]{0.4\textwidth}
        \centering
        \includegraphics[width=\linewidth]{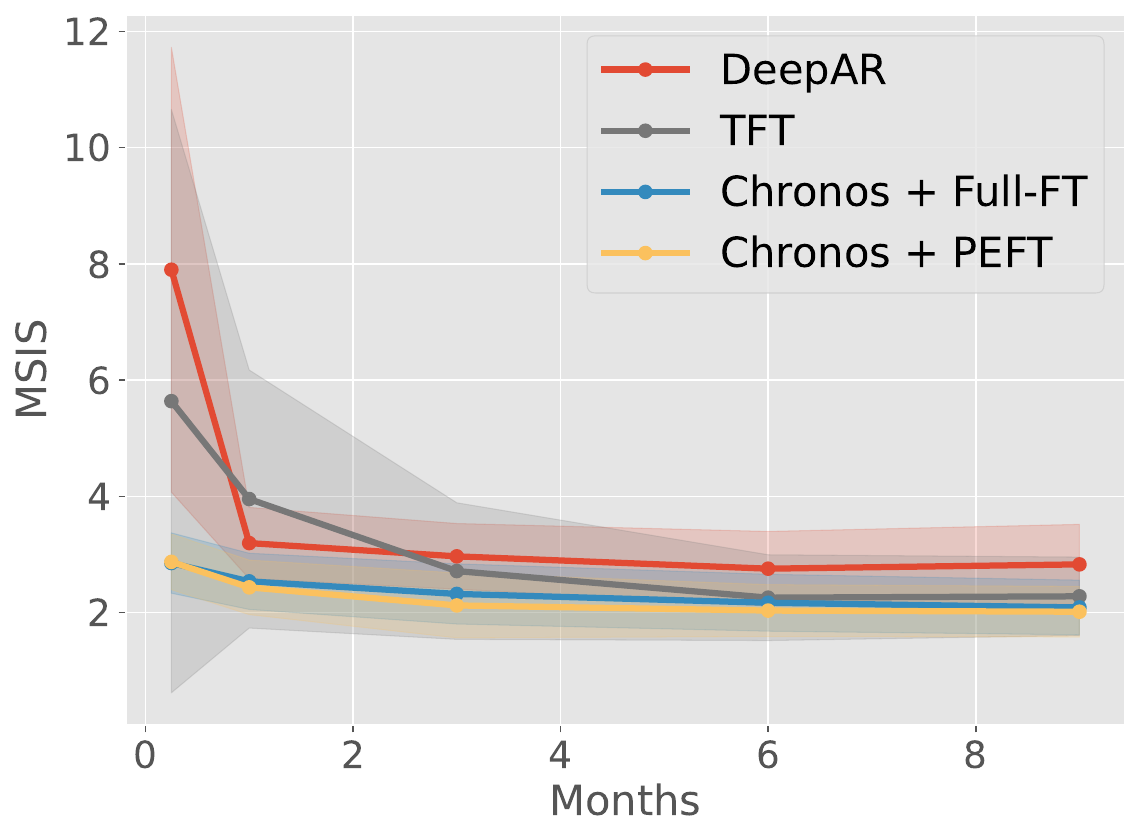}
    \end{subfigure}
    \caption{Forecasting accuracies of the benchmark models and the proposed TSFM approach using PEFT across different training dataset size on the power consumption (\texttt{Light}) signal. The shades represent the standard deviation across different zones and seasons.}
    \vspace{-1em}
    \label{fig:abl-light-full}
\end{figure}

\begin{figure}[!ht]
    \centering
    \begin{subfigure}[]{0.4\textwidth}
        \centering
        \includegraphics[width=\linewidth]{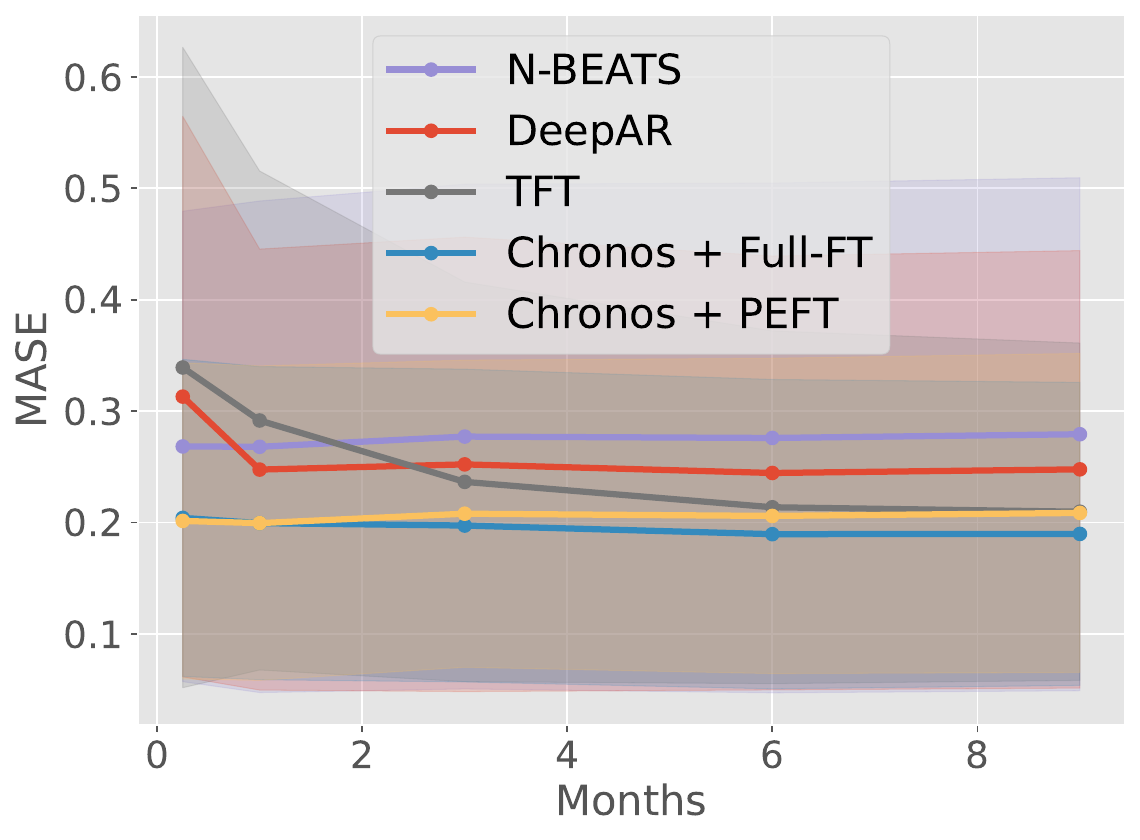}
    \end{subfigure}
    \begin{subfigure}[]{0.4\textwidth}
        \centering
        \includegraphics[width=\linewidth]{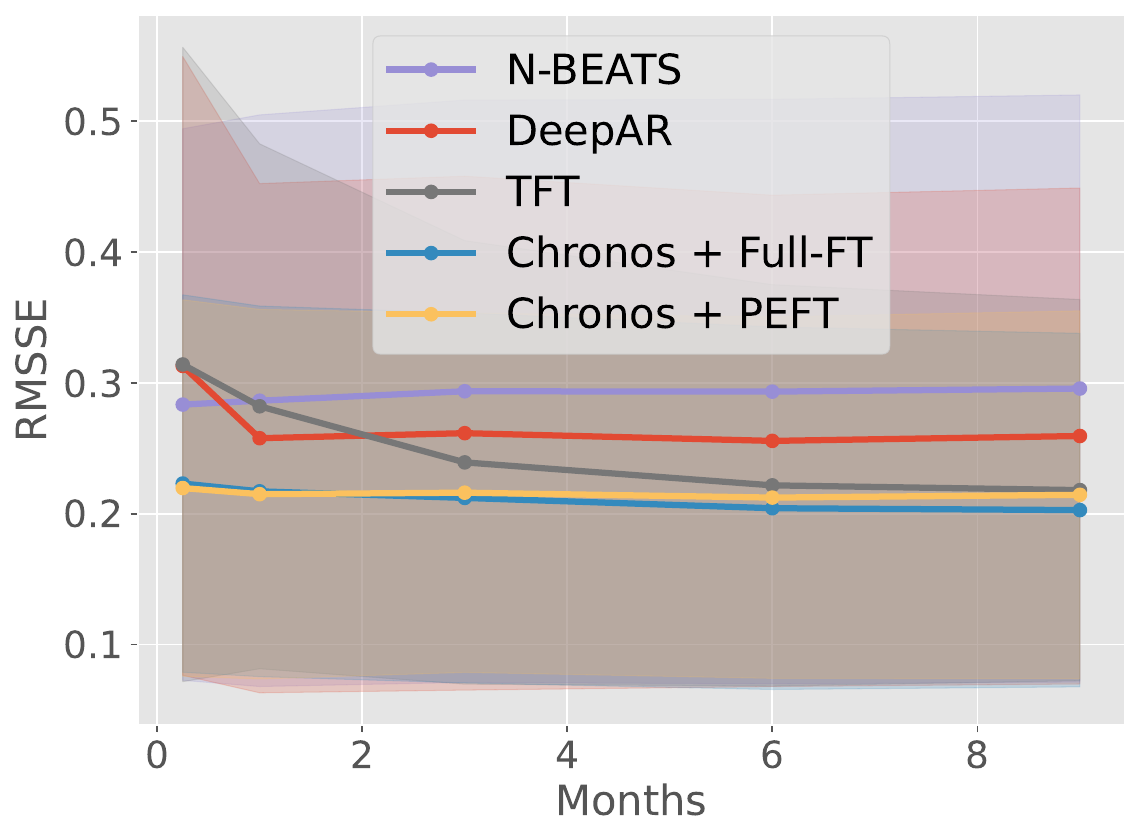}
    \end{subfigure}
    \begin{subfigure}[]{0.4\textwidth}
        \centering
        \includegraphics[width=\linewidth]{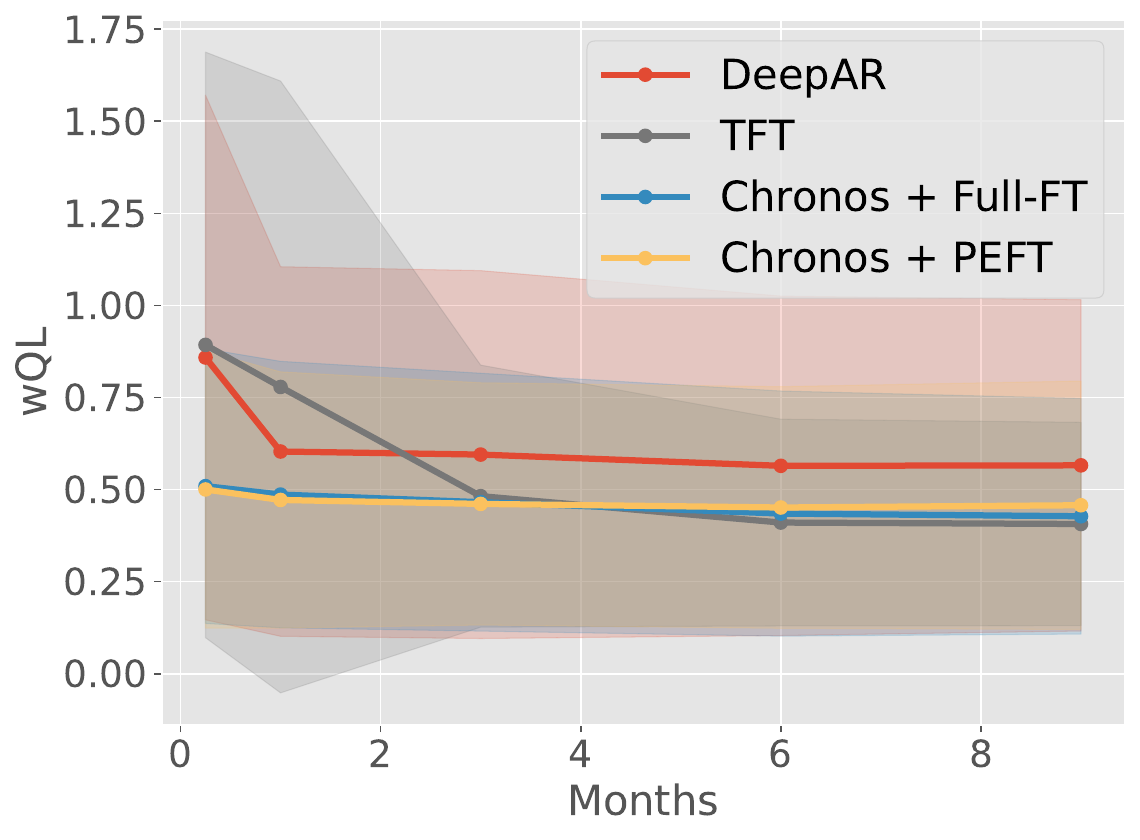}
    \end{subfigure}
    \begin{subfigure}[]{0.4\textwidth}
        \centering
        \includegraphics[width=\linewidth]{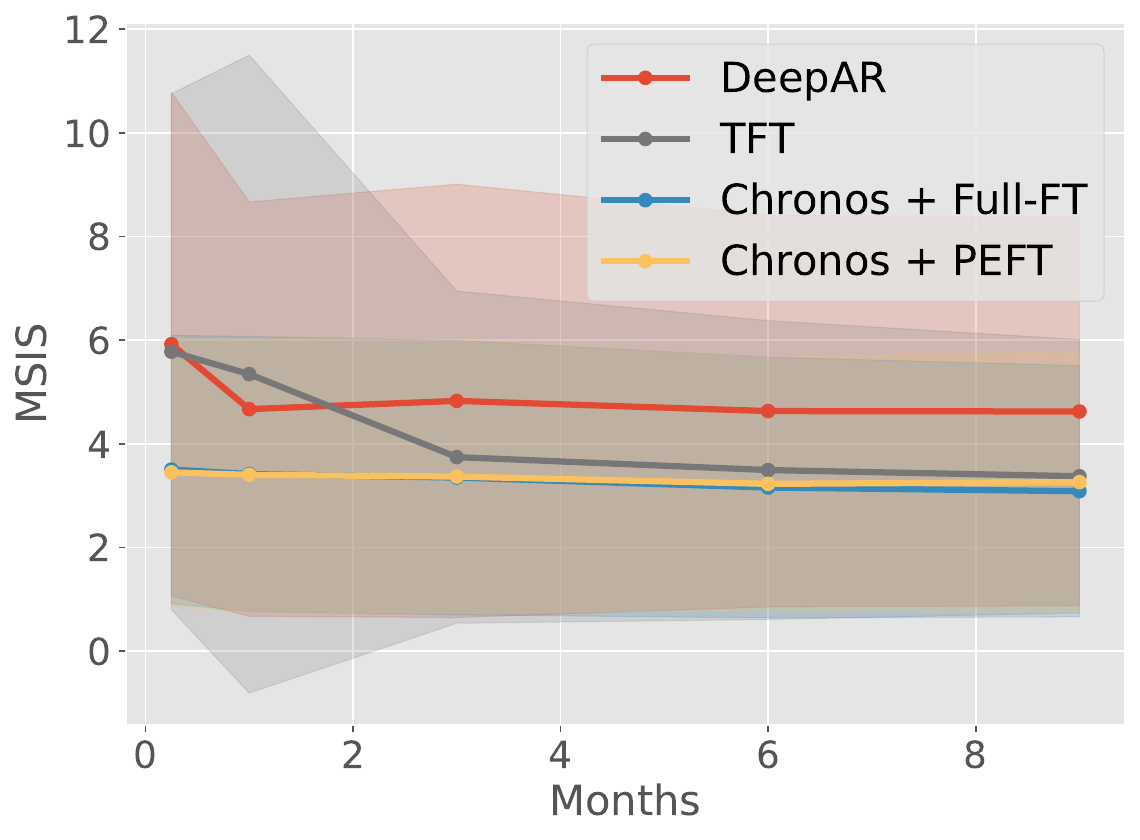}
    \end{subfigure}
    \caption{Forecasting accuracies of the benchmark models and the proposed TSFM approach using PEFT across different training dataset size on the energy consumption (\texttt{HVAC}) signal. The shades represent the standard deviation across different zones and seasons.}
    \vspace{-1em}
    \label{fig:abl-hvac-full}
\end{figure}

We present here the tables and figures omitted from the main text. Table~\ref{tab:comparison_zone_1_8}  complements the results from Section~\ref{sec:result_specific} (i.e. Figures~\ref{fig:zone-occ} and \ref{fig:season-occ}, respectively) with a full breakdown of the predictive performance of TFT and fine-tuned Chronos across different zones and seasons for each signal, respectively. Likewise, Figures \ref{fig:abl-occ-full}, \ref{fig:abl-co2-full}, \ref{fig:abl-light-full}, and \ref{fig:abl-hvac-full} complement the results from Section~\ref{sec:result_robust} (i.e. Figure~\ref{fig:abl-occ}) and report the ablation study results for all 4 signal types.  

A key observation we can make looking at Table~\ref{tab:comparison_zone_1_8} is that the fine-tuned TSFM approach consistently outperforms TFT for the \texttt{CO2} and \texttt{Light} signals under all tested conditions. We see further evidence of this in Figures \ref{fig:abl-co2-full} and \ref{fig:abl-light-full} where fine-tuned TSFM achieves generally higher accuracy and smaller variance on these signals. Notably, this robustness appears even more beneficial when training data are limited, likely because TSFM can leverage related time-series data from its pre-training phase to quickly adapt to downstream tasks.

Another interesting takeaway is that TSFM appears to deliver more robust performance on MASE or RMSSE than on MSIS or wQL. This seems to stem from the correlation between the loss function used to train the TSFM model (i.e., \textsc{Chronos}, in this paper) and the chosen evaluation metrics. Since the TFT model is optimized from scratch to minimize quantile loss, it naturally has a relative advantage in quantile-based evaluation metrics such as MSIS or wQL. In contrast, \textsc{Chronos} adopts a regression-as-classification approach, dividing the output space into finely spaced discrete bins and training the model via cross-entropy loss. Consequently, its loss function is not directly tied to quantile predictions. Nevertheless, the fact that \textsc{Chronos} demonstrates accuracy on par with—or even exceeding—that of a model explicitly optimized for quantile forecasting (i.e., TFT) is a noteworthy finding. In the future, it would be promising to explore ways of fine-tuning TSFM for alignment with domain-specific metrics of interest. Additionally, this may be further indication of the care needed in the more difficult context of probabilistic forecasting, where the priors learned at pre-training would likely be a little less fitting and mitigating overfitting of the massively over-parameterized TSFMs would be more challenging.

\begin{table*}[hbt!]
	\centering
	\caption{Comparison of forecasting performance across different zones (Zones \#1–\#8). The winner is highlighted in light blue. The average scores along with their standard deviations across different seasons are presented together.}
	\label{tab:comparison_zone_1_8}
	\resizebox{.8\linewidth}{!}{%
		\begin{tabular}{ll|rr|rr|rr|rr}
			\toprule
			&  & \multicolumn{2}{c}{\textbf{Zone \#1}} & \multicolumn{2}{c}{\textbf{Zone \#2}} & \multicolumn{2}{c}{\textbf{Zone \#3}} & \multicolumn{2}{c}{\textbf{Zone \#4}} \\
			\cmidrule(lr){3-4} \cmidrule(lr){5-6} \cmidrule(lr){7-8} \cmidrule(lr){9-10}
			\textbf{Signal} & \textbf{Metric} & \textsc{TFT} & \makecell[r]{\textsc{Chronos}\\+PEFT} & \textsc{TFT} & \makecell[r]{\textsc{Chronos}\\+PEFT} & \textsc{TFT} & \makecell[r]{\textsc{Chronos}\\+PEFT} & \textsc{TFT} & \makecell[r]{\textsc{Chronos}\\+PEFT} \\
			\midrule
			\midrule
			\multirow{4}{*}{\texttt{Occ}}
			& MASE  & 0.20 {\scriptsize $\pm$ 0.05} & \cellcolor{bestblue}0.17 {\scriptsize $\pm$ 0.03} & 0.23 {\scriptsize $\pm$ 0.05} & \cellcolor{bestblue}0.21 {\scriptsize $\pm$ 0.05} & 0.19 {\scriptsize $\pm$ 0.03} & \cellcolor{bestblue}0.15 {\scriptsize $\pm$ 0.02} & 0.21 {\scriptsize $\pm$ 0.05} & \cellcolor{bestblue}0.17 {\scriptsize $\pm$ 0.01} \\
			& RMSSE & 0.21 {\scriptsize $\pm$ 0.05} & \cellcolor{bestblue}0.18 {\scriptsize $\pm$ 0.03} & 0.23 {\scriptsize $\pm$ 0.05} & \cellcolor{bestblue}0.22 {\scriptsize $\pm$ 0.05} & 0.19 {\scriptsize $\pm$ 0.03} & \cellcolor{bestblue}0.17 {\scriptsize $\pm$ 0.03} & 0.23 {\scriptsize $\pm$ 0.04} & \cellcolor{bestblue}0.19 {\scriptsize $\pm$ 0.01} \\
			& MSIS  & 3.29 {\scriptsize $\pm$ 1.01} & \cellcolor{bestblue}2.65 {\scriptsize $\pm$ 0.61} & \cellcolor{bestblue}3.13 {\scriptsize $\pm$ 1.04} & 3.42 {\scriptsize $\pm$ 1.19} & \cellcolor{bestblue}2.40 {\scriptsize $\pm$ 0.36} & 2.42 {\scriptsize $\pm$ 0.55} & 3.04 {\scriptsize $\pm$ 1.00} & \cellcolor{bestblue}2.59 {\scriptsize $\pm$ 0.29} \\
			& wQL   & 0.35 {\scriptsize $\pm$ 0.09} & \cellcolor{bestblue}0.32 {\scriptsize $\pm$ 0.05} & \cellcolor{bestblue}0.35 {\scriptsize $\pm$ 0.08} & 0.36 {\scriptsize $\pm$ 0.09} & 0.30 {\scriptsize $\pm$ 0.05} & \cellcolor{bestblue}0.28 {\scriptsize $\pm$ 0.05} & 0.38 {\scriptsize $\pm$ 0.14} & \cellcolor{bestblue}0.31 {\scriptsize $\pm$ 0.02} \\
			\midrule
			\multirow{4}{*}{\texttt{CO2}}
			& MASE  & 0.03 {\scriptsize $\pm$ 0.00} & \cellcolor{bestblue}0.02 {\scriptsize $\pm$ 0.00} & 0.03 {\scriptsize $\pm$ 0.01} & \cellcolor{bestblue}0.02 {\scriptsize $\pm$ 0.00} & 0.03 {\scriptsize $\pm$ 0.00} & \cellcolor{bestblue}0.02 {\scriptsize $\pm$ 0.00} & 0.03 {\scriptsize $\pm$ 0.00} & \cellcolor{bestblue}0.02 {\scriptsize $\pm$ 0.00} \\
			& RMSSE & 0.03 {\scriptsize $\pm$ 0.00} & \cellcolor{bestblue}0.03 {\scriptsize $\pm$ 0.00} & 0.03 {\scriptsize $\pm$ 0.01} & \cellcolor{bestblue}0.03 {\scriptsize $\pm$ 0.00} & 0.03 {\scriptsize $\pm$ 0.00} & \cellcolor{bestblue}0.03 {\scriptsize $\pm$ 0.00} & 0.03 {\scriptsize $\pm$ 0.01} & \cellcolor{bestblue}0.03 {\scriptsize $\pm$ 0.01} \\
			& MSIS  & 0.42 {\scriptsize $\pm$ 0.07} & \cellcolor{bestblue}0.35 {\scriptsize $\pm$ 0.06} & 0.40 {\scriptsize $\pm$ 0.13} & \cellcolor{bestblue}0.33 {\scriptsize $\pm$ 0.07} & 0.42 {\scriptsize $\pm$ 0.10} & \cellcolor{bestblue}0.32 {\scriptsize $\pm$ 0.06} & 0.35 {\scriptsize $\pm$ 0.05} & \cellcolor{bestblue}0.34 {\scriptsize $\pm$ 0.04} \\
			& wQL   & 0.04 {\scriptsize $\pm$ 0.01} & \cellcolor{bestblue}0.04 {\scriptsize $\pm$ 0.01} & 0.04 {\scriptsize $\pm$ 0.01} & \cellcolor{bestblue}0.03 {\scriptsize $\pm$ 0.01} & 0.04 {\scriptsize $\pm$ 0.01} & \cellcolor{bestblue}0.03 {\scriptsize $\pm$ 0.00} & 0.04 {\scriptsize $\pm$ 0.01} & \cellcolor{bestblue}0.03 {\scriptsize $\pm$ 0.00} \\
			\midrule
			\multirow{4}{*}{\texttt{Light}}
			& MASE  & 0.17 {\scriptsize $\pm$ 0.02} & \cellcolor{bestblue}0.14 {\scriptsize $\pm$ 0.02} & 0.16 {\scriptsize $\pm$ 0.04} & \cellcolor{bestblue}0.14 {\scriptsize $\pm$ 0.02} & 0.17 {\scriptsize $\pm$ 0.05} & \cellcolor{bestblue}0.13 {\scriptsize $\pm$ 0.03} & 0.16 {\scriptsize $\pm$ 0.05} & \cellcolor{bestblue}0.14 {\scriptsize $\pm$ 0.04} \\
			& RMSSE & 0.20 {\scriptsize $\pm$ 0.01} & \cellcolor{bestblue}0.17 {\scriptsize $\pm$ 0.02} & 0.19 {\scriptsize $\pm$ 0.03} & \cellcolor{bestblue}0.17 {\scriptsize $\pm$ 0.02} & 0.20 {\scriptsize $\pm$ 0.05} & \cellcolor{bestblue}0.16 {\scriptsize $\pm$ 0.03} & 0.18 {\scriptsize $\pm$ 0.05} & \cellcolor{bestblue}0.17 {\scriptsize $\pm$ 0.04} \\
			& MSIS  & 3.56 {\scriptsize $\pm$ 2.25} & \cellcolor{bestblue}2.14 {\scriptsize $\pm$ 0.26} & 2.36 {\scriptsize $\pm$ 0.59} & \cellcolor{bestblue}2.02 {\scriptsize $\pm$ 0.47} & 2.72 {\scriptsize $\pm$ 0.94} & \cellcolor{bestblue}1.95 {\scriptsize $\pm$ 0.65} & 2.31 {\scriptsize $\pm$ 0.89} & \cellcolor{bestblue}2.07 {\scriptsize $\pm$ 0.94} \\
			& wQL   & 0.30 {\scriptsize $\pm$ 0.15} & \cellcolor{bestblue}0.20 {\scriptsize $\pm$ 0.02} & 0.22 {\scriptsize $\pm$ 0.04} & \cellcolor{bestblue}0.19 {\scriptsize $\pm$ 0.02} & 0.24 {\scriptsize $\pm$ 0.07} & \cellcolor{bestblue}0.19 {\scriptsize $\pm$ 0.04} & 0.22 {\scriptsize $\pm$ 0.07} & \cellcolor{bestblue}0.20 {\scriptsize $\pm$ 0.06} \\
			\midrule
			\multirow{4}{*}{\texttt{HVAC}}
			& MASE  & 0.16 {\scriptsize $\pm$ 0.13} & \cellcolor{bestblue}0.14 {\scriptsize $\pm$ 0.12} & 0.18 {\scriptsize $\pm$ 0.17} & \cellcolor{bestblue}0.15 {\scriptsize $\pm$ 0.15} & 0.17 {\scriptsize $\pm$ 0.12} & \cellcolor{bestblue}0.15 {\scriptsize $\pm$ 0.11} & 0.26 {\scriptsize $\pm$ 0.22} & \cellcolor{bestblue}0.24 {\scriptsize $\pm$ 0.18} \\
			& RMSSE & 0.16 {\scriptsize $\pm$ 0.13} & \cellcolor{bestblue}0.15 {\scriptsize $\pm$ 0.13} & 0.19 {\scriptsize $\pm$ 0.18} & \cellcolor{bestblue}0.17 {\scriptsize $\pm$ 0.16} & 0.18 {\scriptsize $\pm$ 0.12} & \cellcolor{bestblue}0.17 {\scriptsize $\pm$ 0.12} & 0.26 {\scriptsize $\pm$ 0.22} & \cellcolor{bestblue}0.24 {\scriptsize $\pm$ 0.18} \\
			& MSIS  & \cellcolor{bestblue}2.44 {\scriptsize $\pm$ 1.99} & 2.48 {\scriptsize $\pm$ 2.14} & 3.14 {\scriptsize $\pm$ 3.40} & \cellcolor{bestblue}2.79 {\scriptsize $\pm$ 2.84} & 2.67 {\scriptsize $\pm$ 2.25} & \cellcolor{bestblue}2.57 {\scriptsize $\pm$ 2.04} & 4.36 {\scriptsize $\pm$ 4.33} & \cellcolor{bestblue}3.66 {\scriptsize $\pm$ 3.24} \\
			& wQL   & 0.32 {\scriptsize $\pm$ 0.23} & \cellcolor{bestblue}0.32 {\scriptsize $\pm$ 0.25} & \cellcolor{bestblue}0.34 {\scriptsize $\pm$ 0.32} & 0.35 {\scriptsize $\pm$ 0.35} & \cellcolor{bestblue}0.32 {\scriptsize $\pm$ 0.21} & 0.32 {\scriptsize $\pm$ 0.21} & 0.58 {\scriptsize $\pm$ 0.50} & \cellcolor{bestblue}0.54 {\scriptsize $\pm$ 0.43} \\ \bottomrule\toprule
			&  & \multicolumn{2}{c}{\textbf{Zone \#5}} & \multicolumn{2}{c}{\textbf{Zone \#6}} & \multicolumn{2}{c}{\textbf{Zone \#7}} & \multicolumn{2}{c}{\textbf{Zone \#8}} \\
			\cmidrule(lr){3-4} \cmidrule(lr){5-6} \cmidrule(lr){7-8} \cmidrule(lr){9-10}
			\textbf{Signal} & \textbf{Metric} & \textsc{TFT} & \makecell[r]{\textsc{Chronos}\\+PEFT} & \textsc{TFT} & \makecell[r]{\textsc{Chronos}\\+PEFT} & \textsc{TFT} & \makecell[r]{\textsc{Chronos}\\+PEFT} & \textsc{TFT} & \makecell[r]{\textsc{Chronos}\\+PEFT} \\
			\midrule
			\midrule
			\multirow{4}{*}{\texttt{Occ}}
			& MASE  & 0.19 {\scriptsize $\pm$ 0.02} & \cellcolor{bestblue}0.18 {\scriptsize $\pm$ 0.02} & 0.19 {\scriptsize $\pm$ 0.03} & \cellcolor{bestblue}0.17 {\scriptsize $\pm$ 0.02} & 0.20 {\scriptsize $\pm$ 0.03} & \cellcolor{bestblue}0.19 {\scriptsize $\pm$ 0.02} & 0.20 {\scriptsize $\pm$ 0.03} & \cellcolor{bestblue}0.16 {\scriptsize $\pm$ 0.01} \\
			& RMSSE & 0.20 {\scriptsize $\pm$ 0.03} & \cellcolor{bestblue}0.19 {\scriptsize $\pm$ 0.02} & 0.21 {\scriptsize $\pm$ 0.03} & \cellcolor{bestblue}0.17 {\scriptsize $\pm$ 0.02} & 0.21 {\scriptsize $\pm$ 0.03} & \cellcolor{bestblue}0.19 {\scriptsize $\pm$ 0.02} & 0.21 {\scriptsize $\pm$ 0.03} & \cellcolor{bestblue}0.18 {\scriptsize $\pm$ 0.02} \\
			& MSIS  & \cellcolor{bestblue}2.37 {\scriptsize $\pm$ 0.33} & 2.62 {\scriptsize $\pm$ 0.48} & \cellcolor{bestblue}2.41 {\scriptsize $\pm$ 0.28} & 2.57 {\scriptsize $\pm$ 0.44} & 3.02 {\scriptsize $\pm$ 0.67} & \cellcolor{bestblue}3.00 {\scriptsize $\pm$ 0.67} & 2.93 {\scriptsize $\pm$ 1.21} & \cellcolor{bestblue}2.41 {\scriptsize $\pm$ 0.33} \\
			& wQL   & \cellcolor{bestblue}0.31 {\scriptsize $\pm$ 0.04} & 0.32 {\scriptsize $\pm$ 0.04} & \cellcolor{bestblue}0.31 {\scriptsize $\pm$ 0.04} & 0.31 {\scriptsize $\pm$ 0.04} & 0.35 {\scriptsize $\pm$ 0.05} & 0.35 {\scriptsize $\pm$ 0.04} & 0.37 {\scriptsize $\pm$ 0.13} & \cellcolor{bestblue}0.29 {\scriptsize $\pm$ 0.02} \\[1mm]
			\midrule
			\multirow{4}{*}{\texttt{CO2}}
			& MASE  & 0.03 {\scriptsize $\pm$ 0.01} & \cellcolor{bestblue}0.03 {\scriptsize $\pm$ 0.01} & 0.02 {\scriptsize $\pm$ 0.00} & \cellcolor{bestblue}0.02 {\scriptsize $\pm$ 0.00} & 0.03 {\scriptsize $\pm$ 0.01} & \cellcolor{bestblue}0.03 {\scriptsize $\pm$ 0.01} & 0.02 {\scriptsize $\pm$ 0.00} & \cellcolor{bestblue}0.02 {\scriptsize $\pm$ 0.00} \\
			& RMSSE & 0.03 {\scriptsize $\pm$ 0.01} & \cellcolor{bestblue}0.03 {\scriptsize $\pm$ 0.00} & 0.03 {\scriptsize $\pm$ 0.00} & \cellcolor{bestblue}0.03 {\scriptsize $\pm$ 0.00} & 0.03 {\scriptsize $\pm$ 0.00} & \cellcolor{bestblue}0.03 {\scriptsize $\pm$ 0.00} & 0.02 {\scriptsize $\pm$ 0.00} & \cellcolor{bestblue}0.02 {\scriptsize $\pm$ 0.00} \\
			& MSIS  & 0.43 {\scriptsize $\pm$ 0.11} & \cellcolor{bestblue}0.34 {\scriptsize $\pm$ 0.12} & 0.35 {\scriptsize $\pm$ 0.04} & \cellcolor{bestblue}0.28 {\scriptsize $\pm$ 0.06} & 0.27 {\scriptsize $\pm$ 0.06} & \cellcolor{bestblue}0.21 {\scriptsize $\pm$ 0.04} & 0.32 {\scriptsize $\pm$ 0.07} & \cellcolor{bestblue}0.23 {\scriptsize $\pm$ 0.08} \\
			& wQL   & 0.04 {\scriptsize $\pm$ 0.01} & \cellcolor{bestblue}0.04 {\scriptsize $\pm$ 0.01} & 0.04 {\scriptsize $\pm$ 0.01} & \cellcolor{bestblue}0.03 {\scriptsize $\pm$ 0.01} & 0.04 {\scriptsize $\pm$ 0.01} & \cellcolor{bestblue}0.03 {\scriptsize $\pm$ 0.00} & 0.03 {\scriptsize $\pm$ 0.00} & \cellcolor{bestblue}0.03 {\scriptsize $\pm$ 0.00} \\
			\midrule
			\multirow{4}{*}{\texttt{Light}}
			& MASE  & 0.16 {\scriptsize $\pm$ 0.05} & \cellcolor{bestblue}0.14 {\scriptsize $\pm$ 0.02} & 0.16 {\scriptsize $\pm$ 0.05} & \cellcolor{bestblue}0.14 {\scriptsize $\pm$ 0.02} & 0.24 {\scriptsize $\pm$ 0.06} & \cellcolor{bestblue}0.20 {\scriptsize $\pm$ 0.02} & 0.16 {\scriptsize $\pm$ 0.04} & \cellcolor{bestblue}0.14 {\scriptsize $\pm$ 0.02} \\
			& RMSSE & 0.19 {\scriptsize $\pm$ 0.05} & \cellcolor{bestblue}0.17 {\scriptsize $\pm$ 0.02} & 0.19 {\scriptsize $\pm$ 0.05} & \cellcolor{bestblue}0.17 {\scriptsize $\pm$ 0.02} & 0.25 {\scriptsize $\pm$ 0.06} & \cellcolor{bestblue}0.22 {\scriptsize $\pm$ 0.03} & 0.18 {\scriptsize $\pm$ 0.03} & \cellcolor{bestblue}0.17 {\scriptsize $\pm$ 0.02} \\
			& MSIS  & 2.48 {\scriptsize $\pm$ 0.96} & \cellcolor{bestblue}2.02 {\scriptsize $\pm$ 0.59} & 2.37 {\scriptsize $\pm$ 0.45} & \cellcolor{bestblue}1.97 {\scriptsize $\pm$ 0.39} & 3.87 {\scriptsize $\pm$ 1.27} & \cellcolor{bestblue}2.86 {\scriptsize $\pm$ 0.36} & 1.99 {\scriptsize $\pm$ 0.29} & \cellcolor{bestblue}1.96 {\scriptsize $\pm$ 0.36} \\
			& wQL   & 0.23 {\scriptsize $\pm$ 0.07} & \cellcolor{bestblue}0.20 {\scriptsize $\pm$ 0.04} & 0.23 {\scriptsize $\pm$ 0.03} & \cellcolor{bestblue}0.20 {\scriptsize $\pm$ 0.03} & 0.34 {\scriptsize $\pm$ 0.11} & \cellcolor{bestblue}0.27 {\scriptsize $\pm$ 0.03} & 0.22 {\scriptsize $\pm$ 0.05} & \cellcolor{bestblue}0.20 {\scriptsize $\pm$ 0.03} \\
			\midrule
			\multirow{4}{*}{\texttt{HVAC}}
			& MASE  & 0.26 {\scriptsize $\pm$ 0.22} & \cellcolor{bestblue}0.24 {\scriptsize $\pm$ 0.16} & 0.27 {\scriptsize $\pm$ 0.22} & \cellcolor{bestblue}0.25 {\scriptsize $\pm$ 0.18} & 0.32 {\scriptsize $\pm$ 0.24} & \cellcolor{bestblue}0.27 {\scriptsize $\pm$ 0.16} & 0.28 {\scriptsize $\pm$ 0.19} & \cellcolor{bestblue}0.21 {\scriptsize $\pm$ 0.09} \\
			& RMSSE & 0.18 {\scriptsize $\pm$ 0.12} & \cellcolor{bestblue}0.17 {\scriptsize $\pm$ 0.12} & 0.18 {\scriptsize $\pm$ 0.12} & \cellcolor{bestblue}0.17 {\scriptsize $\pm$ 0.12} & 0.26 {\scriptsize $\pm$ 0.22} & \cellcolor{bestblue}0.24 {\scriptsize $\pm$ 0.18} & 0.28 {\scriptsize $\pm$ 0.21} & \cellcolor{bestblue}0.26 {\scriptsize $\pm$ 0.18} \\
			& MSIS  & 4.69 {\scriptsize $\pm$ 4.14} & \cellcolor{bestblue}3.91 {\scriptsize $\pm$ 3.22} & \cellcolor{bestblue}4.01 {\scriptsize $\pm$ 3.78} & 4.05 {\scriptsize $\pm$ 3.79} & 5.02 {\scriptsize $\pm$ 4.86} & \cellcolor{bestblue}4.38 {\scriptsize $\pm$ 3.55} & 3.61 {\scriptsize $\pm$ 1.87} & \cellcolor{bestblue}3.12 {\scriptsize $\pm$ 1.62} \\
			& wQL   & 0.57 {\scriptsize $\pm$ 0.44} & \cellcolor{bestblue}0.56 {\scriptsize $\pm$ 0.40} & \cellcolor{bestblue}0.57 {\scriptsize $\pm$ 0.43} & 0.59 {\scriptsize $\pm$ 0.49} & 0.67 {\scriptsize $\pm$ 0.48} & \cellcolor{bestblue}0.60 {\scriptsize $\pm$ 0.37} & 0.50 {\scriptsize $\pm$ 0.27} & \cellcolor{bestblue}0.42 {\scriptsize $\pm$ 0.17} \\[1mm]
			\bottomrule
		\end{tabular}%
}
\end{table*}

\newpage

\end{document}